\documentclass[preprint,12pt]{elsarticle}
\usepackage{times}
\usepackage{amsmath}
\usepackage{amssymb}
\usepackage{graphicx}
\usepackage{xcolor}
\usepackage{mathrsfs}
\usepackage{caption}

\usepackage[bottom]{footmisc}
\usepackage{bm}
\usepackage{algorithm}
\usepackage{algpseudocode}

\usepackage{mathtools}
\usepackage{subcaption}

\usepackage{float}
\usepackage{adjustbox}
\usepackage[margin=1in]{geometry}
\usepackage[small,compact]{titlesec}
\usepackage[labelfont=bf,labelsep=colon,skip=2pt]{caption} 
\usepackage[titles]{tocloft}
\usepackage{bold-extra}
\usepackage{hyperref}

\usepackage{changes}
\definechangesauthor[name={Reviewer 1}, color = blue]{R1}
\definechangesauthor[name={Reviewer 2}, color = red]{R2}
\definechangesauthor[name={All reviewers}, color = green]{All}
\definechangesauthor[name={Editor}, color = cyan]{Editor}
\definechangesauthor[name={Author}, color = olive]{Author}

\definechangesauthor[name={Deleted}, color = gray]{Deleted}

\DeclareMathOperator*{\argmax}{arg\,max}
\DeclareMathOperator*{\tr}{tr}

\usepackage[left]{lineno}

\begin{document}
\begin{frontmatter}
\title{Bayesian Experimental Design for Model Discrepancy Calibration: A Rivalry between Kullback--Leibler Divergence \\and Wasserstein Distance}

\author{Huchen Yang}
\author{Xinghao Dong}
\author{Jin-Long Wu\corref{cor1}} \ead{jinlong.wu@wisc.edu} 
\cortext[cor1]{Corresponding author}

\address{Department of Mechanical Engineering, University of Wisconsin–Madison, Madison, WI 53706}

\begin{abstract}
Designing experiments that systematically gather data from complex physical systems is central to accelerating scientific discovery. While Bayesian experimental design (BED) provides a principled, information-based framework that integrates experimental planning with probabilistic inference, the selection of utility functions in BED is a long-standing and active topic, where different criteria emphasize different notions of information. Although Kullback--Leibler (KL) divergence has been one of the most common choices, recent studies have proposed Wasserstein distance as an alternative. In this work, we first employ a toy example to illustrate an issue with the Wasserstein distance: the value assigned to a fixed-shape posterior depends on the relative position of its main mass within the support and can exhibit false rewards unrelated to information gain, especially with a non-informative prior (e.g., uniform distribution). We then further provide a systematic comparison between these two criteria through a classical source inversion problem in the BED literature, revealing that the KL divergence tends to lead to faster convergence in the absence of model discrepancy, while Wasserstein metrics provide more robust sequential BED results if model discrepancy is non-negligible. These findings clarify the trade-offs between KL divergence and Wasserstein metrics for the utility function and provide guidelines for selecting suitable criteria in practical BED applications.
\end{abstract}

\begin{keyword}
Bayesian experimental design \sep Wasserstein distance \sep Model discrepancy
\end{keyword}

\end{frontmatter}

\section{Introduction}

Quantifying the differences between probability distributions is a fundamental task across machine learning~\cite{sriperumbudur2012empirical, reid2011information}, statistics~\cite{pardo2018statistical,gibbs2002choosing}, biology~\cite{chanda2020information, adami2004information}, and scientific computing~\cite{owhadi2013optimal, stuart2010inverse}. Various types of methods have been proposed for this purpose, including the family of $f$–divergences such as Kullback--Leibler (KL) and Jensen–Shannon (JS) divergences \cite{csiszar1967information,liese2006divergences,lin2002divergence, sriperumbudur2009integral}, as well as integral probability metrics such as the Wasserstein, energy and Kolmogorov distances~\cite{muller1997integral,marsaglia2003evaluating,rizzo2016energy, Stephens1992}.

In recent years, there has been a growing interest in exploring the Wasserstein distance as an alternative to the commonly used KL divergence in machine learning and statistical inference tasks~\cite{peyre2019computational, villani2008optimal, kato2023unified}. In the field of generative models, it was shown in~\cite{arjovsky2017wasserstein} that replacing KL- or JS-based losses with Wasserstein distance (e.g., in generative adversarial networks) leads to more stable training and alleviates mode collapse, while \cite{tolstikhin2017wasserstein} demonstrated that Wasserstein auto-encoders improve sample quality compared to standard VAEs that essentially rely on KL divergence. For variational inference, \cite{ambrogioni2018wasserstein} proposed Wasserstein variational inference and showed that it remains stable even when the true and approximate distributions have disjoint supports, whereas \cite{yi2023sliced} argued that sliced Wasserstein distance provides a proper metric and avoids the asymmetry and instability issues of KL divergence. In robust optimization, \cite{namkoong2016stochastic} analyzed KL-based distributional robustness and showed that it effectively reweights samples with higher losses to improve generalization guarantees, while \cite{sinha2017certifying} highlighted that Wasserstein distance naturally captures input perturbations and provides stronger robustness against adversarial attacks. In machine learning theory, \cite{rodriguez2021tighter} proved that Wasserstein-based generalization bounds are tighter than KL-based ones, and \cite{lee2018minimax} further showed that Wasserstein distance provides guarantees for domain adaptation when the source and target distributions are close in this metric. 

In Bayesian inference, the two distributions of interest are typically the prior and the posterior over latent quantities. The change from prior to posterior is interpreted as information gained from the data. Building on this perspective, Bayesian experimental design (BED) provides a principled framework for optimizing data acquisition by leveraging Bayesian inference to guide the design process \cite{chaloner1995bayesian, lindley1956measure, lindley1972bayesian}. Rooted in information theory, BED aims to identify experimental conditions that maximize the expected information gain about target latent variables. This probabilistic approach not only ensures that experimental designs are tailored to the specific goals of the study but also allows for adaptive, sequential experimentation, where newly acquired data refine subsequent designs \cite{jones2016bayes, rainforth2024modern, huan2024optimal}. Various popular utility functions and criteria for selecting optimal designs have been summarized in~\cite{chaloner1995bayesian, ryan2016review}. Among them, the KL divergence between the prior and posterior distributions \cite{raiffa2000applied, degroot2005optimal, sebastiani2000maximum} is one of the most common choices, as it provides a straightforward measure of the knowledge gain due to an experiment by quantifying the difference between two distributions. Formally, when the expectation is taken with respect to the prior predictive distribution of the data, this expected KL divergence coincides with the mutual information between the parameters and the observations \cite{huan2024optimal, huan2013simulation, houlsby2011bayesian}.

Despite its popularity in Bayesian inference tasks to evaluate information gain, KL divergence also presents certain drawbacks. Its asymmetry, combined with the commonly adopted form ``posterior relative to prior'', induces a statistical preference toward designs that concentrate posterior mass more sharply, rather than accounting for multiple plausible modes \cite{helin2025bayesian}. These characteristics are generally not problematic when the forward model is well specified. Because in these cases, the posterior mainly refines the prior by concentrating its mass within the main density region, without relocating it elsewhere. However, under model discrepancy \cite{kennedy2001bayesian, nott2023bayesian}, where likelihood misspecification distorts the update, the posterior’s high-probability region may lie far from the prior’s \cite{grunwald_inconsistency_2017, catanach2023metrics}. In such cases, the value of KL divergence can become extremely large, which in turn may cause the expected information gain to exhibit unstable gradients or to favor design choices that induce wrong posteriors \cite{YANG2025118198}. Recent work has proposed using the Wasserstein distance, instead of the KL divergence, as the criterion for quantifying the difference between prior and posterior distributions~\cite{helin2025bayesian} in BED. This work rigorously established that the expected Wasserstein utility is a well-posed and proper Bayesian information measure, satisfying key axioms such as sufficiency ordering. Among the various properties they highlight, two are particularly important in the context of experimental design: (i) the Wasserstein distance admits a clear geometric interpretation in terms of the optimal transport cost required to transform one distribution into another; and (ii) unlike KL divergence, it remains well-defined even when the prior and posterior have non-overlapping supports, making it especially suitable for applications involving surrogate or empirical models. 

The geometric interpretation of the Wasserstein distances (e.g., earth mover's distance as 1-Wasserstein distance) leads to an important observation: starting from a uniform prior, moving all probability mass to two posterior distributions (e.g., Gaussian) of the same shape but centered at different locations incurs different transport costs. This behavior stands in sharp contrast to the KL divergence: for a uniform prior, two posteriors of the same shape yield the same value of KL divergence regardless of their relative location. This sensitivity of the Wasserstein distance to the absolute location of probability mass is not specific to BED. Several works in optimal transport and statistics have noted that classical Wasserstein distances are inherently sensitive to global translations~\cite{panaretos2019statistical,chhachhi20231}. In some applications, such as comparing images or spatio–temporal fields, this sensitivity to global shifts has been viewed as a limitation, which has motivated the construction of translation–invariant or relative–translation–invariant variants of Wasserstein distance that explicitly factor out rigid displacements~\cite{wang2024relative,sejourne2022faster}. These studies show that Wasserstein distances inherently couple shape and location information, in contrast to divergences that depend only on local density ratios. To the best of our knowledge, however, this location dependence of Wasserstein distances has not been systematically analyzed in the context of information gain via Bayesian inference, nor have its consequences for experimental design utilities been explored in detail.

On the other hand, the performance of BED relies on the accuracy of the forward model~\cite{rainforth2024modern}, while the model discrepancy~\cite{kennedy2001bayesian,feng_optimal_2015} is inevitable in most real-world applications. The model discrepancy primarily introduces errors in the forward evaluation of BED, leading to the following consequences: (i) introduction of bias in parameter estimation; (ii) continuous generation of similar designs and low-quality datasets, which present significant challenges in BED practice \cite{grunwald_inconsistency_2017,brynjarsdottir2014learning,catanach2023metrics}. Although recent developments in residual learning~\cite{levine2022framework, ebers2024discrepancy, dong2025data, wu2024learning, YANG2025118198} are promising for characterizing the model discrepancy, they often involve neural networks with a large number of unknown coefficients. Although systematically gathering informative training data for network calibration could also be addressed by BED, updating beliefs and evaluating information gain in the high-dimensional network parameter space via a full Bayesian perspective (e.g., Monte Carlo methods) are often computationally infeasible. Recently, an automatic differentiable ensemble Kalman inversion (AD-EKI) approach we proposed~\cite{yang2025bayesian} efficiently addresses the belief updating of high-dimensional network parameters for model discrepancies within the BED framework, building on the ensemble Kalman methods~\cite{iglesias2013ensemble, chen2022autodifferentiable} that have been actively studied in the past decade as an efficient derivative-free method for Bayesian inverse problems. 

In this work, we assess the use of the Wasserstein distance as the utility function of BED, both in the absence and in the presence of model discrepancy, using the AD-EKI framework~\cite{yang2025bayesian}. We consider both parametric and structural forms of model discrepancy in different examples, and investigate a comprehensive comparison between Wasserstein distance and KL divergence. The key highlights of our work are summarized below:
\begin{itemize}
    \item We investigate a limitation of using Wasserstein distance to construct the utility function in BED -- when compared with an approximately uniform prior, its location dependence can induce false reward unrelated to information gain.
    \item We compare Wasserstein distance and KL divergence without model discrepancy. Starting from a uniform prior, Wasserstein distance tends to favor designs whose induced posteriors place more mass near the boundary of the prior support. Though this effect is most pronounced only in the early stages, our experiments further show that such design preference often retains higher posterior uncertainty, thereby slowing down belief updating in subsequent stages of prior-based EIG methods.
    \item We extend the comparison to settings with model discrepancy, and show under actual utilities that KL's posterior tends to over-concentrate on incorrect regions, while Wasserstein designs preserve uncertainty around the true parameter and correct both belief and model more effectively. When switching to expected utilities, the performance gap narrows because the expected procedure introduces additional uncertainty, and the metric that retains more probability mass around the truth achieves faster updating and convergence.

\end{itemize}

This paper is organized as follows. Section \ref{sec: Methodology} introduces the general formulation of BED with model discrepancy, using the Wasserstein distance to construct the utility function. Section \ref{sec: Numerical Results} provides a comprehensive study of a classical BED problem to facilitate the comparison between the performances of Wasserstein distance and KL divergence. Section \ref{Conclusion} concludes the paper.

\section{Methodology}\label{sec: Methodology}

\subsection{Bayesian experimental design}\label{Sec: BED}

The Bayesian experimental design~\cite{lindley1972bayesian, rainforth2024modern,huan2024optimal} provides a general framework to systematically seek the optimal design by solving the optimization problem: 
\begin{equation}
\label{eq:oed_opt}
    \begin{aligned}
        \mathbf{d}^* &= \argmax_{\mathbf{d}\in\mathcal{D}} \mathbb{E}[U(\boldsymbol{\theta},\mathbf{y},\mathbf{d})]\\
        &= \argmax_{\mathbf{d}\in\mathcal{D}} \int_\mathcal{Y} \int_{\boldsymbol{\Theta}} U(\boldsymbol{\theta},\mathbf{y},\mathbf{d}) p(\boldsymbol{\theta},\mathbf{y}|\mathbf{d}) \mathrm{d}\boldsymbol{\theta}\mathrm{d}\mathbf{y},
    \end{aligned}
\end{equation}
where $\mathbf{y} \in \mathcal{Y} \subset \mathbb{R}^{d_{\mathbf{y}}}$ is data from the experimental design $\mathbf{d} \in \mathcal{D}\subset \mathbb{R}^{d_{\mathbf{d}}}$, $\boldsymbol{\theta} \in \boldsymbol{\Theta}\subset \mathbb{R}^{d_{\boldsymbol{\theta}}}$ denotes the target parameters, and $U$ is the utility function taking the inputs of $\mathbf{y}$ and $\boldsymbol{\theta}$ and returning a real value, which reflects the specific purpose of designing the experiment. The optimal design $\mathbf{d}^*$ is obtained by maximizing the expected utility function $\mathbb{E}[U(\boldsymbol{\theta},\mathbf{y},\mathbf{d})]$ over the design space $\mathcal{D}$. The term $p(\boldsymbol{\theta},\mathbf{y}|\mathbf{d})$ is the joint conditional distribution of data and parameters.

The utility function $U(\boldsymbol{\theta},\mathbf{y},\mathbf{d})$ can be regarded as the information gain obtained from the data $\mathbf{y}$ for the corresponding design $\mathbf{d}$. A most common choice in Bayesian setting is the Kullback-Leibler (KL) divergence between the posterior and the prior distribution of parameters $\boldsymbol{\theta}$:
\begin{equation}
    \begin{aligned}
        U(\boldsymbol{\theta},\mathbf{y},\mathbf{d}) &= D_{\textrm{KL}} \bigl(p(\boldsymbol\theta|\mathbf{y},\mathbf{d})\| p(\boldsymbol\theta)\bigr)\\
        &=\mathbb{E}_{\boldsymbol\theta|\mathbf{d},\mathbf{y}}(\log p(\boldsymbol{\theta}|\mathbf{y},\mathbf{d}) - \log p(\boldsymbol{\theta}))\\
        &=\int_{\boldsymbol{\Theta}} p(\boldsymbol{\theta}|\mathbf{y},\mathbf{d}) \log(\frac{p(\boldsymbol{\theta}|\mathbf{y},\mathbf{d})}{p(\boldsymbol{\theta})})\mathrm{d}\boldsymbol{\theta},
    \end{aligned}
    \label{eq: kld utility}
\end{equation}
where $p(\boldsymbol\theta|\mathbf{y},\mathbf{d})$ is the posterior distribution of $\boldsymbol\theta$ given a design $\mathbf{d}$ and the data $\mathbf{y}$. Note that data $\mathbf{y}$ could be either from the actual experimental measurement or the numerical model simulation. The use of KL divergence as the utility function leads to the definition of expected information gain (EIG,~\cite{lindley1956measure}) that integrates the information gain over all possible predicted data:
\begin{equation}
\label{eq:EIG}
    \begin{aligned}
        \text{EIG} (\mathbf{d}) &=\mathbb{E}_{\mathbf{y} | \mathbf{d}} [D_{\textrm{KL}}(p(\boldsymbol\theta|\mathbf{y},\mathbf{d})\| p(\boldsymbol\theta))]\\
  &=\int_{\mathcal{Y}}D_{\textrm{KL}}(p(\boldsymbol\theta|\mathbf{y},\mathbf{d})\| p(\boldsymbol\theta)) p(\mathbf{y}|\mathbf{d})\mathrm{d}\mathbf{y},\\      
    \end{aligned}
\end{equation}
where $p(\mathbf{y}|\mathbf{d}):=\mathbb{E}_{\boldsymbol{\theta}}[p(\mathbf{y}|\boldsymbol{\theta},\mathbf{d})]$ is the distribution of the predicted data among all possible parameter values given a certain design.

\subsection{Wasserstein distance criteria}

Recent work \cite{helin2025bayesian} has proposed using the Wasserstein distance, which is also a metric to quantify divergence between two distributions, as the utility function in BED. Let $\rho_0$ and $\rho_1$ be two probability measures on a domain $\Omega \subset \mathbb{R}^d$. The Wasserstein-$p$ distance, with $p \ge 1$, is defined as
\begin{equation}
    W_p(\rho_0, \rho_1) = \left(\inf_{\gamma \in \Gamma(\rho_0, \rho_1)} 
    \int_{\Omega \times \Omega} \| x - y \|^p \, d\gamma(x, y)\right)^{1/p},
\end{equation}
where $\Gamma(\rho_0, \rho_1)$ denotes the set of all couplings with marginals $\rho_0$ and $\rho_1$. 
In this study, we focus on $p=2$ and $p=1$, corresponding to the $W_2$ and $W_1$ distances, respectively. 
For Gaussian measures $\rho_0 = \mathcal{N}(m_0, C_0)$ and $\rho_1 = \mathcal{N}(m_1, C_1)$, the $W_2$ distance admits the well-known closed-form expression \cite{gelbrich1990formula}:
\begin{equation}
\label{eq:w2_gaussian}
    W_2(\rho_0, \rho_1) = \left[\| m_0 - m_1 \|_2^2 
    + \tr \left( C_0 + C_1 - 2 \left( C_1^{1/2} C_0 C_1^{1/2} \right)^{1/2} \right)\right]^{1/2},
\end{equation}
where $\tr(\cdot)$ denotes the trace operator. 
The $W_1$ distance between Gaussian measures does not generally admit a simple closed form, except in one dimension.

In most practical scenarios of BED, the prior and posterior distributions of the physical parameters are non-Gaussian due to nonlinear forward models and non-Gaussian priors. In such cases, the Gaussian closed form \eqref{eq:w2_gaussian} is no longer applicable, and the computation of $W_2$ or $W_1$ must rely on numerical methods for optimal transport.

In \cite{helin2025bayesian}, the authors address the non-Gaussian setting by solving the Monge--Amp\`ere PDE associated with the optimal transport map. 
Specifically, they represent the convex potential function via a radial basis function (RBF) expansion and employ the Kansa collocation method to solve the PDE under convexity constraints. 
Once the optimal map is obtained, the Wasserstein distance is computed by numerical integration over the parameter domain.

In the present work, to facilitate automatic differentiation, we adopt a simpler and more flexible computational strategy based on the discrete optimal transport formulation. Let $\mu=\sum_{i=1}^n a_i \delta_{x_i}$ and $\nu=\sum_{j=1}^m b_j \delta_{y_j}$ be empirical measures of the prior and posterior, with $a_i,b_j>0$ and $\sum_i a_i=\sum_j b_j=1$. 
We approximate $W_2$ using the entropic-regularized discrete optimal transport formulation
\begin{equation}
\label{eq:sinkhorn_primal}
\mathrm{OT}_{\varepsilon}^{(2)}(\mu,\nu)
=\min_{\gamma\in\mathbb{R}_+^{n\times m}}
\left\langle C,\gamma\right\rangle
+\varepsilon\, D_\text{KL}\!\left(\gamma\,\middle\|\, a\,b^\top\right)
\quad \text{s.t.}\quad
\gamma \mathbf{1}_m = a,\;\; \gamma^\top \mathbf{1}_n = b,
\end{equation}
where $C_{ij}=\|x_i-y_j\|^2$ is the quadratic ground cost, $\langle C,\gamma\rangle=\sum_{ij} C_{ij}\gamma_{ij}$, and $D_\text{KL}(\gamma\|ab^\top)=\sum_{ij}\gamma_{ij}\log(\gamma_{ij}/(a_i b_j))$. 
The regularization parameter $\varepsilon>0$ smooths the transport plan and enables fast convergence of the Sinkhorn iterations implemented in \texttt{ott-jax}. In practice, we form the empirical weight vectors $a$ and $b$ from normalized histograms of the prior and posterior samples, apply a small jitter to avoid zero entries, and solve~\eqref{eq:sinkhorn_primal} with the \texttt{LinearProblem} and \texttt{Sinkhorn} solver in \texttt{ott-jax}. 
The approximate Wasserstein-2 distance is then obtained as
\begin{equation}
    \widehat{W}_2(\mu,\nu) = \sqrt{\mathrm{OT}_{\varepsilon}^{(2)}(\mu,\nu)},
\end{equation}
where $\mathrm{OT}_{\varepsilon}^{(2)}(\mu,\nu)$ is returned by the solver as the regularized OT cost.
We note that, due to entropic regularization, $\widehat{W}_2$ is a biased approximation to the unregularized $W_2$, but offers substantial computational speedup, GPU scalability, and automatic differentiation, which are critical in repeated evaluations of EIG and gradient-based design search of BED. A sensitivity check for the regularization parameter can be found in~\ref{apd:sensitivity}. The results show that the regularization parameter does not appreciably change the utility landscape or the resulting design ranking.

\subsection{False reward of Wasserstein distance}

Although Wasserstein distance admits several advantages as a utility function in BED, it might mislead design choices. The underlying reason is that, even when two designs yield posteriors with the same shape but centered at different locations, the one located nearer to the boundary of the prior’s bounded support will register a larger Wasserstein distance, purely as a boundary effect. This increase in distance does not necessarily reflect a true gain in information, but rather arises from the location-dependent nature of the Wasserstein metric. When such cases occur, the Wasserstein utility may favor a suboptimal design that produces a posterior with greater uncertainty.

This location effect is further compounded when the posterior is more diffuse. Although such posteriors may cover the true value, their mass is spread more broadly, and most of it may still be located far from the true value. Moving prior mass to these far-away regions can incur a larger transport cost than moving it to the true value, which can result in a higher Wasserstein distance despite the posterior being less informative.

To illustrate this location-dependent effect, consider a one-dimensional parameter space with a uniform prior on the support $[-A, A]$. We examine three posterior distributions:

\begin{itemize}
    \item a delta distribution at 0, $\delta(0)$;
    \item two symmetric delta distributions, $0.5\delta(-\Delta x)+0.5\delta(\Delta x)$;
    \item a shifted symmetric delta distribution, $0.5\delta(-\Delta x+\mu)+0.5\delta(\Delta x+\mu)$, with  $\Delta x < A$ and $\Delta x +\mu< A$.
\end{itemize}

These three distributions can be viewed as a single delta that is first split symmetrically into two equally weighted deltas and then shifted together along the axis. We give the analytical solutions of the Wasserstein-2 distance between them and a uniform prior, respectively, to show that this false reward does happen.

The Wasserstein-2 distances of these three distributions from the uniform prior are:
\begin{equation}
    W_2^{(1)}=A/\sqrt{3},\quad W_2^{(2)}=\sqrt{\Delta x^2-A\Delta x + A^2/3},\quad W_2^{(3)}=\sqrt{\mu^2+\Delta x^2-A\Delta x + A^2/3},
\end{equation}
and detailed derivations can be found in \ref{appendix: derivation}.

Intuitively, a delta distribution at the center of the support exhibits greater discrepancy from the uniform prior than two equally weighted deltas symmetrically located within the support. Consequently, $ W_2^{(1)} \geq W_2^{(2)}$, with equality only when $\Delta x = 0$ or $\Delta x = A$. However, when the symmetric pair is shifted toward the boundary, its corresponding Wasserstein distance will increase, i.e., $W_2^{(3)} \geq W_2^{(2)}$ . For sufficiently large $\mu$, specifically when $\mu^2 > A\Delta x - \Delta x^2$, we have $W_2^{(3)} = \sqrt{\mu^2 + \Delta x^2 - A\Delta x + A^2/3} > W_2^{(1)} = A/\sqrt{3}$. In this regime, the Wasserstein distance for the two-point configuration exceeds that for the single point mass.

This example shows that a more diffuse distribution, if shifted toward the boundaries, can be assigned a larger Wasserstein distance from the uniform prior than a concentrated distribution at the center. In other words, the location-dependent property of Wasserstein distance can make a shifted, diffuse distribution appear to carry more information relative to the prior, even when it does not. Once the symmetric delta distribution is shifted toward the boundaries, some prior mass from the opposite side must be transported across the entire domain to reach the posterior’s concentration region. This additional transport inflates the distance metric and creates a false information gain in the design context.

It is noteworthy that, for fixed-shape posteriors shifted within the support of a uniform prior, the KL divergence remains unchanged regardless of the shift, underscoring a key difference between Wasserstein distance and KL divergence as measures of distributional discrepancy.

In summary, this simplified setting illustrates why Wasserstein-based and KL-based design choices can disagree. A diffuse posterior shifted toward the boundary may appear more different from the prior in terms of Wasserstein distance than a concentrated posterior at the center, despite being no more informative. This location-dependent property can assign high utility values to certain posteriors, leading to design selections that differ from those favored by KL divergence. While the posteriors here are chosen arbitrarily for illustration, more design-relevant examples are presented in Section~\ref{sec: Numerical Results}.

\subsection{Correction strategy under model discrepancy}\label{Sec: methodology framework}

We not only compare the performance between classical KL divergence and Wasserstein distance in the traditional setting (i.e., exact model), but also in the presence of model discrepancy. When there is no model discrepancy, we follow the standard sequential BED framework with a greedy strategy in each experiment as described in Section~\ref{Sec: BED}. However, model discrepancy is ubiquitous in real-world applications, posing challenges to both forward and inverse problems, not to mention active learning and BED. Under model discrepancy scenarios, we build on the framework in \cite{yang2025bayesian} and additionally specify a re-update process to improve effectiveness for the expected utilities. Although this framework has proved effective in structural error cases with data-driven methods, we consider parametric error in the main-text numerical example, as our focus is to compare different metrics rather than dealing with complicated structural discrepancies. Also, the parametric error will make it easier to compare the performance gap between different metrics. An additional example with structural model discrepancy and neural-network-based correction is presented in the appendix.

Specifically, to define the model discrepancy, the true system is written in general form
\begin{equation}
    \mathbf{y}=\mathcal{G}^\dagger(\boldsymbol{\theta}_\mathcal{G};\mathbf{d})+\boldsymbol\epsilon, \quad \boldsymbol{\epsilon} \sim \mathcal{N}(\mathbf{0}, \Sigma^2),
\end{equation}
while the incorrect model is
\begin{equation}
    \mathbf{y}=\mathcal{G}(\boldsymbol{\theta}_\mathcal{G},\boldsymbol{\theta}_\mathcal{E} ;\mathbf{d}),
\end{equation}
where $\boldsymbol{\theta}_\mathcal{G}$ represents the original physical parameter, $\boldsymbol{\theta}_\mathcal{E}$ represents the error parameter, and $\boldsymbol\epsilon$ represents the zero-mean Gaussian noise with covariance matrix $\Sigma^2$. Note that the system $\mathcal{G}^\dagger$ and model $\mathcal{G}$ may also serve as the right-hand-side of dynamic systems, and in that case design $\mathbf{d}$ usually appears in the additional measuring operator, rather than the input of dynamic systems. To calibrate the model discrepancy, we seek to optimize the error parameter $\boldsymbol{\theta}_\mathcal{E}$ such that the model $\mathcal{G}$ could approximate the system $\mathcal{G}^\dagger$.

In the context of BED, this framework determines optimal designs for physical parameters $\boldsymbol{\theta}_\mathcal{G}$ and error parameters $\boldsymbol{\theta}_\mathcal{E}$, respectively. The purpose of separating these two types of parameters rather than jointly seeking designs and updating is to avoid the high-dimensional joint parameter space $\{\boldsymbol{\theta}_\mathcal{G}, \boldsymbol{\theta}_\mathcal{E}\}$, especially when using a neural network to calibrate structural model discrepancy. At each stage in the sequential process, for relatively low-dimensional physical parameters, we retain standard BED methods to allow arbitrary distributions. For high-dimensional error parameters, we (1) assume Gaussian distributions and follow the automatic differential ensemble Kalman inversion (AD-EKI) framework to find designs, (2) use a gradient-based method to update their deterministic values. The two steps are repeated within each stage: the updated model improves belief updating at later stages, and the refined belief in turn enables more effective model correction. A flowchart can be found in Algorithm \ref{alg:bed_workflow}.

At each stage of sequential BED, the optimal design for physical parameters is determined using standard BED: 
\begin{equation}
\label{eq:BED_optimization_G}
    \mathbf{d}^*_\mathcal{G} = \argmax_{\mathbf{d}\in\mathcal{D}} \mathbb{E}[U(p(\boldsymbol{\theta}_\mathcal{G}|\mathbf{y};\boldsymbol{\theta}_\mathcal{E}) || p(\boldsymbol{\theta}_\mathcal{G})  )],
\end{equation}
where the posterior distribution of $\boldsymbol{\theta}_\mathcal{G}$ from the previous stage serves as the prior distribution $p(\boldsymbol{\theta}_\mathcal{G})$ for the current stage. The utility function $U$ could either be the KL divergence or the Wasserstein distance. Considering that $\boldsymbol{\theta}_\mathcal{G}$ is often low-dimensional, standard BED methods can be employed to solve the optimization problem in Eq.~\eqref{eq:BED_optimization_G}. With the optimal design $\mathbf{d}^*_\mathcal{G}$, the corresponding data $\mathbf{y}_\mathcal{G}$ is then used to update the belief of physical parameters by the Bayesian theorem conditioned on the current error parameters $\boldsymbol\theta_\mathcal{E}$:
\begin{equation}
    p(\boldsymbol\theta_\mathcal{G}|\mathbf{y}_\mathcal{G},\mathbf{d}^*_\mathcal{G};\boldsymbol\theta_\mathcal{E})=\frac{p(\mathbf{y}_\mathcal{G}|\boldsymbol\theta_\mathcal{G},\mathbf{d}^*_\mathcal{G};\boldsymbol\theta_\mathcal{E})p(\boldsymbol\theta_\mathcal{G})  }{p(\mathbf{y}_\mathcal{G}|\mathbf{d}^*_\mathcal{G};\boldsymbol\theta_\mathcal{E})},
\end{equation}
which also provides the MAP estimation of physical parameters $\boldsymbol\theta_\mathcal{G}^*$:
\begin{equation}
    \label{eq:selet 1 theta}\boldsymbol\theta_\mathcal{G}^*=\argmax_{\boldsymbol\theta_\mathcal{G}}
 \{ p(\boldsymbol\theta_\mathcal{G}|\mathbf{y}_\mathcal{G},\mathbf{d}^*_\mathcal{G};\boldsymbol\theta_\mathcal{E})\}.
\end{equation}

On the other hand, the optimal design for error parameters is found by BED with the ensemble-based utility function:
\begin{equation}
    \begin{aligned}
        \mathbf{d}^*_\mathcal{E} &= \argmax_{\mathbf{d}\in\mathcal{D}} \mathbb{E}[U(p(\boldsymbol{\theta}_\mathcal{E}|\mathbf{y};\boldsymbol{\theta}_\mathcal{G}^*) || p(\boldsymbol{\theta}_\mathcal{E})  )],\\
    &=\argmax_{\mathbf{d}\in\mathcal{D}} \mathbb{E}[U(\{\tilde{\boldsymbol\theta}_\mathcal{E}^{(j)}\}_{j=1}^J || \{\boldsymbol\theta_\mathcal{E}^{(j)}\}_{j=1}^J )]
    \end{aligned}
\end{equation}
where $\{\tilde{\boldsymbol\theta}_\mathcal{E}^{(j)}\}_{j=1}^J$ and $\{\boldsymbol\theta_\mathcal{E}^{(j)}\}_{j=1}^J$ represent the updated and initial ensemble in EKI, respectively. The metric choice for $U$ aligns with that in Eq.~\eqref{eq:BED_optimization_G}. No matter what kind of metric is chosen, the AD-EKI introduced in \cite{yang2025bayesian} allows direct acquisition of the gradient of expected utility to design. At each stage of the sequential BED, we assume a Gaussian prior over the error parameters centered at their current value to enable ensemble update for EKI. Once the design $\mathbf{d}^*_\mathcal{E}$ is determined, the resulting data $\mathbf{y}_\mathcal{E}$ is then used to update the error parameters via standard gradient-based optimization:
\begin{equation}
    \boldsymbol\theta^*_\mathcal{E}=\argmax_{\boldsymbol\theta_\mathcal{E}} p(\mathbf{y}_\mathcal{E}|\boldsymbol\theta_\mathcal{G}^*,\mathbf{d}^*_\mathcal{E},\boldsymbol\theta_\mathcal{E}).
\end{equation}

At each stage, we immediately use the updated model ($\boldsymbol\theta^*_\mathcal{E}$) and the previously collected data for physical parameters, $\mathbf{y}_\mathcal{G}$, to refine the belief of the physical parameter $\boldsymbol{\theta}_\mathcal{G}$. Unlike the standard sequential BED approach, which propagates the posterior from the previous stage as the new prior, we adopt a re-update strategy that incorporates all accumulated data into the posterior update from the initial prior distribution. Specifically, the physical parameter posterior is recomputed as:
\begin{equation}
p(\boldsymbol\theta_\mathcal{G}|\mathbf{Y}_\mathcal{G},\mathbf{D}_\mathcal{G};\boldsymbol\theta^*_\mathcal{E})=\frac{p(\mathbf{Y}_\mathcal{G}|\boldsymbol\theta_\mathcal{G},\mathbf{D}_\mathcal{G};\boldsymbol\theta^*_\mathcal{E})p(\boldsymbol\theta_\mathcal{G})  }{p(\mathbf{Y}_\mathcal{G}|\mathbf{D}_\mathcal{G};\boldsymbol\theta^*_\mathcal{E})},
\end{equation}
where $\mathbf{D}_\mathcal{G}=\{\mathbf{d}_\mathcal{G}^n,\mathbf{d}_\mathcal{G}^{n-1},\cdots,\mathbf{d}_\mathcal{G}^1\}$ and $\mathbf{Y}_\mathcal{G}=\{\mathbf{y}_\mathcal{G}^n,\mathbf{y}_\mathcal{G}^{n-1},\cdots,\mathbf{y}_\mathcal{G}^1\}$ denote the sequence of historical designs and corresponding observations, and $p(\boldsymbol\theta_\mathcal{G})$ is the original prior, which may be uniform.

This re-update strategy helps correct the accumulating effect of early-stage model errors. When the model is not yet well corrected in early stages, posteriors updated via incorrect model may inherit inaccuracies from the model discrepancy. In sequential setting, by re-initializing the inference of $\boldsymbol\theta_\mathcal{G}$ from the original prior and incorporating all accumulated data with an improved model, we effectively remove the influence of previously biased posteriors and instead rely more directly on the observed data and the current, better-corrected model.

Notice that the alternating re-update and model-correction steps require no additional experiments or data acquisition. They are performed using accumulated data, solely by post computation. When computational cost is small relative to measurement cost, the procedure can be repeated multiple times. Conceptually, this alternating post-process scheme can progressively improve the posterior belief over physical parameters and the corrected model, and may converge to a stationary point. The attained solution is not necessarily globally optimal: its accuracy is constrained by the measurement noise in the available data and by limitations of the alternating-update framework. Formal convergence criteria and dedicated denoising or regularization strategies remain topics for future work.

\begin{algorithm}[H]
\caption{Workflow of the BED procedure used in this study. We compare different utility choices, including KL divergence, Wasserstein-1, and Wasserstein-2.}
\label{alg:bed_workflow}
\small
\begin{algorithmic}[1]
\Require Original prior $p(\boldsymbol{\theta}_\mathcal{G})$, initial error parameter $\boldsymbol{\theta}_\mathcal{E}^0$, utility $U\in\{D_{\mathrm{KL}},W_1,W_2\}$, and empty histories $\mathbf D_\mathcal{G}$ and $\mathbf Y_\mathcal{G}$.
\For{BED stage $n=1,\ldots,N$}
    \State Select the physical-parameter design $\mathbf d_\mathcal{G}^{n}$ under the current model with $\boldsymbol{\theta}_\mathcal{E}^{n-1}$.
    \State Obtain the corresponding observation $\mathbf y_\mathcal{G}^{n}$ and append $(\mathbf d_\mathcal{G}^{n},\mathbf y_\mathcal{G}^{n})$ to $(\mathbf D_\mathcal{G},\mathbf Y_\mathcal{G})$.
    \State Update the posterior belief $p(\boldsymbol{\theta}_\mathcal{G}\mid \mathbf y_\mathcal{G}^{n},\mathbf d_\mathcal{G}^{n};\boldsymbol{\theta}_\mathcal{E}^{n-1})$ by Bayesian inference.
    \If{model discrepancy is not considered}
        \State Use the updated posterior as the prior for the next BED stage.
    \Else
        \State Compute the MAP estimate $\boldsymbol{\theta}_\mathcal{G}^{*,n}$ from the current posterior.
        \State Select the error parameter design $\mathbf d_\mathcal{E}^{n}$ using AD-EKI conditioned on $\boldsymbol{\theta}_\mathcal{G}^{*,n}$ and the same utility $U$.
        \State Obtain the corresponding observation $\mathbf y_\mathcal{E}^{n}$.
        \State Update the error parameter by gradient-based optimization
        \State Recompute the physical-parameter posterior from the original prior $p(\boldsymbol{\theta}_\mathcal{G})$ using all accumulated physical data pairs $(\mathbf D_\mathcal{G},\mathbf Y_\mathcal{G})$ and the updated model.
        \State Use the re-updated posterior as the prior for the next BED stage.
    \EndIf
\EndFor
\end{algorithmic}
\end{algorithm}

\section{Numerical Results}
\label{sec: Numerical Results}

In this work, we mainly consider a classical source inversion problem in BED community~\cite{huan2014gradient, wu2023large, helin2025bayesian, yang2025bayesian}. This problem assumes a plume source located at an unknown position in a two-dimensional spatial domain, governed by certain PDE (i.e., combinations of convection, diffusion, and reaction). For each experimental design, a scalar concentration value can be measured at a certain location, and the goal is to infer the probability distribution of the source location based on the measurement, with a series of optimally designed measurement locations. Sometimes other properties of the system, e.g., source strength~\cite{shen2023bayesian} and convective velocity~\cite{YANG2025118198},  are also assumed to be unknown and left for inference. 

In this work, we aim to investigate the BED performance based on the expected Wasserstein metric and compare it with the classical expected KL divergence. We first study the performance comparison in the setting without model discrepancy and then extend to a parametric model-discrepancy setting.

The key motivations and conclusions of the numerical results are summarized below:
\begin{itemize}
\item We first study a single-stage two-dimensional toy source inversion example to demonstrate that Wasserstein distances can produce false increases in reward when the posterior mass is shifted toward the boundary of the prior support. This location-driven effect arises from transport geometry rather than information gain, and can mislead the design choice.
\item We then compare KL divergence and Wasserstein distances using a convection–diffusion plume inversion example without model discrepancy under a sequential BED setting. With actual utilities, KL leads to posteriors that converge more rapidly to the true source location, while Wasserstein metrics can be misled by location bias with a non-informative prior and also yield more uncertainties in posteriors at subsequent stages of sequential BED. With expected utilities, the performance gap narrows because the expectation introduces additional uncertainty to both KL- and Wasserstein-based utilities. In this setting, KL still yields superior designs in most cases. 
\item We further study the performance comparison under model discrepancy. Interestingly, the smaller uncertainties in the posteriors produced by the KL-based utility become a disadvantage. With actual utilities, KL starts to lead biased posteriors and thus slows down the model correction, while results via Wasserstein-based utility retain more uncertainties in posteriors and thus avoid a rapid convergence to biased estimation of source location and incorrect model discrepancy. With expected utilities, the performance gap also narrows. In this setting, Wasserstein yields superior designs in most cases by retaining greater probability mass near the true value and provides a faster correction of model discrepancy. 

\end{itemize}

An additional PDE example based on the acoustic inverse problem with and without model discrepancy can be found in~\ref{apd: Numerical case: acoustic inverse}.

\subsection{A toy example of 2-D inverse problem}\label{Sec: toy example}

We first consider a simplified but general two-dimensional toy example in a single-step design to illustrate the false reward effect of Wasserstein distance in BED. This example aims to show how the Wasserstein distance between posterior and prior can misleadingly increase with certain design choices, even when information gain does not actually improve.

\subsubsection{Setup}

We begin with a simplified toy example formulated within the source inversion framework. This example involves two key modeling assumptions. First, rather than specifying a single governing equation, we consider a broad class of systems in which the observation is represented as a function of the relative position between the observation point and the source. Specifically, the observation model takes the form
\begin{equation}
    y=f(\mathbf{z}-\boldsymbol\theta),
\end{equation}
where \( \mathbf{z} \in \mathbb{R}^2 \) denotes the sensor(observation) location, \( \boldsymbol\theta \in \mathbb{R}^2 \) is the unknown source location, and \( y \in \mathbb{R} \) is the observed scalar concentration. The function \( f \) is assumed to be isotropic and monotonically decreasing with respect to the Euclidean distance \( \|\mathbf{z} - \boldsymbol\theta\| \). This assumption defines a family of models that includes many commonly used physical processes with a decaying source, such as diffusive or dispersive transport. For example, admissible forms of \( f \) include
\begin{equation}
    y = -\|\mathbf{z} - \boldsymbol\theta\| + B \quad \text{or} \quad y = \exp(-\|\mathbf{z} - \boldsymbol\theta\|^2),
\end{equation}
both of which are radially symmetric and consistent with the physical intuition that concentration decreases with increasing distance from the source.

Second, we assume that the measurement is noise-free, which makes this a toy example. As a consequence, the likelihood function becomes a Dirac delta distribution: for any candidate parameter \( \boldsymbol\theta \), the likelihood equals 1 if the model prediction exactly matches the observation, and $0$ otherwise. Under this assumption, a single measurement induces a posterior distribution over the parameter space in the shape of a circular ring. The ring is centered at the measurement location \( \mathbf{z} \), and the true source \( \boldsymbol\theta \) lies on its circumference. Due to isotropic assumption, the posterior has zero thickness and behaves like a delta distribution over the circle in \( \mathbb{R}^2 \).

For numerical testing, we define the spatial domain as the square region \( [0, 3]^2 \), which bounds both the sensor deployment region and the parameter space. The true source location is set at \( \boldsymbol\theta = \{0.9, 1.2\} \), and the prior distribution is assumed to be uniform over the same domain, representing non-informative beliefs about the source position.

We select several candidate sensor locations and calculate the corresponding posteriors and their KL divergence / Wasserstein distance against the prior. Specifically, these sensors are located along the straight line connecting the source location and the lower-left corner of the domain. Our goal is to analyze how a single measurement affects the posterior distribution and to identify which candidate measurement location provides the greatest actual information gain. 

\subsubsection{Results}

\begin{figure}[H]
  \centering
  \begin{subfigure}[t]{0.48\textwidth}
    \centering
    \includegraphics[width=0.85\linewidth]{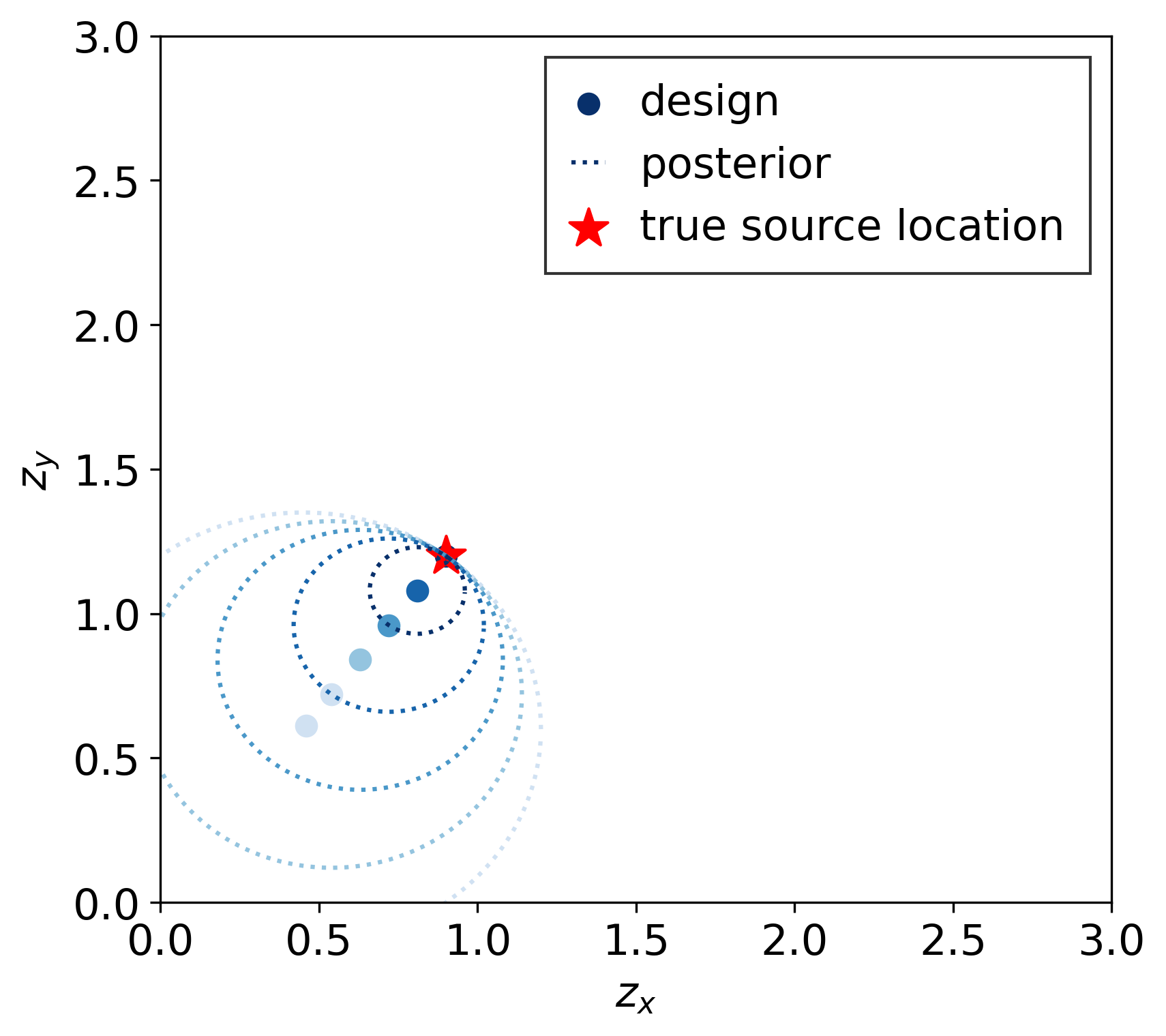}
    \caption{Design and posterior}
    \label{fig:fig1a}
  \end{subfigure}
  \hfill
  \begin{subfigure}[t]{0.48\textwidth}
    \centering
    \includegraphics[width=\linewidth]{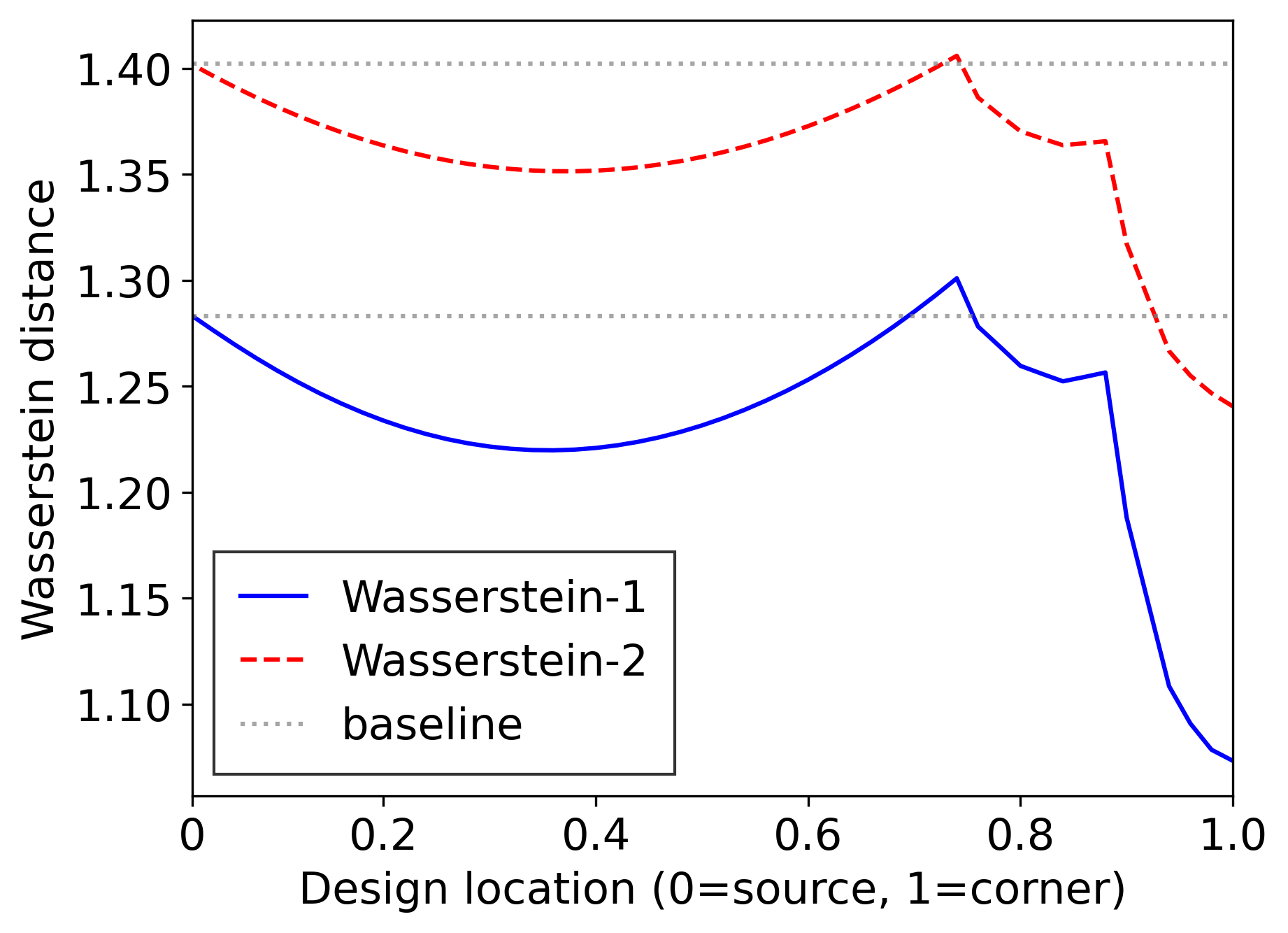}
    \caption{Reward vs. design}
    \label{fig:fig1b}
  \end{subfigure}
  \caption{Illustration of design and associated reward. Panel (a) Design and posterior: each filled circle denotes a design point, and the dotted circle of the same color indicates its corresponding posterior distribution. The red star marks the true source location. Panel (b) Reward vs. design: Wasserstein-1 and Wasserstein-2 distances are plotted as functions of the design location, normalized so that $0$ corresponds to the source location and $1$ to the domain corner. The grey dotted line shows the baseline. }
  \label{fig:fig1}
\end{figure}

Figure~\ref{fig:fig1a} illustrates the overall setup. As the design point moves further from the source location toward the boundary of the domain, the posterior circles grow in radius, eventually touching or exceeding the boundary. If the posterior mass extends beyond the parameter support, it would be truncated in standard Bayesian computations. 

Figure~\ref{fig:fig1b} shows how the Wasserstein distance between posterior and prior varies with the design location. When the design is within approximately 40\% of the diagonal distance (along its way from source to corner), the Wasserstein distance decreases. This agrees with intuition: a posterior highly concentrated around the true source (a delta distribution) should be more informative than a circular posterior with probability spread in multiple directions. Beyond this range, however, the Wasserstein distance begins to increase. The rise is not due to reduced uncertainty, but rather to transport geometry: moving prior mass from the far upper-right region of the support to a posterior increasingly concentrated near the lower-left incurs a large relocation cost. We refer to this behavior as a false increase in reward, since it exaggerates the apparent informativeness of the design without reflecting actual information gain. Notably, around 70\% along the diagonal, the Wasserstein distance associated with a circular posterior can exceed that of the delta posterior centered at the truth, incorrectly suggesting that such designs are more informative. As the design approaches the boundary, the posterior is truncated by the support and the Wasserstein distance decreases again, creating a local maximum. Such a maximum of information gain indicates that if Wasserstein distance were used directly as an information metric in BED, this would incorrectly suggest that these designs yield greater information, even though they actually produce misleading posteriors.

This example highlights the potential limitation of using Wasserstein distance as information gain: its value depends not only on posterior uncertainty but also on the relative spatial position of posterior mass with respect to the prior support. Such spatial effects can therefore mislead design selection. A simple mitigation is to enlarge the support domain used to define the prior, which effectively shifts the true parameter away from the boundary and reduces the chance of encountering false increases, at the cost of increasing computational burden.

\subsection{An inverse problem of convection-diffusion equation without model discrepancy}

\subsubsection{Setup}

To compare the KL divergence and Wasserstein distance in a more realistic case, we study the source inversion problem of a contaminant in the convection-diffusion field, which is a classical test example in BED and has been previously studied in \cite{shen2023bayesian, YANG2025118198}. More specifically, the contaminant concentration $\mathbf{u}$ at two-dimensional spatial location $\mathbf{z} = \{z_x, z_y\}$ and time $t$ is governed by the following equation:

\begin{equation}
    \frac{\partial \mathbf{u}(\mathbf{z},t;\boldsymbol\theta)}{\partial t}=D\nabla^2\mathbf{u}-\mathbf{v}(t) \cdot \nabla \mathbf{u}+S(\mathbf{z},t;\boldsymbol\theta),~~~\mathbf{z} \in [z_L,z_R]^2,~~t>0
    \label{eq:true_system_example},
\end{equation}
where $D$ is the diffusion coefficient assumed to be known with a true value of $1$, $\mathbf{v}=\{v_x,v_y\} \subseteq \mathbb{R}^2$ is a time-dependent convection velocity assumed to be known with true value of $v_x=v_y=50t$, $S$ denotes the source term with some parameters $\boldsymbol\theta$. In this work, the true system has an exponentially decaying source term in the following form with the parameters $\boldsymbol\theta=\{\theta_x,\theta_y,\theta_h,\theta_s\} \in \mathbb{R}^4$:
\begin{equation}
    S(\mathbf{z},t;\boldsymbol\theta)=\frac{\theta_s}{2\pi\theta_h^2}\exp \left(-\frac{(\theta_x-z_x)^2+(\theta_y-z_y)^2}{2\theta_h^2} \right),
    \label{eq:true_source_term}
\end{equation}
where \(\theta_x\) and \(\theta_y\) denote the source location, \(\theta_h\) and \(\theta_s\) represent the source width and source strength. The initial condition is \(\mathbf{u}(\mathbf{z}, 0;\boldsymbol\theta) = \mathbf{0}\), and a homogeneous Neumann boundary condition is imposed on all sides of the square domain. 

Note that in this work, the inversion is performed over a finite-dimensional parameter set $\boldsymbol\theta$ which governs the source term (e.g., source location, width, and strength). The state variable $\mathbf{u}$ is not directly inferred but is obtained by solving the governing PDE given $\boldsymbol\theta$.

In all numerical examples, the physics-based unknown parameters $\boldsymbol{\theta}_\mathcal{G}$ are the source location $\{\theta_x,\theta_y\}$. The parameter space is set as $[0,1]^2$. The true value varies in different cases:
\begin{table}[H]
\centering
\begin{tabular}{|c|c|c|}
\hline
\textbf{Case} & $\boldsymbol{\theta_x}$ & $\boldsymbol{\theta_y}$ \\
\hline
1 \& 2 \& 4 & 0.3 & 0.4 \\
3          & 0.4 & 0.4 \\
5          & 0.1& 0.1\\
\hline
\end{tabular}
\caption{Values of $\theta_x$ and $\theta_y$ for different cases. Case 5 is presented in the appendix.}
\end{table}

The true values of the remaining source parameters are fixed at $\theta_s=2$ and $\theta_h=0.05$ for all cases. The PDE is numerically solved on $[-2,3]^2$, which is larger than the parameter space, to avoid any boundary effects. 

The design $\mathbf{d}=\{d_x,d_y\}$ in this problem refers to the spatial coordinate to measure the concentration value. More specifically, each design involves measuring just one point in the domain. The temporal coordinates of the measurement are set as $0.05+n\times0.005$ time units, where $n$ is the stage number ($n=1,2,\cdots$). The design searching area is set as $[0,1]^2$, which happens to be the same as the parameter space. Measurements contain a Gaussian noise $y=\mathbf{u}(\mathbf{d})+\epsilon, \quad \epsilon \sim \mathcal{N}(0,0.05^2)$.

\subsubsection{Case 1: Actual information gain}

We conduct a six-stage sequential BED using three utility functions: KL divergence, Wasserstein-1 ($W_1$), and Wasserstein-2 ($W_2$). In this section, designs are optimized with the actual utility computed from the observed data at each step, rather than the expected utility; results based on expected utility are presented in the next section. The purpose of using actual utility in this section is to clearly reveal how each metric influences the design selection process and the resulting posterior distributions. This choice is potentially reasonable in scenarios where (i) a small preliminary measurement budget allows one to obtain actual rewards at a handful of locations and then extrapolate to higher-reward designs for subsequent measurements, or (ii) multi-fidelity sensing is available, so that low-precision probes can identify promising design points before committing high-precision measurements.

\begin{figure}[H]
    \centering
    \includegraphics[width=\linewidth]{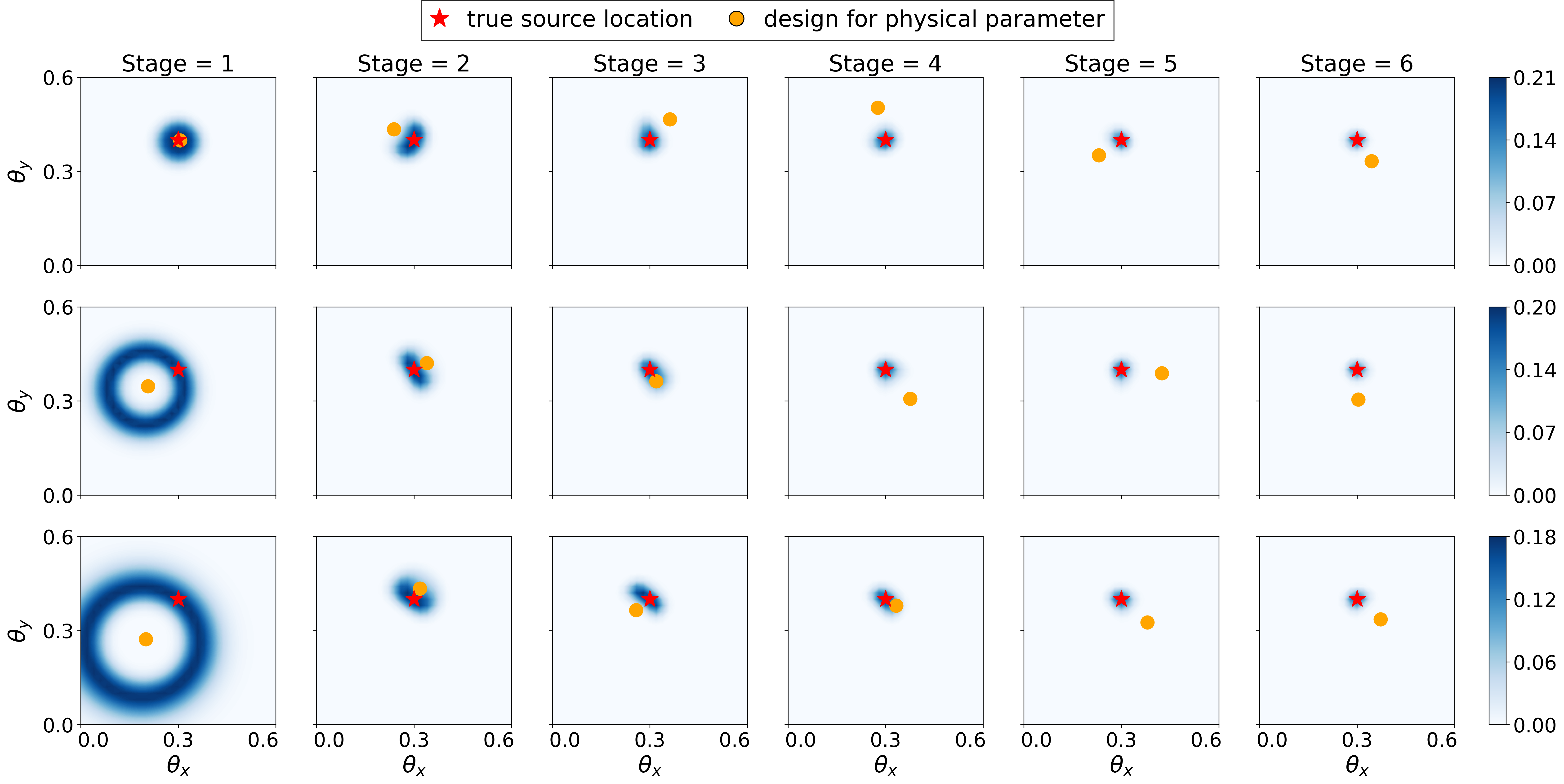}
    \caption{Without error (case 1), using actual information gain: posterior. From top to bottom: KL, $W_2$, $W_1$. The parameter space is set as $[0,1]^2$ but zoomed in as $[0,0.6]^2$. This holds for all the remaining posterior figures.}
    \label{fig: no error actual posterior}
\end{figure}

Figure~\ref{fig: no error actual posterior} presents the evolution of the posterior across six stages. After the first measurement, the design chosen under KL divergence yields a posterior that is tightly centered near the true parameter with low uncertainty. By contrast, the designs selected by $W_1$ and $W_2$ tend to lie closer to the boundary of the support; their posteriors still cover the truth but place substantial probability mass away from it. This reproduces the conclusion of the previous section: the location-driven false reward effect causes Wasserstein-based utilities to prefer suboptimal designs when used directly as information metrics. This bias degrades the performance of the Wasserstein-based design.

\begin{figure}[H]
  \centering
  \begin{subfigure}[t]{0.48\textwidth}
    \centering
    \includegraphics[width=0.8\linewidth]{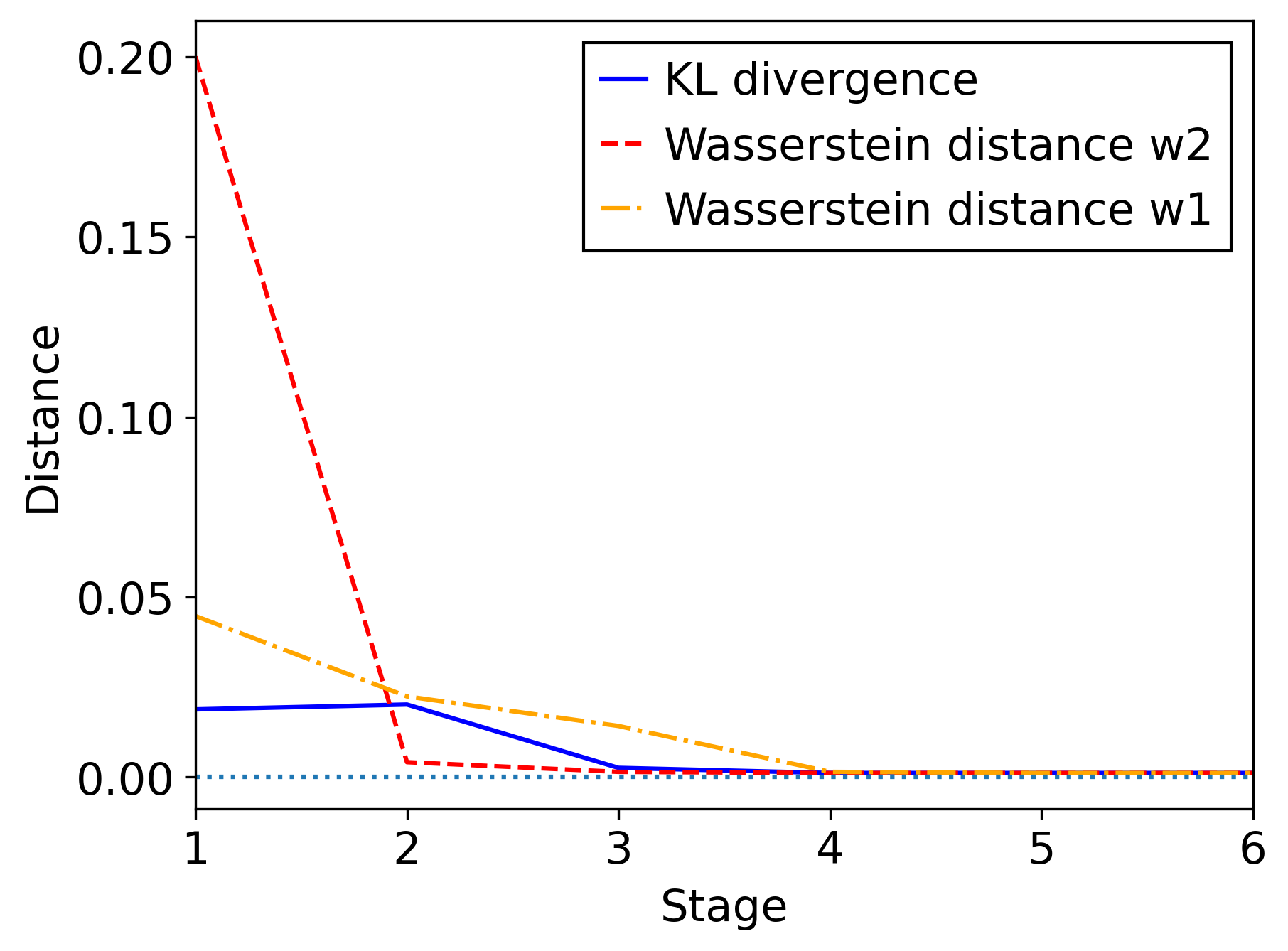}
    \caption{Distance}
    \label{fig: no error_actual distance}
  \end{subfigure}
  \hfill
  \begin{subfigure}[t]{0.48\textwidth}
    \centering
    \includegraphics[width=0.8\linewidth]{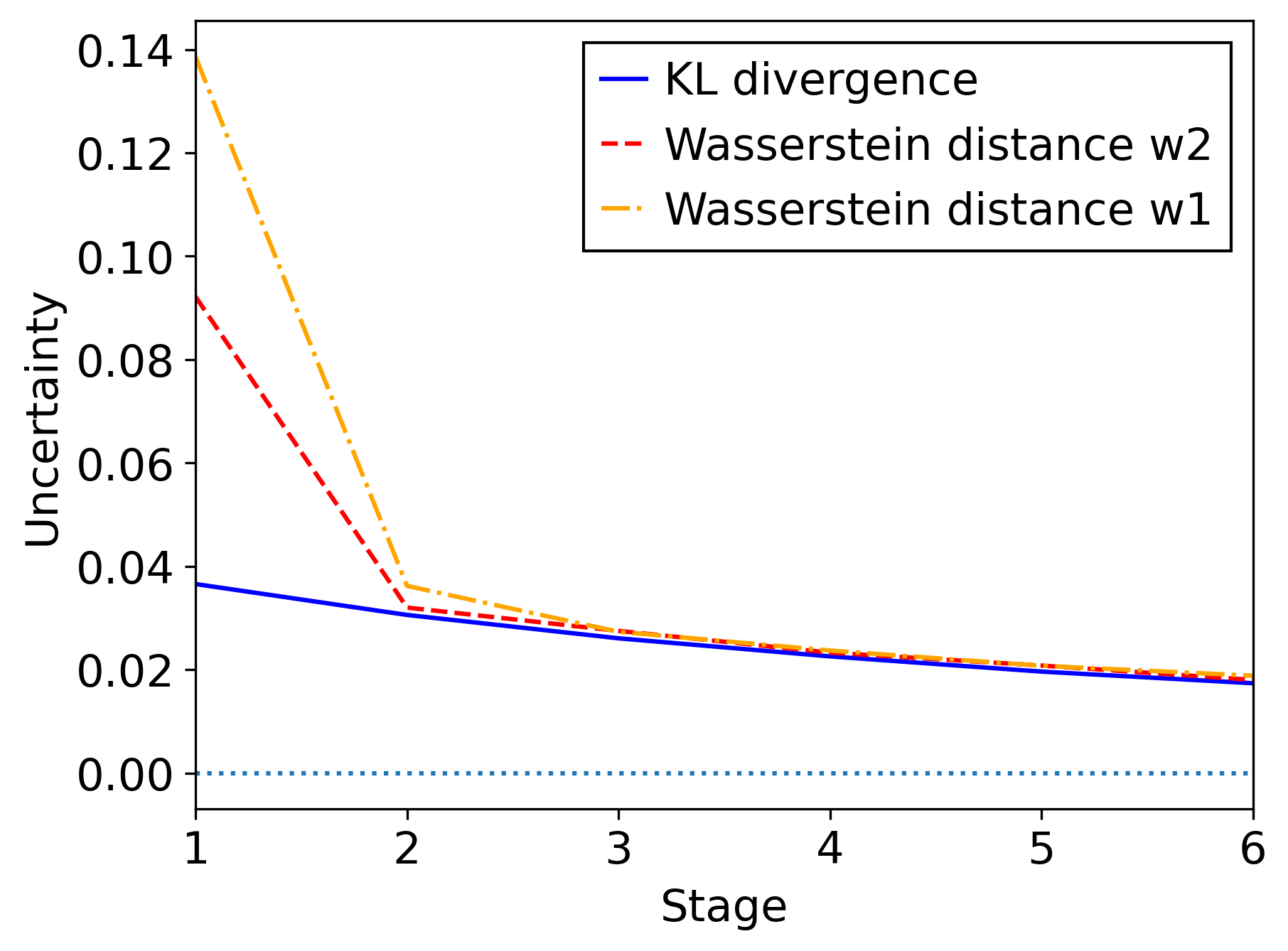}
    \caption{Uncertainty}
    \label{fig: no error expected uncertainty}
  \end{subfigure}
  \caption{Distance and uncertainty evolution for Case 1, where the designs are selected using actual information gain without model discrepancy.}
  \label{fig: no error_actual distance and uncertainty}
\end{figure}

We further look into some quantitative metrics to evaluate the posterior distribution of the source location $\{\theta_x, \theta_y\}$. We define the distance between the MAP point $\boldsymbol{\theta}_\mathcal{G}^*=\{\theta_x^*,\theta_y^*\}$ and the true value $\boldsymbol{\theta}_\mathcal{G}^\dagger=\{\theta_x^\dagger,\theta_y^\dagger\}$ in each stage:
\begin{equation}
\label{eq: distance}
    D = \|\boldsymbol{\theta}_\mathcal{G}^* - \boldsymbol{\theta}_\mathcal{G}^\dagger\|_2 =\sqrt{(\theta_x^*-\theta_x^\dagger)^2+(\theta_y^*-\theta_y^\dagger)^2},
\end{equation}
and the magnitude of uncertainty:
\begin{equation}
\label{eq: uncertainty}
    \sigma_{\text{eq}} = \sqrt[4]{\lambda_1 \lambda_2} .
\end{equation}
where $\lambda_1$ and $\lambda_2$ are the eigenvalues of the posterior covariance matrix. Corresponding results are presented in Fig.~\ref{fig: no error_actual distance and uncertainty}. The least uncertainty given by KL at stage 1 further confirms its better performance. The reason for the nearly the same uncertainty starting from stage 2 is that two data points are sufficient to constrain the two-dimensional unknown parameter when using actual information gain.

We also observe that the $W_1$- and $W_2$-selected designs differ slightly, as do the resulting posterior shapes. A plausible explanation is their different behavior to boundary-induced truncation. In the previous example in Section \ref{Sec: toy example}, the two metrics behaved similarly because the posteriors consisted of delta masses. By contrast, in this case, the posterior decays smoothly outside the circular shape, so truncation induces more intricate (and metric-dependent) behavior. 

Notice we are using a greedy strategy at each stage. The difference between the KL divergence and the Wasserstein distance in long-term or accumulated reward is left for future work. 

\subsubsection{Case 2: Expected information gain}

We now perform a six-stage sequential BED experiment using expected utility functions. An expected utility formulation corresponds to the practical scenario where no additional measurement budget is available, and design selection must rely entirely on the predictive capability of the current model. In this setting, candidate measurement outcomes are simulated from the forward model and used to compute the expected information gain. This substantially reduces the need for real measurements but increases computational cost. 

\begin{figure}[H]
  \centering
  \begin{subfigure}[t]{0.32\textwidth}
    \centering
    \includegraphics[width=\linewidth]{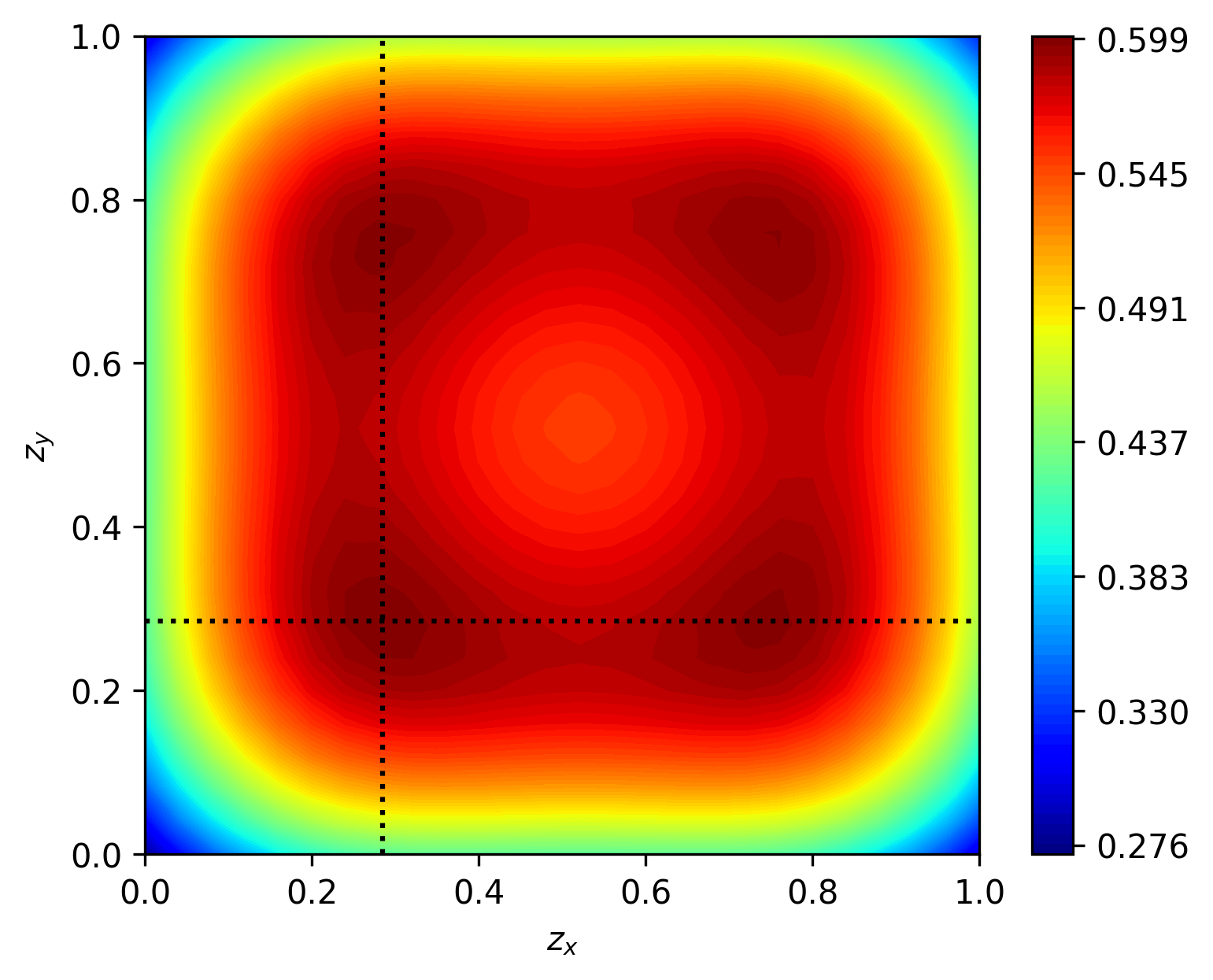}
    \caption{KL}
    \label{fig: no error reward KL}
  \end{subfigure}
  \begin{subfigure}[t]{0.32\textwidth}
    \centering
    \includegraphics[width=\linewidth]{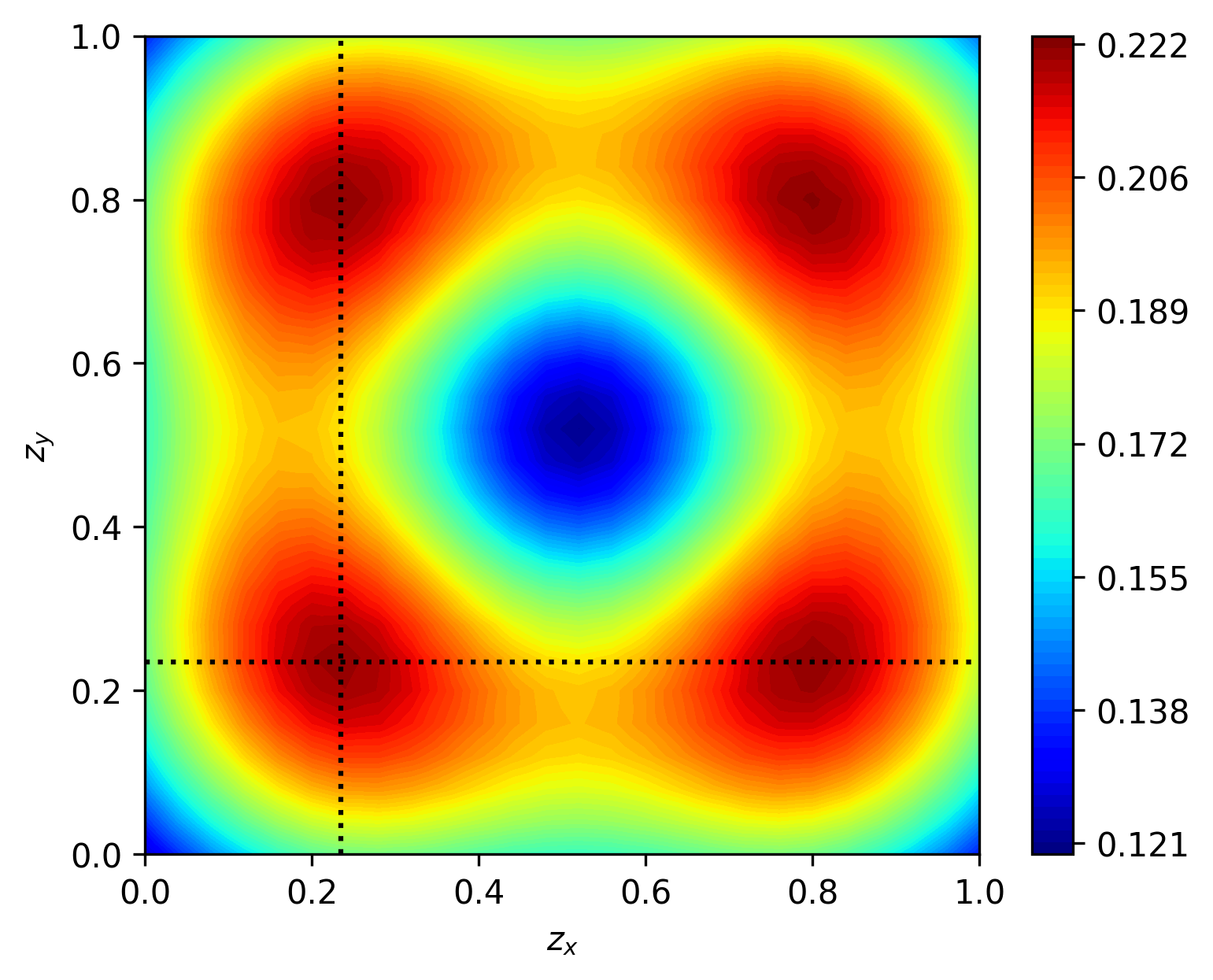}
    \caption{$W_2$}
    \label{fig: no error reward w2}
  \end{subfigure}
  \begin{subfigure}[t]{0.32\textwidth}
    \centering
    \includegraphics[width=\linewidth]{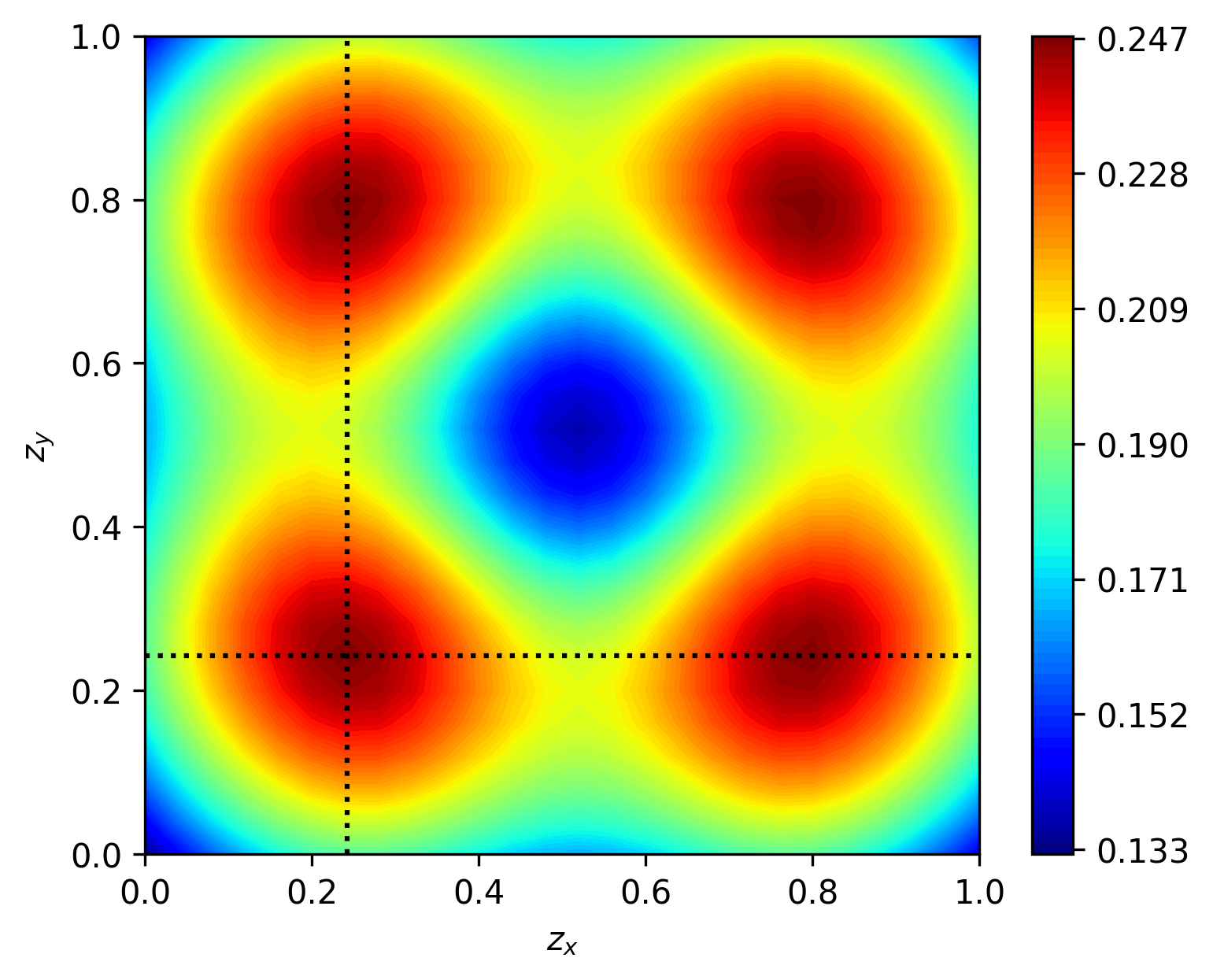}
    \caption{$W_1$}
    \label{fig: no error reward w1}
  \end{subfigure}
  \caption{Reward map with no model discrepancy at stage 1 for Case 2, where no discrepancy is present. The black dotted lines indicate the $z_x$ and $z_y$ coordinates of the global maximizer.}
  \label{fig: no error reward map}
\end{figure}

\begin{figure}[H]
    \centering
    \includegraphics[width=\linewidth]{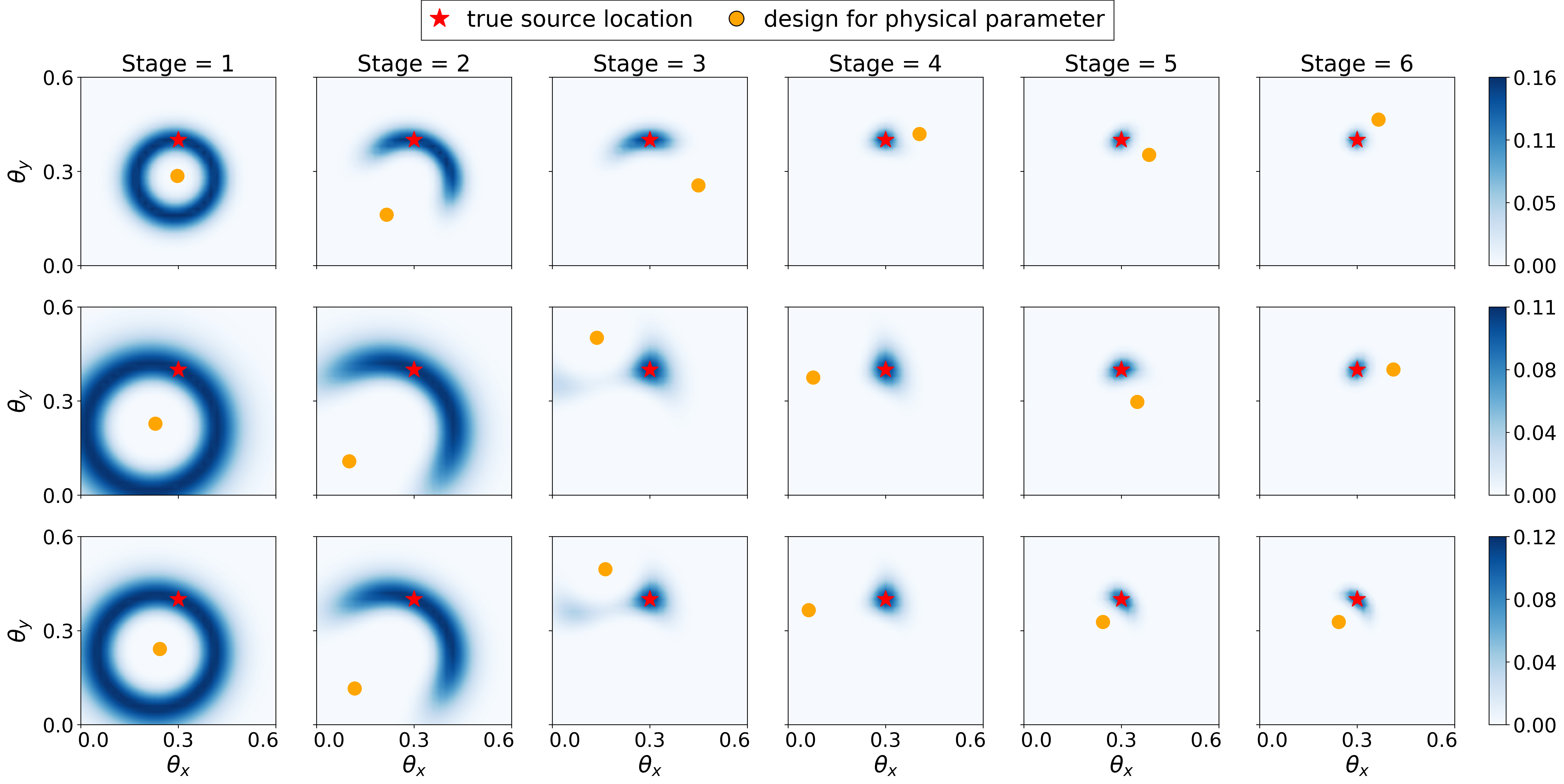}
    \caption{Without error (case 2), using expected information gain: posterior. From top to bottom: KL, $W_2$, $W_1$. }
    \label{fig: no error expected posterior}
\end{figure}

Figure~\ref{fig: no error reward map} presents the reward map of different expected utilities at stage 1, depicting the contour of expected utility values across various design points. The reward map of expected KL divergence (Fig.~\ref{fig: no error reward KL}) closely resembles Figure 10 in \cite{shen2023bayesian}, where a similar case setup was used, indicating the reliability of the numerical results. Overall, the reward map of the expected Wasserstein distance is largely similar to that of the expected KL divergence, indicating that the expected Wasserstein distance is also suitable as a BED utility function. However, the locations of the extrema exhibit certain differences. The design points that yield the highest expected Wasserstein distance are located near those of expected KL divergence, suggesting that Wasserstein distance is a potential candidate for utility function, similar to KL divergence. However, a slight difference remains: the optimal design point based on expected Wasserstein distance lies slightly closer to the boundary compared to that of expected KL divergence, as more clearly illustrated in the first stage of Fig.\ref{fig: no error expected posterior}. This reflects the same false reward effect discussed in the previous section.

\begin{figure}[H]
  \centering
  \begin{subfigure}[t]{0.48\textwidth}
    \centering
    \includegraphics[width=0.8\linewidth]{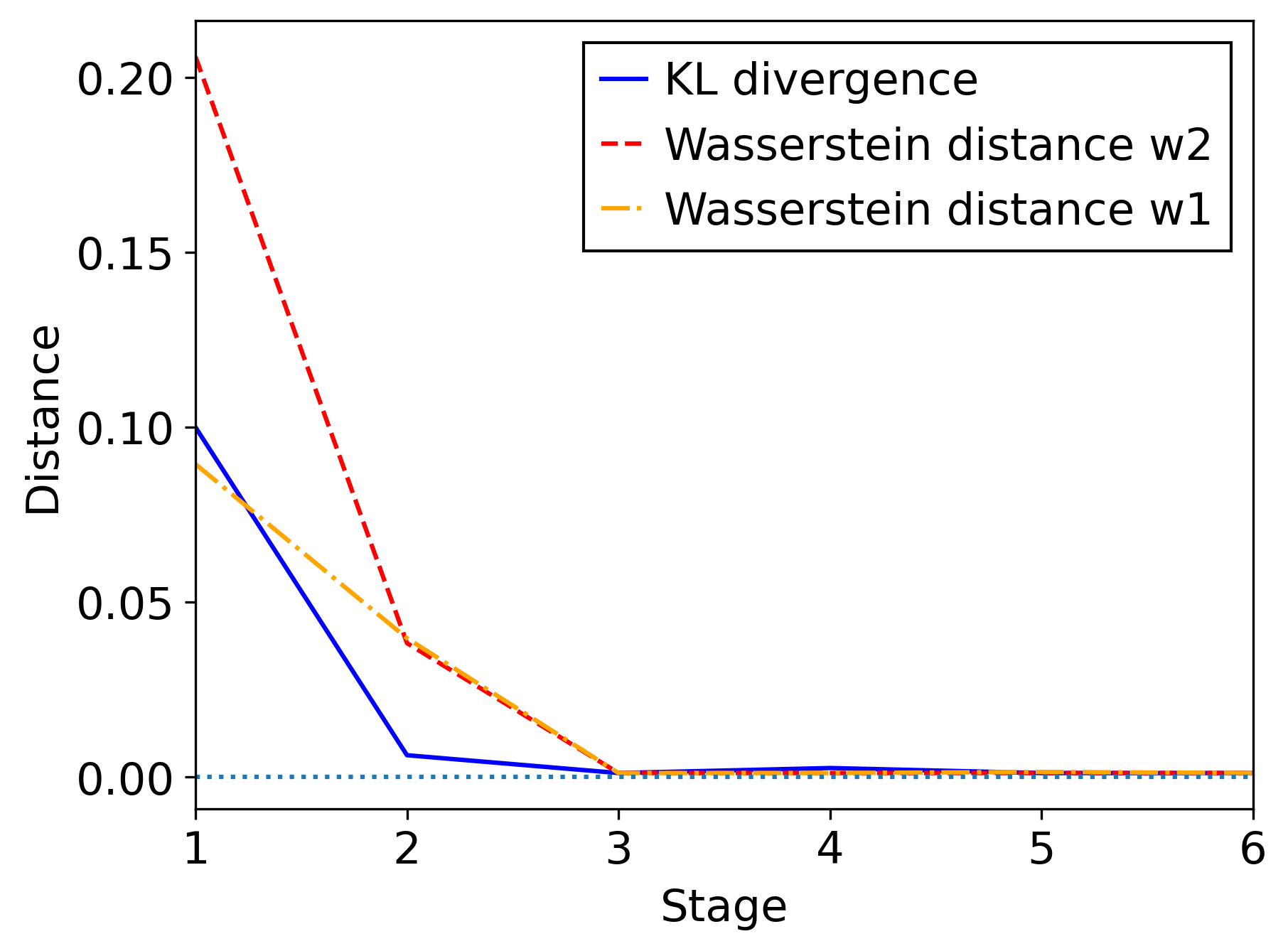}
    \caption{Distance}
    \label{fig: no error expected distance}
  \end{subfigure}
  \hfill
  \begin{subfigure}[t]{0.48\textwidth}
    \centering
    \includegraphics[width=0.8\linewidth]{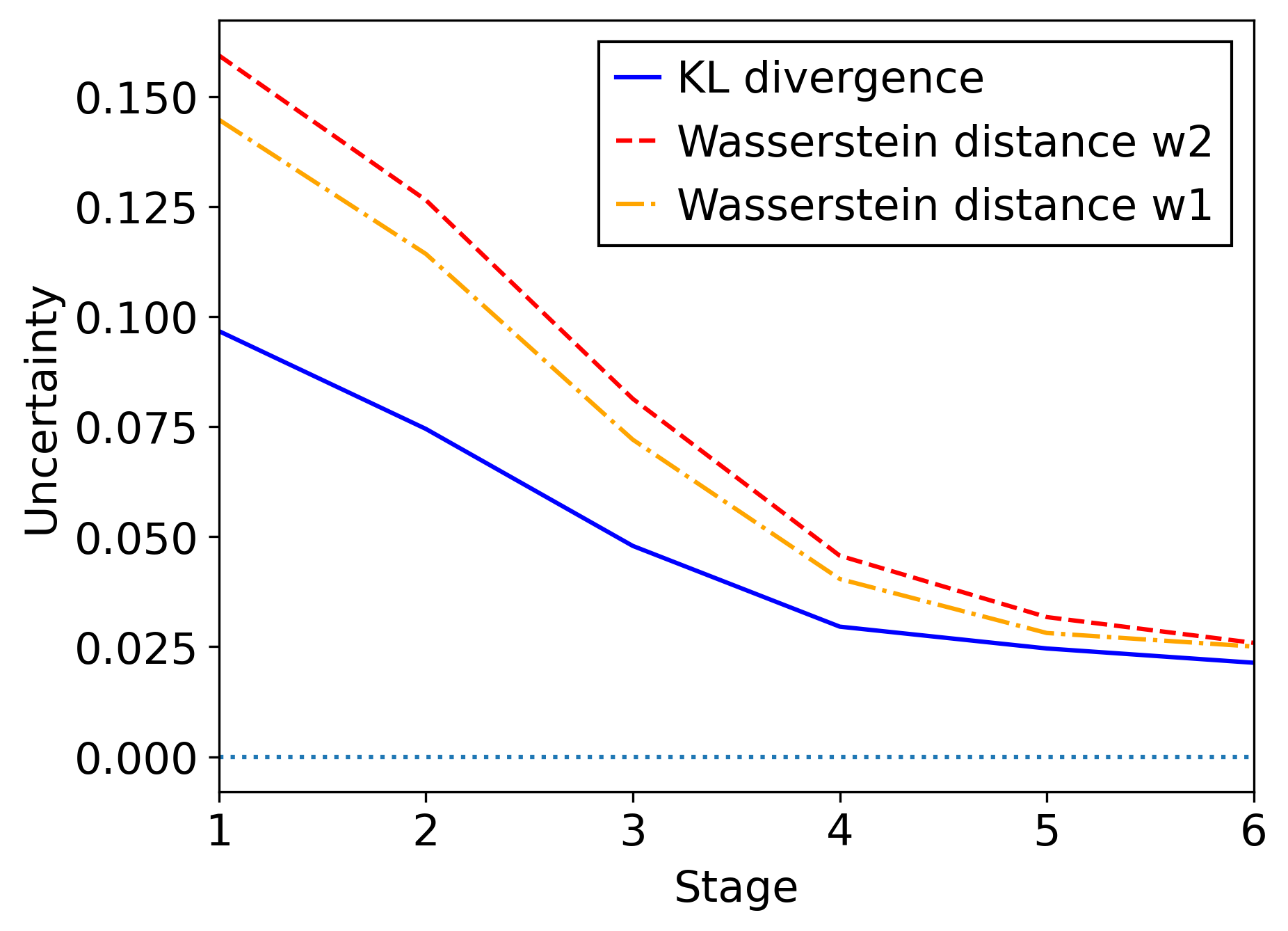}
    \caption{Uncertainty}
    \label{fig: no error expected uncertainty}
  \end{subfigure}
  \caption{Distance and uncertainty evolution for Case 2, where the designs are selected using expected information gain without model discrepancy.}
  \label{fig: no error expected distance and uncertainty}
\end{figure}

Figure~\ref{fig: no error expected posterior} shows the evolution of the posterior distributions over six stages. In stage~1, both $W_1$ and $W_2$ select designs that lie closer to the boundary of the domain compared to the KL divergence, resulting in more diffuse posteriors. Because these early-stage posteriors, which serve as priors for subsequent stages, retain substantial probability mass away from the true parameter, the following design choices become less effective at quickly eliminating uncertainty. As a result, the posteriors of $W_1$ and $W_2$ exhibit higher dispersion than those of KL even in the later stages. This inferior performance of Wasserstein distance is further confirmed by the uncertainty results in Fig.~\ref{fig: no error expected distance and uncertainty}. In other words, the uncertainty in the posterior induced by the Wasserstein-based design requires one additional measurement to converge to the same level as that induced by the KL-based design.

It is instructive to discuss the belief updating speed under actual information gain and under expected information gain. Empirically, in stage 1 the posteriors induced by expected-utility designs are less concentrated than those obtained with actual utility (see the first rows of Figs. \ref{fig: no error actual posterior} and \ref{fig: no error expected posterior}). The posterior mass becomes centered at the truth by stage 3, which requires one additional measurement relative to the actual-utility setting. The reason is that expected information gain is optimal only in the prior-averaged sense. The design that maximizes the expectation need not be the most informative if evaluating its true information gain, especially when the prior is weak. In other words, the expectation step injects additional uncertainty. This uncertainty propagates across stages because a more diffuse early-stage posterior becomes the prior for later stages, which slows contraction toward the truth.

This additional uncertainty is pivotal when comparing discrepancy measures. Under actual utility, KL divergence typically outperforms the Wasserstein distance: KL favors designs that minimize residual uncertainty and tends to select high-signal, noise-robust measurements, producing concentrated posteriors. By contrast, Wasserstein distances can be affected by a false-increase effect that shifts designs toward the boundary, leading to more diffuse posteriors. After switching to expected utility, the expectation-induced uncertainty weakens the absoluteness of this ranking. In most cases the expected KL still dominates the expected Wasserstein distance, yet regimes exist in which the Wasserstein-based designs yield lower uncertainty in early stages, owing to the interaction between weak priors and the stage-1 design choice (see \ref{appendix: reverse case}).

In summary, the false increase effect is still present under expected utilities, predominantly appearing in the early stages and subsequently slowing down the uncertainty reduction in the sequential process due to the residual posterior uncertainty.  Because expected utilities introduce additional uncertainty, the design bias of Wasserstein distances does not always translate into a large performance disadvantage. Nevertheless, in most cases, expected KL divergence still achieves better performance than expected Wasserstein distances.

\subsection{An inverse problem of convection-diffusion equation with model discrepancy}

In this section, we discuss a more challenging case. We assume the model we use is different from the true system, that is, under model discrepancy~\cite{feng_optimal_2015, YANG2025118198}. In this convection--diffusion example, we only consider the parametric error case as our focus is on comparing different metrics. We assume knowing an incorrect source strength parameter $\theta_s=3$, against the true value 2. Under this setting, it can be expected that normal inference and BED methods can not produce satisfying results. We follow the framework described in Section \ref{Sec: methodology framework} to find designs and iteratively update both the belief of the physical location parameter $\{\theta_x,\theta_y\}$ and the value of the error parameter $\theta_s$.

\subsubsection{Case 3: Actual information gain}\label{sec: actual error}

\begin{figure}
    \centering
    \includegraphics[width=\linewidth]{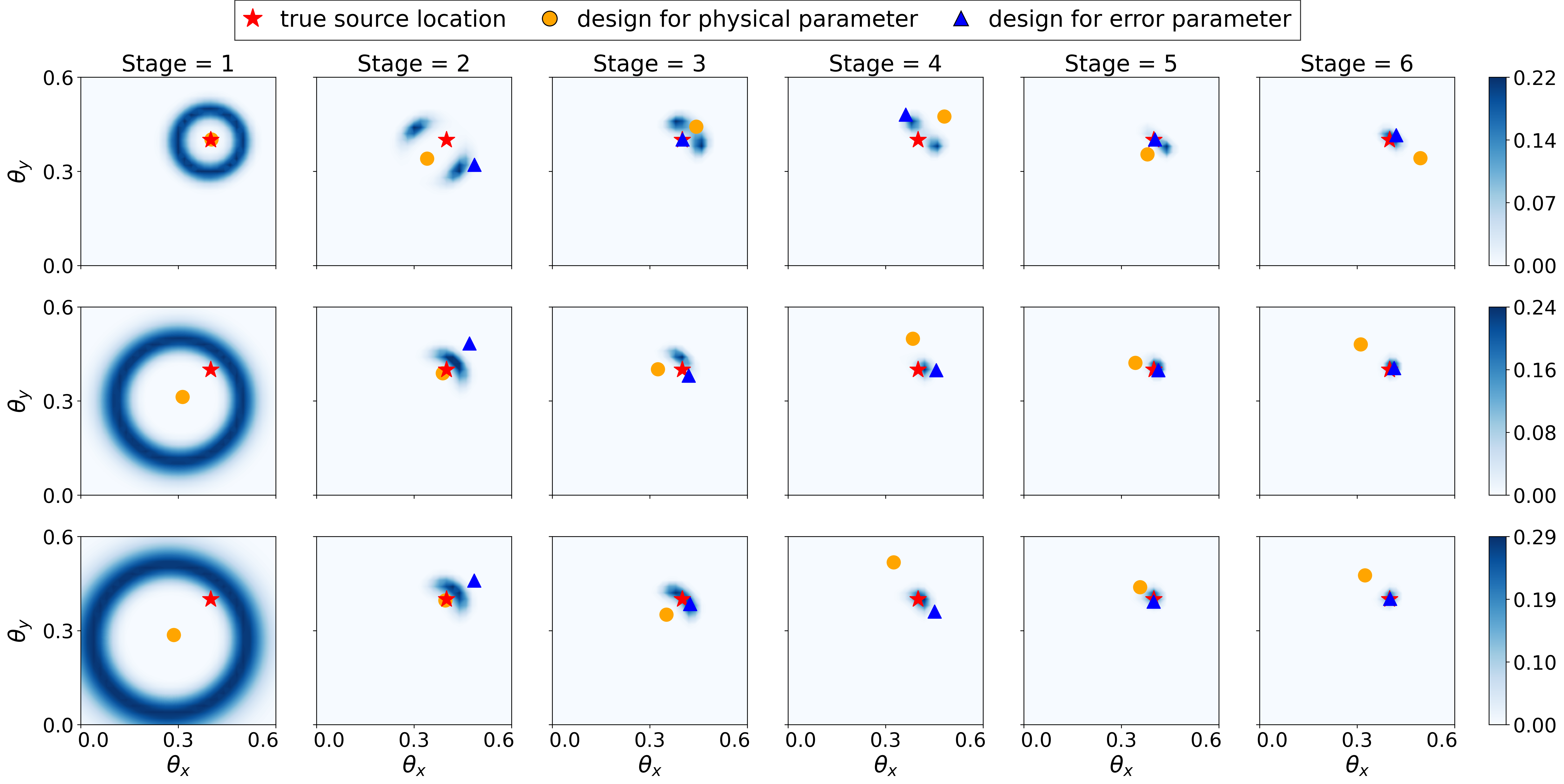}
    \caption{With error (case 3), using actual information gain: posterior. From top to bottom: KL, $W_2$, $W_1$. }
    \label{fig: error actual posterior}
\end{figure}

Figure~\ref{fig: error actual posterior} shows the evolution of the posterior distributions over six stages in the presence of model discrepancy. In stage~1, KL divergence still selects the design at the source location, producing the largest measurement value and the most concentrated posterior. However, because the forward model is incorrect, the posterior does not cover the true parameter but instead forms a ring around it. Despite being misaligned, this posterior is still more compact than those obtained from other measurement locations, which produce larger rings. Designs selected by the Wasserstein metrics are shifted slightly outward relative to KL, resulting in more diffuse posteriors.

\begin{figure}[H]
  \centering
  \begin{subfigure}[t]{0.48\textwidth}
    \centering
    \includegraphics[width=0.8\linewidth]{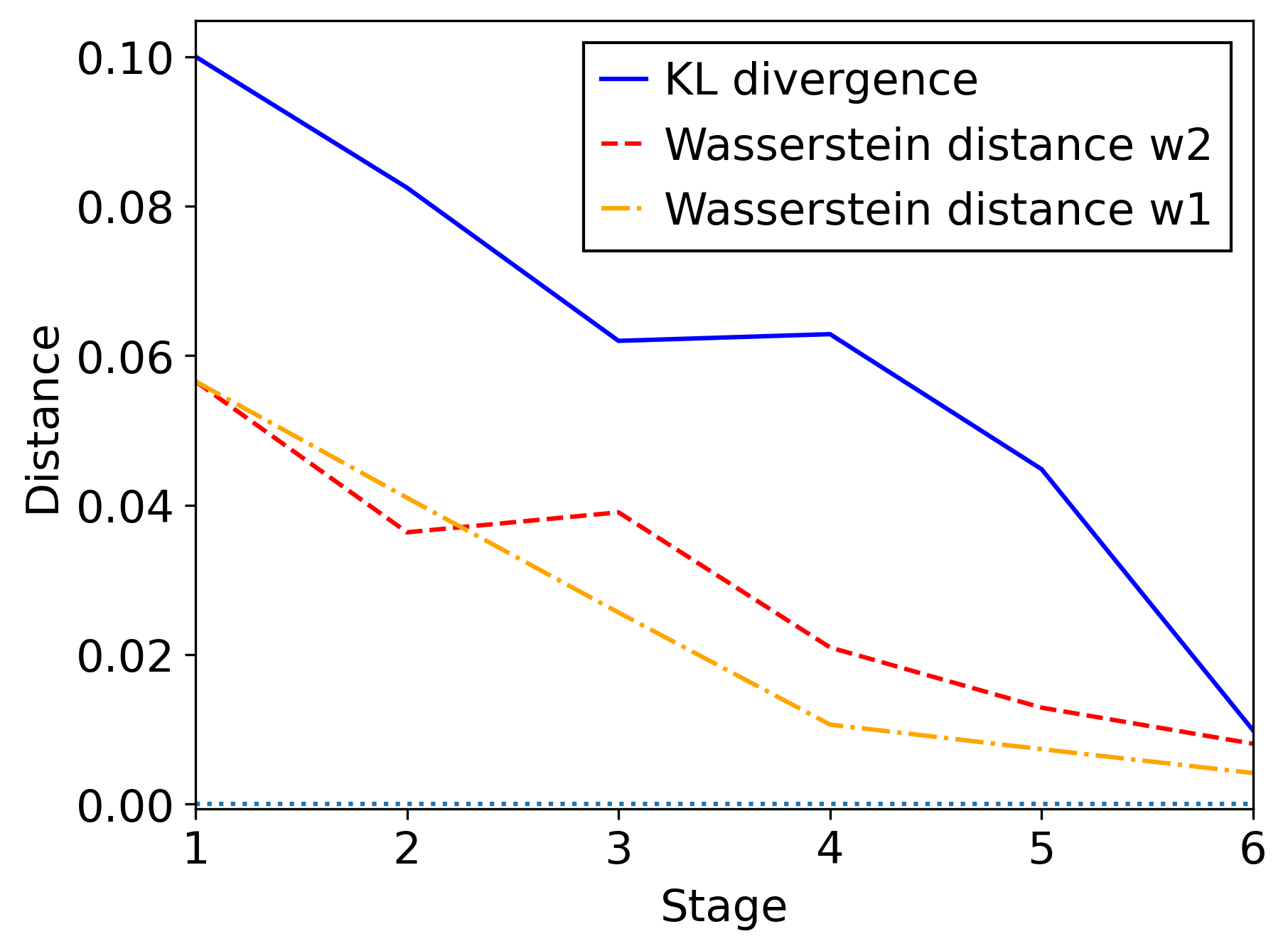}
    \caption{Distance}
    \label{fig: error actual distance}
  \end{subfigure}
  \hfill
  \begin{subfigure}[t]{0.48\textwidth}
    \centering
    \includegraphics[width=0.8\linewidth]{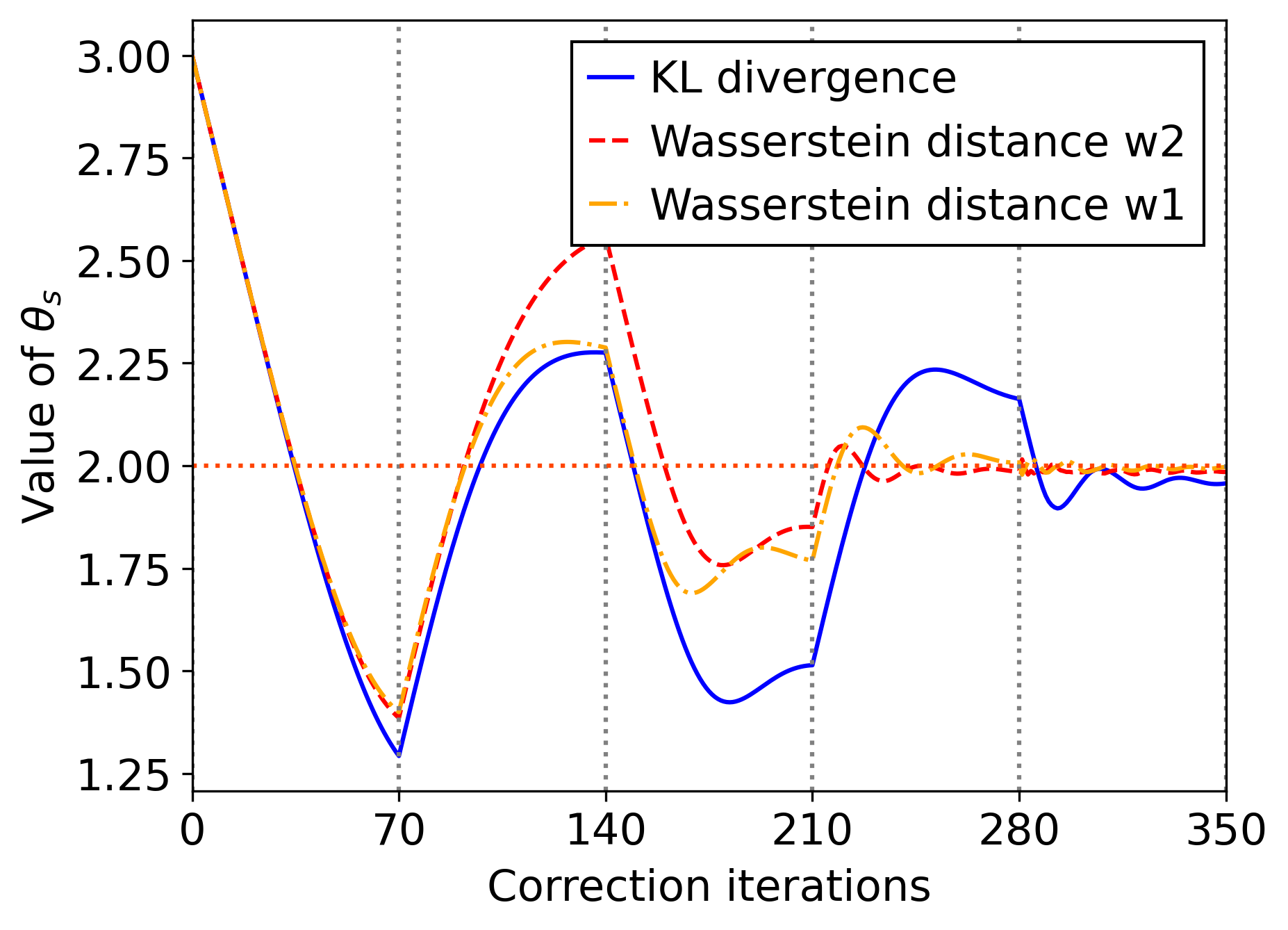}
    \caption{Error parameter $\theta_s$}
    \label{fig: error actual theta}
  \end{subfigure}
  \caption{Distance and parametric error evolution for Case 3, where designs are selected using actual information gain in the presence of model discrepancy.}
  \label{fig: error actual distance and theta}
\end{figure}

Interestingly, in this setting, the Wasserstein-based posterior places more weight on the true source location than the KL-based posterior. As a result, by stage~2, after two experiments, the posteriors obtained using the Wasserstein utilities are centered more closely to the true value than the posterior obtained using the KL utility. The reason can be traced to the stage~1 posteriors: KL’s posterior was overly concentrated in the wrong region due to the model error, leading its stage~2 uncorrected posterior to remain more peaked in areas overlapping its stage~1 prior. As a result, when the model is later corrected, the KL posterior requires more substantial probability mass transport to move toward the truth, while the Wasserstein posterior can be adjusted with less effort. In contrast, the more diffuse Wasserstein posterior spread probability more evenly over a wider area, making it less overconfident in an incorrect location. This behavior is quantitatively visible in Fig.~\ref{fig: error actual distance and theta}, where the distance between the MAP estimate and the true source, as well as the updates to the model error parameter, evolve more slowly for KL than for the Wasserstein metrics.

We also note that, because model correction can cause the posterior mass to be displaced relative to the main mass of the prior, evaluating KL divergence and its gradient w.r.t. design in this situation can, in principle, be numerically unstable: if the prior assigns near-zero probability to regions where the posterior places mass, $D_\text{KL}(p(\theta|y)||p(\theta))$ may diverge. While this instability was not observed in our experiments, it is a theoretical and potential issue.

In summary, under model discrepancy, the false reward effect associated with Wasserstein distances is also observed. However, the presence of model error alters the relative performance: KL’s tendency to concentrate posterior mass in the region favored by the incorrect model makes it overconfident in the wrong location, slowing subsequent corrections. Wasserstein distances, by selecting designs slightly offset from the source location, produce more diffuse and spatially balanced posteriors, which are easier to adjust after the model is corrected. This advantage in correction speed does not imply an inherent superiority of Wasserstein metrics, but within our model correction framework, Wasserstein-based designs achieve faster and more effective convergence to the truth than those based on KL divergence.

\subsubsection{Case 4: Expected information gain}\label{sec: case 4}

\begin{figure}[H]
  \centering
  \begin{subfigure}[t]{0.32\textwidth}
    \centering
    \includegraphics[width=\linewidth]{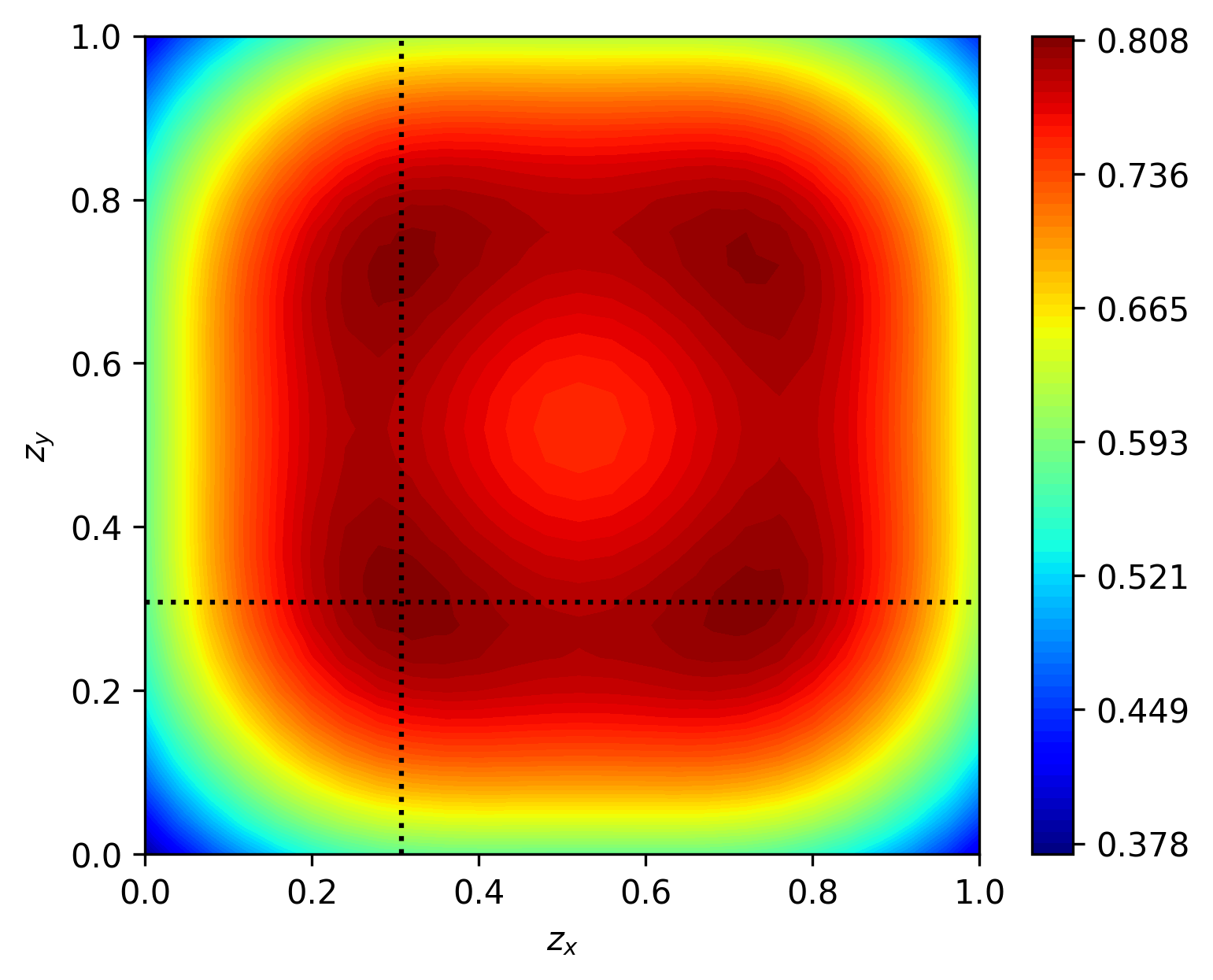}
    \caption{KL}
    \label{fig: error reward KL}
  \end{subfigure}
  \begin{subfigure}[t]{0.32\textwidth}
    \centering
    \includegraphics[width=\linewidth]{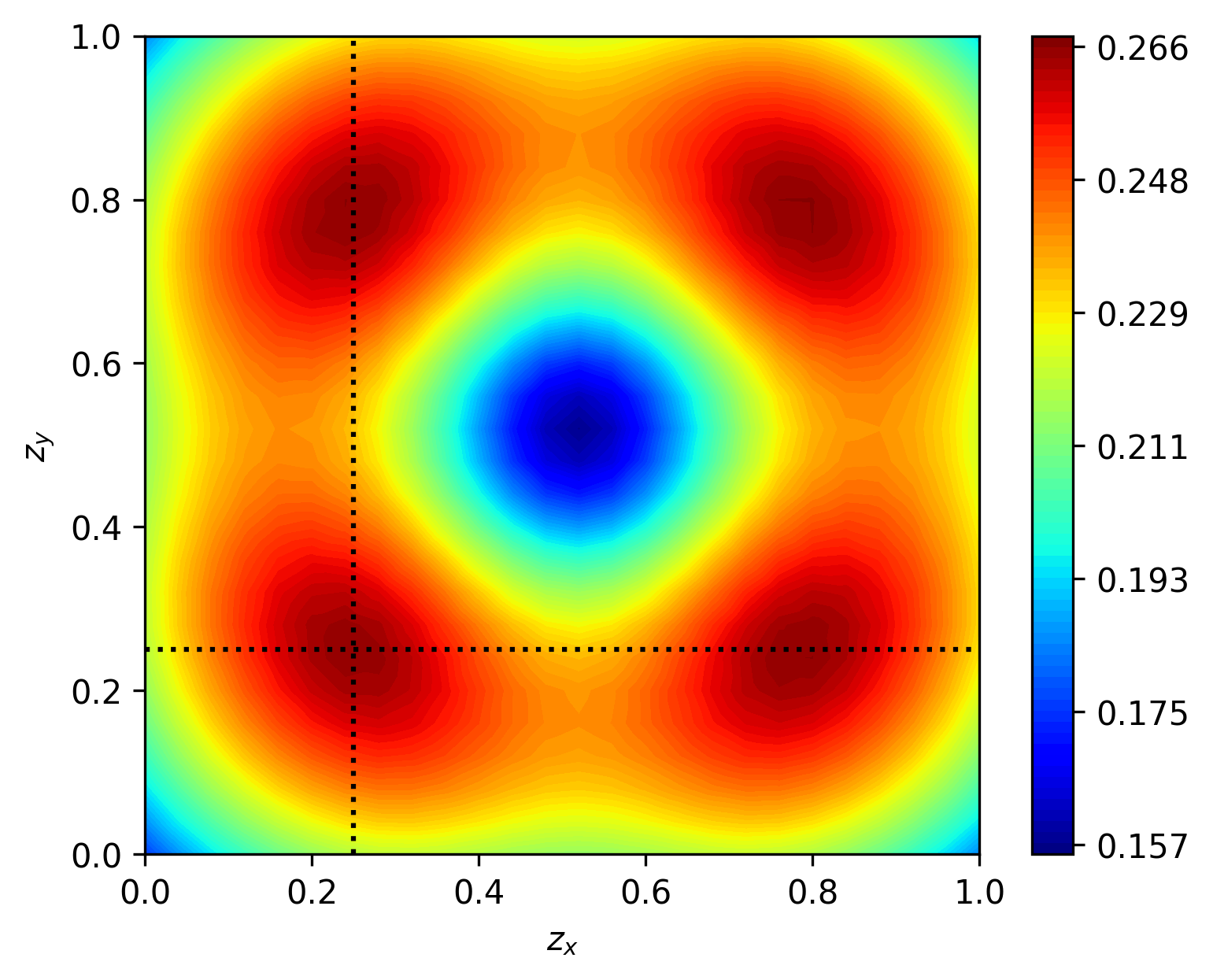}
    \caption{$W_2$}
    \label{fig: error reward w2}
  \end{subfigure}
  \begin{subfigure}[t]{0.32\textwidth}
    \centering
    \includegraphics[width=\linewidth]{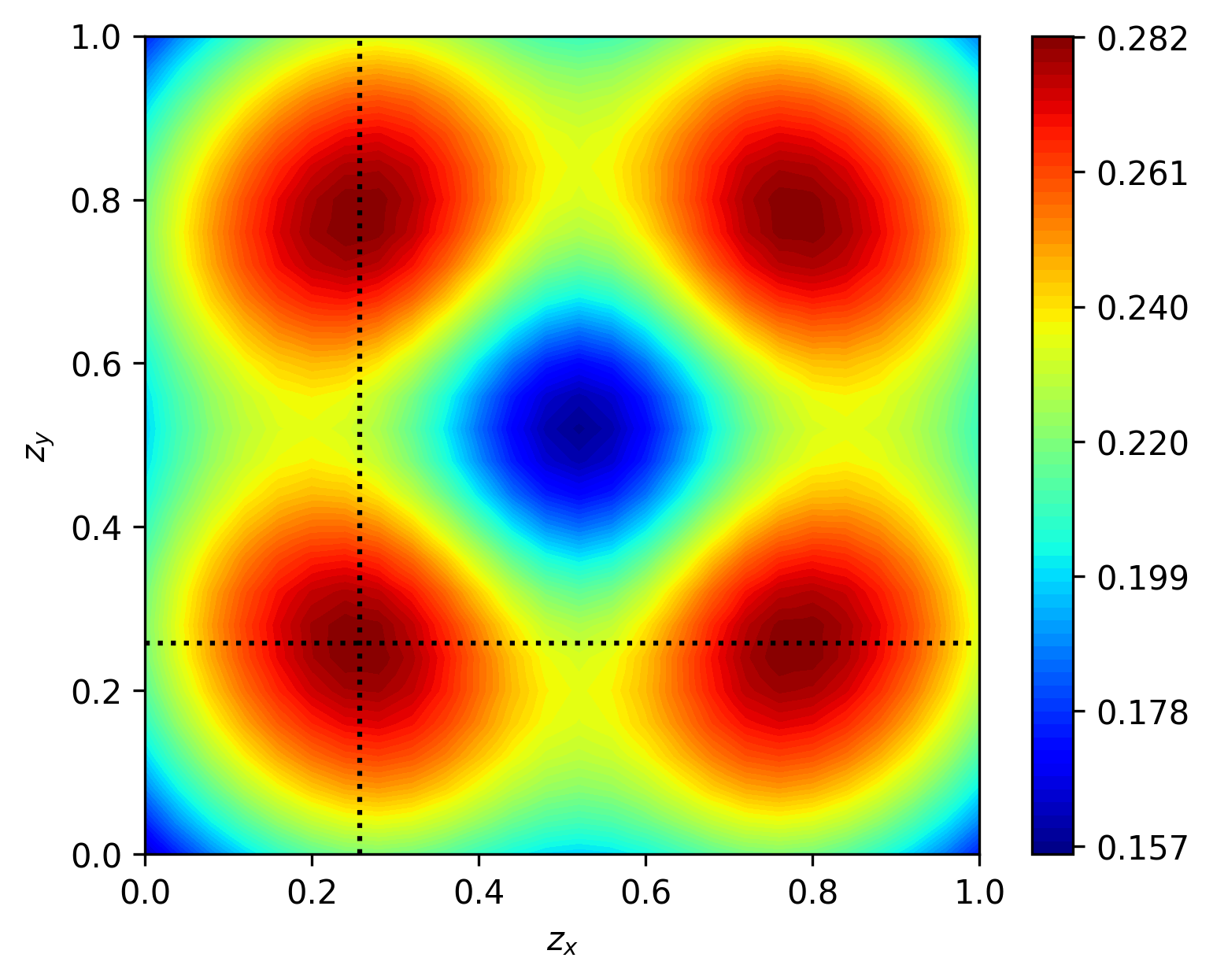}
    \caption{$W_1$}
    \label{fig: error reward w1}
  \end{subfigure}
  \caption{Reward map with model discrepancy at stage 1 for Case 4, where model discrepancy is present. The black dotted lines indicate the $z_x$ and $z_y$ coordinates of the global maximizer.}
  \label{fig: error reward map}
\end{figure}

In this section, we switch to the expected utility functions. Figure~\ref{fig: error reward map} shows the reward map at stage 1 under model discrepancy. Again, the design point preferred by the expected Wasserstein distance lies farther toward the boundary than that of the expected KL divergence. Notably, although the result with model discrepancy (Fig.~\ref{fig: error reward map}) is very similar to that without discrepancy (Fig.~\ref{fig: no error reward map}), they are not the same: with discrepancy, the most favored design is slightly closer to the center. The similarity is because we assume a parametric error, which preserves the model form and thus yields a similar overall pattern. 

\begin{figure}[H]
    \centering
    \includegraphics[width=\linewidth]{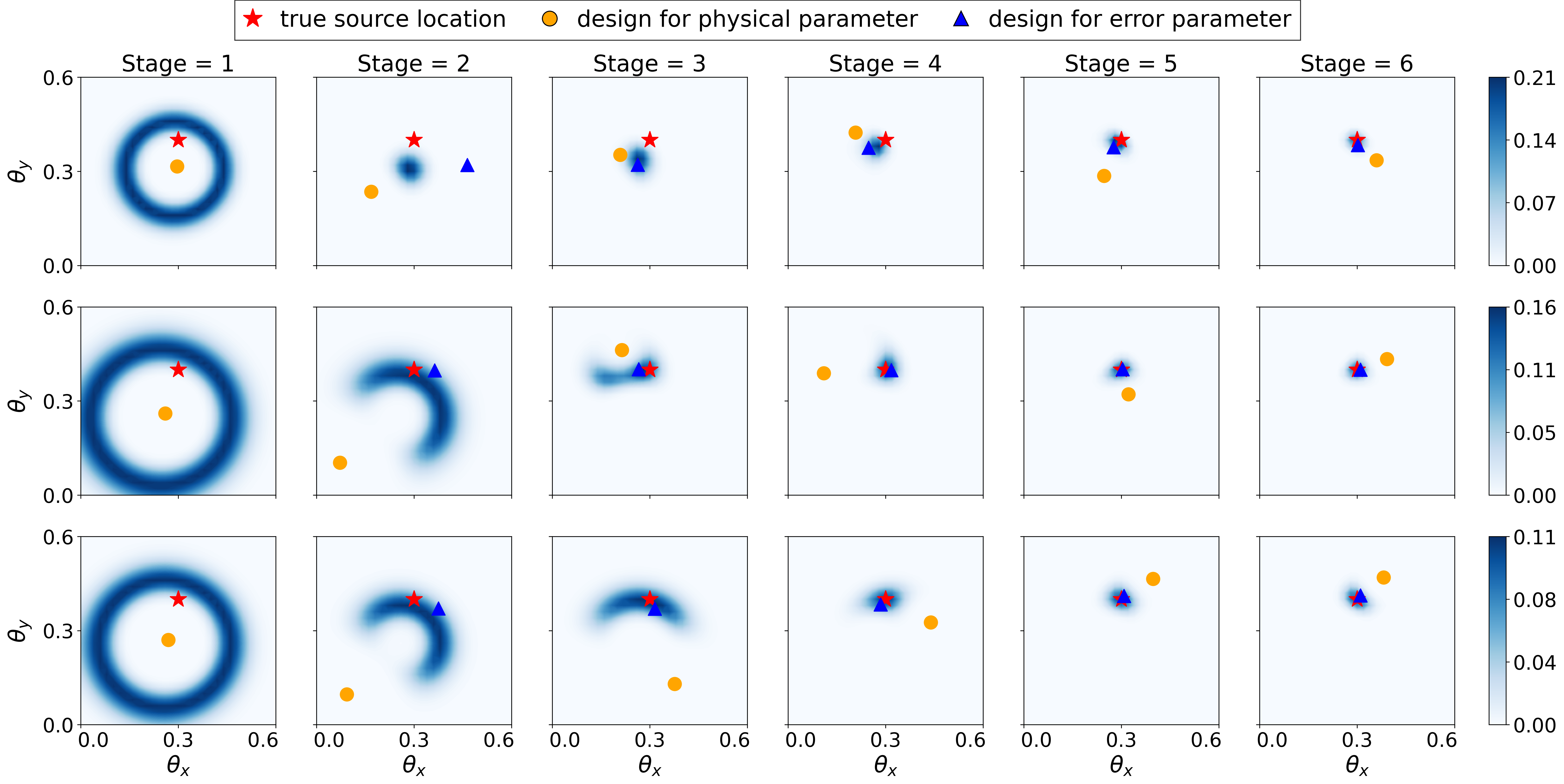}
    \caption{With error (case 4), using expected information gain: posterior. From top to bottom: KL, $W_2$, $W_1$. }
    \label{fig: error expected posterior}
\end{figure}

Figure~\ref{fig: error expected posterior} shows the evolution of the posterior distributions when using expected utility in the presence of model discrepancy (case 4). Both expected utilities exhibit the same vulnerability to model discrepancy as in the actual-utility case: at stage 1, the true parameter lies entirely outside the main posterior mass for both criteria. However, although the expected KL divergence produces a sharper posterior, model discrepancy drives most of its probability mass away from the true value. By contrast, the design selected by the expected Wasserstein distance is influenced by the false reward effect and sits farther from the center; its posterior is therefore less concentrated but places higher density on the truth.

\begin{figure}[H]
  \centering
  \begin{subfigure}[t]{0.48\textwidth}
    \centering
    \includegraphics[width=0.8\linewidth]{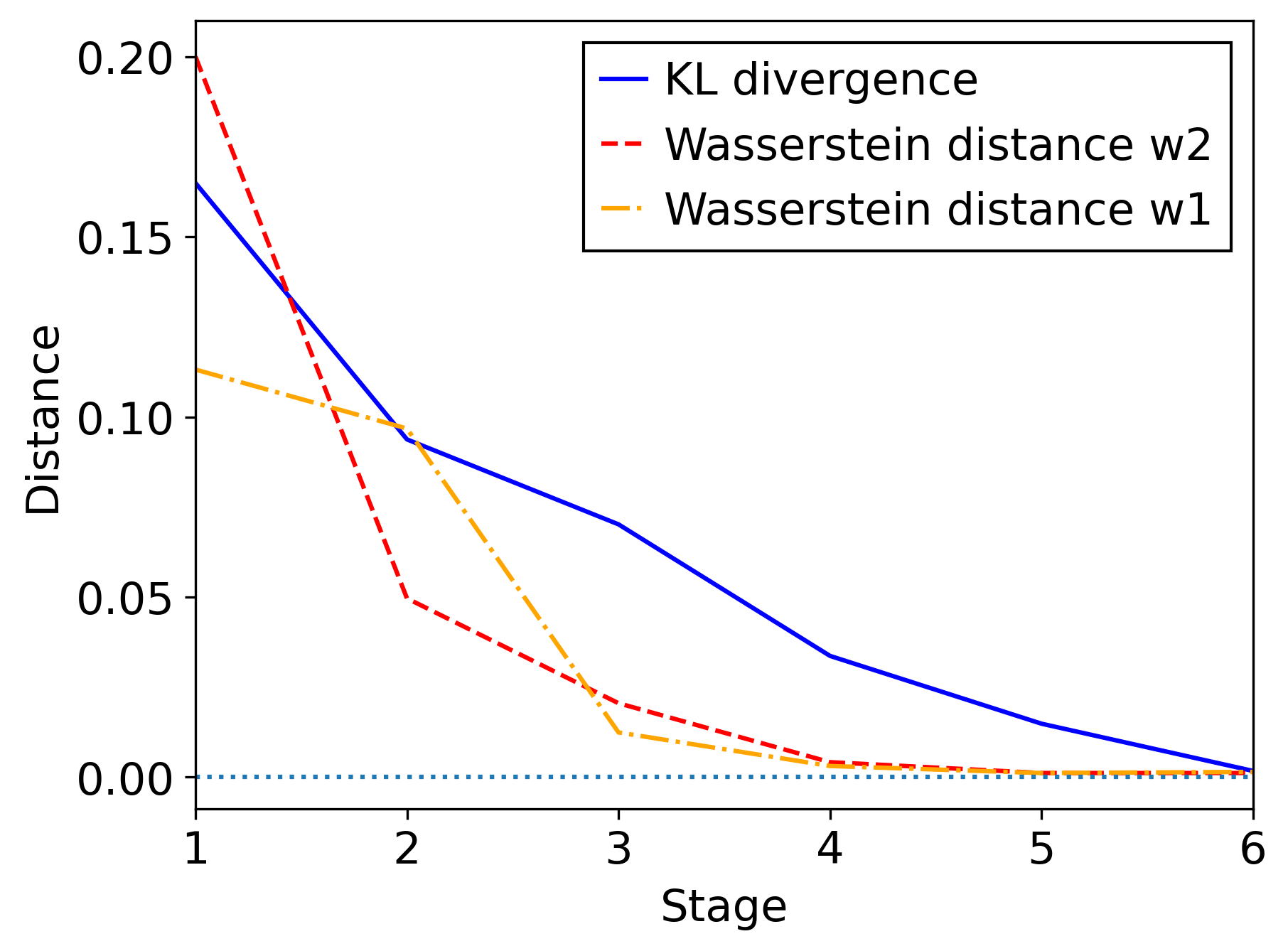}
    \caption{Distance}
    \label{fig: error expected distance}
  \end{subfigure}
  \hfill
  \begin{subfigure}[t]{0.48\textwidth}
    \centering
    \includegraphics[width=0.8\linewidth]{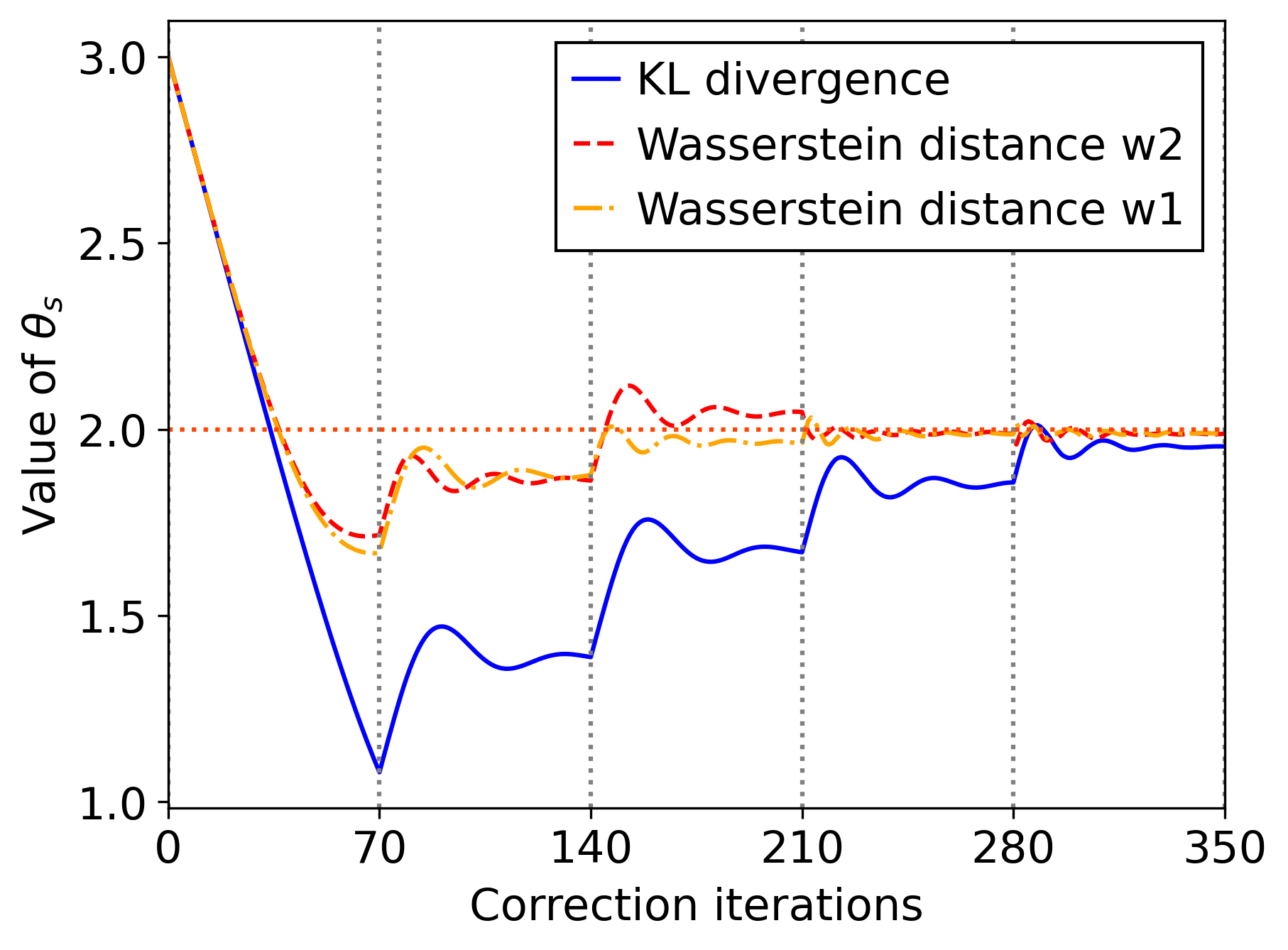}
    \caption{Error parameter $\theta_s$}
    \label{fig: error expected theta}
  \end{subfigure}
  \caption{Distance and parametric error evolution  for Case 4, where designs are selected using expected information gain in the presence of model discrepancy.}
  \label{fig: error expected distance and theta}
\end{figure}

In stages 2 and 3, the expected KL divergence continues to prefer over-peaked posteriors located at an incorrect value. Such over-concentration requires more steps to correct through Bayesian updating, so its posterior reaches and covers the truth only by stage 4. The expected Wasserstein distance keeps the main posterior mass closer to the truth throughout the first three stages and already covers the truth by stage 3. A quantitative comparison leads to the same conclusion: the MAP estimate under the expected Wasserstein distance remains consistently closer to the true source than that obtained under the expected KL divergence (see Fig.~\ref{fig: error expected distance}). These results indicate that, under model discrepancy, the expected Wasserstein distance yields more reliable belief updating than the expected KL divergence.

Figure~\ref{fig: error expected theta} further reports the evolution of the model‐error parameter. When using the expected Wasserstein distance, the discrepancy parameter converges to the true value significantly faster than under the expected KL divergence. The reason is twofold. First, in the early stages, the over-concentrated but mislocated posterior produced by the expected KL divergence yields a poor MAP estimate, which in turn provides a suboptimal input for the discrepancy correction. Second, the slower discrepancy correction feeds back into the inference loop and delays the subsequent posterior refinement. Taken together, the results in this case demonstrate that within our discrepancy-correction framework, the expected Wasserstein distance yields more effective belief updating and error correction than the expected KL divergence.

In comparison with Section~\ref{sec: actual error} (actual utility), the underlying mechanism with model discrepancy is similar but not identical. Under the actual-utility setting, Wasserstein distance consistently outperforms KL divergence across our tested configurations. The efficiency of correction hinges on how much uncertainty each utility retains around the truth at early stages: the more uncertainty retained around the truth, the faster the subsequent correction. With expected utilities, however, the outer expectation introduces extra uncertainty. This uncertainty primarily narrows the performance gap between the two criteria, while the expected Wasserstein distance still outperforms the expected KL divergence in most cases.

In summary, when model discrepancy is present, the false increase effect of Wasserstein utilities can still be observed. However, the introduction of expected utilities narrows the relative performance in probability. In most cases, Wasserstein-based designs leave more balanced uncertainty and thus converge more effectively in both parameter inference and model error updating. The expected KL divergence occasionally outperforms the expected Wasserstein distance when the true source lies in certain regions of the domain, but such cases are rare in our experiments. Overall, expected Wasserstein utilities provide more reliable performance than the expected KL divergence under model discrepancy.

\section{Conclusion}\label{Conclusion}
In this work, we first design a toy example to illustrate a potential limitation of Wasserstein distances in BED: the Wasserstein distance is more sensitive to shifts in the posterior mean, especially under a non-informative prior such as uniform distribution. Such a sensitivity can potentially lead to false rewards unrelated to any useful information gain. We then systematically compared Wasserstein distance and KL divergence for utility functions of Bayesian experimental design through a classical source inversion problem. In the absence of model discrepancy, KL divergence provides more effective early-stage designs under actual utilities, producing faster convergence of posteriors toward the truth. On the contrary, Wasserstein distances are affected by location bias and yield more diffuse posteriors. The difference between the two metrics narrows under expected utilities, while the KL divergence still outperforms Wasserstein metrics in most cases. On the other hand, Wasserstein metrics provide a superior performance than the KL divergence if the model discrepancy is non-negligible. More specifically, KL divergence leads to more confident design choices that concentrate posterior mass around incorrect parameter values, while Wasserstein metrics retain more uncertainties in the posterior distribution and thus avoid the convergence to biased results of both physics-based parameters and the model corrections. These results suggest that Wasserstein metrics, though not universally superior, can offer more robust BED results under model discrepancy, while KL divergence tends to be more efficient in BED tasks when the forward model is accurate.

\section{Discussion}\label{discussion}

We clarify the relation between the sufficiency-ordering property in \cite{helin2025bayesian} and the model-discrepancy setting considered in this paper: they address distinct questions. The sufficiency-ordering property concerns an information criterion for a specified Bayesian experimental design problem. Once the target parameter, prior, and observation model used to compute the expected utility are specified, whether one design is sufficient for another is determined by that Bayesian model, rather than by realized measurements from the true physical system. Therefore, even when the model used for design differs from the true system, the expected Wasserstein utility retains the sufficiency-ordering interpretation for the specified design problem.

The model-discrepancy issue studied in this paper is different. We investigate whether designs selected by KL- or Wasserstein-based expected utilities lead to effective posterior updates and model-error correction when measurements are generated from the true system. Under model discrepancy, a design that is preferred by the expected utility under the model used for design is not necessarily the design that gives the best posterior behavior relative to the true system. This does not contradict the sufficiency-ordering property. Instead, it reflects the gap between expected-utility optimality under the model used for design and practical performance under the true data-generating system. Our numerical results focus on this practical gap and show that, in the discrepancy regimes considered here, Wasserstein-based designs retain more uncertainty around the truth and thereby improve subsequent belief updating and model error correction.

Relatedly, the potential instability of KL divergence under weak prior-posterior overlap can be mitigated by several standard modifications. One simple mitigation is likelihood tempering, or more generally, a generalized-Bayes update, which downweights a potentially misspecified likelihood and produces a less concentrated posterior \cite{holmes2017assigning,chhachhi20231,wu2023comparison}. Another possibility is to robustify the EIG objective itself, such as by using a KL-ambiguity-set formulation of robust EIG \cite{go2022robust}. These modifications introduce additional tuning parameters or would change the design objective or posterior update. We therefore do not include them in our comparison.

\section*{Acknowledgments}
H.Y., X.D., and J.W. are supported by the University of Wisconsin-Madison, Office of the Vice Chancellor for Research and Graduate Education with funding from the Wisconsin Alumni Research Foundation.

\section*{Data Availability}
The codes and data that support the findings of this study are available in the link: \url{https://github.com/AIMS-Madison/BED-AD-EKI-Wasserstein}..
  
\bibliographystyle{unsrt}
\bibliography{references}

@article{wu2024learning,
  title={Learning about structural errors in models of complex dynamical systems},
  author={Wu, Jin-Long and Levine, Matthew E and Schneider, Tapio and Stuart, Andrew},
  journal={Journal of Computational Physics},
  pages={113157},
  year={2024},
  publisher={Elsevier}
}

@article{gelbrich1990formula,
  title={On a formula for the L2 {Wasserstein} metric between measures on {Euclidean} and {Hilbert} spaces},
  author={Gelbrich, Matthias},
  journal={Mathematische Nachrichten},
  volume={147},
  number={1},
  pages={185--203},
  year={1990},
  publisher={Wiley Online Library}
}

@article{peyre2019computational,
  title={Computational optimal transport: With applications to data science},
  author={Peyr{\'e}, Gabriel and Cuturi, Marco and others},
  journal={Foundations and Trends{\textregistered} in Machine Learning},
  volume={11},
  number={5-6},
  pages={355--607},
  year={2019},
  publisher={Now Publishers, Inc.}
}

@book{villani2008optimal,
  title={Optimal transport: old and new},
  author={Villani, C{\'e}dric and others},
  volume={338},
  year={2008},
  publisher={Springer}
}

@article{rizzo2016energy,
  title={Energy distance},
  author={Rizzo, Maria L and Sz{\'e}kely, G{\'a}bor J},
  journal={wiley interdisciplinary reviews: Computational statistics},
  volume={8},
  number={1},
  pages={27--38},
  year={2016},
  publisher={Wiley Online Library}
}

@article{huan2013simulation,
  title={Simulation-based optimal {Bayesian} experimental design for nonlinear systems},
  author={Huan, Xun and Marzouk, Youssef M},
  journal={Journal of Computational Physics},
  volume={232},
  number={1},
  pages={288--317},
  year={2013},
  publisher={Elsevier}
}

@article{houlsby2011bayesian,
  title={Bayesian active learning for classification and preference learning},
  author={Houlsby, Neil and Husz{\'a}r, Ferenc and Ghahramani, Zoubin and Lengyel, M{\'a}t{\'e}},
  journal={arXiv preprint arXiv:1112.5745},
  year={2011}
}

@article{nott2023bayesian,
  title={Bayesian inference for misspecified generative models},
  author={Nott, David J and Drovandi, Christopher and Frazier, David T},
  journal={Annual Review of Statistics and Its Application},
  volume={11},
  year={2023},
  publisher={Annual Reviews}
}

@article{helin2025bayesian,
  title={Bayesian optimal experimental design with {Wasserstein} information criteria},
  author={Helin, Tapio and Marzouk, Youssef and Rojo-Garcia, Jose Rodrigo},
  journal={arXiv preprint arXiv:2504.10092},
  year={2025}
}

@article{tolstikhin2017wasserstein,
  title={Wasserstein auto-encoders},
  author={Tolstikhin, Ilya and Bousquet, Olivier and Gelly, Sylvain and Schoelkopf, Bernhard},
  journal={arXiv preprint arXiv:1711.01558},
  year={2017}
}

@inproceedings{yi2023sliced,
  title={Sliced {Wasserstein} variational inference},
  author={Yi, Mingxuan and Liu, Song},
  booktitle={Asian conference on machine learning},
  pages={1213--1228},
  year={2023},
  organization={PMLR}
}

@article{ambrogioni2018wasserstein,
  title={Wasserstein variational inference},
  author={Ambrogioni, Luca and G{\"u}{\c{c}}l{\"u}, Umut and G{\"u}{\c{c}}l{\"u}t{\"u}rk, Ya{\u{g}}mur and Hinne, Max and van Gerven, Marcel A and Maris, Eric},
  journal={Advances in Neural Information Processing Systems},
  volume={31},
  year={2018}
}

@article{namkoong2016stochastic,
  title={Stochastic gradient methods for distributionally robust optimization with f-divergences},
  author={Namkoong, Hongseok and Duchi, John C},
  journal={Advances in neural information processing systems},
  volume={29},
  year={2016}
}

@article{sinha2017certifying,
  title={Certifying some distributional robustness with principled adversarial training},
  author={Sinha, Aman and Namkoong, Hongseok and Volpi, Riccardo and Duchi, John},
  journal={arXiv preprint arXiv:1710.10571},
  year={2017}
}

@article{rodriguez2021tighter,
  title={Tighter expected generalization error bounds via {Wasserstein} distance},
  author={Rodr{\'\i}guez G{\'a}lvez, Borja and Bassi, Germ{\'a}n and Thobaben, Ragnar and Skoglund, Mikael},
  journal={Advances in Neural Information Processing Systems},
  volume={34},
  pages={19109--19121},
  year={2021}
}

@article{arjovsky2017wasserstein,
  title={Wasserstein GAN. arXiv e-prints},
  author={Arjovsky, Martin and Chintala, Soumith and Bottou, L{\'e}on},
  journal={arXiv preprint arXiv:1701.07875},
  volume={685},
  year={2017}
}

@article{lee2018minimax,
  title={Minimax statistical learning with {Wasserstein} distances},
  author={Lee, Jaeho and Raginsky, Maxim},
  journal={Advances in Neural Information Processing Systems},
  volume={31},
  year={2018}
}

@article{wu2023large,
  title={Large-scale {Bayesian} optimal experimental design with derivative-informed projected neural network},
  author={Wu, Keyi and O’Leary-Roseberry, Thomas and Chen, Peng and Ghattas, Omar},
  journal={Journal of Scientific Computing},
  volume={95},
  number={1},
  pages={30},
  year={2023},
  publisher={Springer}
}

@article{sriperumbudur2012empirical,
  title={On the empirical estimation of integral probability metrics},
  author={Sriperumbudur, Bharath K and Fukumizu, Kenji and Gretton, Arthur and Sch{\"o}lkopf, Bernhard and Lanckriet, Gert RG},
  year={2012}
}

@article{o2022learning,
  title={Learning high-dimensional parametric maps via reduced basis adaptive residual networks},
  author={O’Leary-Roseberry, Thomas and Du, Xiaosong and Chaudhuri, Anirban and Martins, Joaquim RRA and Willcox, Karen and Ghattas, Omar},
  journal={Computer Methods in Applied Mechanics and Engineering},
  volume={402},
  pages={115730},
  year={2022},
  publisher={Elsevier}
}

@article{stuart2010inverse,
  title={Inverse problems: a {Bayesian} perspective},
  author={Stuart, Andrew M},
  journal={Acta numerica},
  volume={19},
  pages={451--559},
  year={2010},
  publisher={Cambridge University Press}
}

@article{chanda2020information,
  title={Information theory in computational biology: where we stand today},
  author={Chanda, Pritam and Costa, Eduardo and Hu, Jie and Sukumar, Shravan and Van Hemert, John and Walia, Rasna},
  journal={Entropy},
  volume={22},
  number={6},
  pages={627},
  year={2020},
  publisher={MDPI}
}

@article{adami2004information,
  title={Information theory in molecular biology},
  author={Adami, Christoph},
  journal={Physics of Life Reviews},
  volume={1},
  number={1},
  pages={3--22},
  year={2004},
  publisher={Elsevier}
}

@article{sriperumbudur2009integral,
  title={On integral probability metrics,$\backslash$phi-divergences and binary classification},
  author={Sriperumbudur, Bharath K and Fukumizu, Kenji and Gretton, Arthur and Sch{\"o}lkopf, Bernhard and Lanckriet, Gert RG},
  journal={arXiv preprint arXiv:0901.2698},
  year={2009}
}

@article{reid2011information,
  title={Information, divergence and risk for binary experiments},
  author={Reid, Mark and Williamson, Robert},
  year={2011},
  publisher={MIT Press}
}

@article{chhachhi20231,
  title={On the 1-{Wasserstein} distance between location-scale distributions and the effect of differential privacy},
  author={Chhachhi, Saurab and Teng, Fei},
  journal={arXiv preprint arXiv:2304.14869},
  year={2023}
}

@article{panaretos2019statistical,
  title={Statistical aspects of {Wasserstein} distances},
  author={Panaretos, Victor M and Zemel, Yoav},
  journal={Annual review of statistics and its application},
  volume={6},
  number={1},
  pages={405--431},
  year={2019},
  publisher={Annual Reviews}
}

@inproceedings{sejourne2022faster,
  title={Faster unbalanced optimal transport: Translation invariant {Sinkhorn} and 1-d {Frank-Wolfe}},
  author={S{\'e}journ{\'e}, Thibault and Vialard, Fran{\c{c}}ois-Xavier and Peyr{\'e}, Gabriel},
  booktitle={International Conference on Artificial Intelligence and Statistics},
  pages={4995--5021},
  year={2022},
  organization={PMLR}
}

@article{wang2024relative,
  title={Relative-Translation Invariant {Wasserstein} Distance},
  author={Wang, Binshuai and Di, Qiwei and Yin, Ming and Wang, Mengdi and Gu, Quanquan and Wei, Peng},
  journal={arXiv preprint arXiv:2409.02416},
  year={2024}
}

@inproceedings{kato2023unified,
  title={Unified perspective on probability divergence via the density-ratio likelihood: Bridging {KL}-divergence and integral probability metrics},
  author={Kato, Masahiro and Imaizumi, Masaaki and Minami, Kentaro},
  booktitle={International Conference on Artificial Intelligence and Statistics},
  pages={5271--5298},
  year={2023},
  organization={PMLR}
}

@article{owhadi2013optimal,
  title={Optimal uncertainty quantification},
  author={Owhadi, Houman and Scovel, Clint and Sullivan, Timothy John and McKerns, Mike and Ortiz, Michael},
  journal={Siam Review},
  volume={55},
  number={2},
  pages={271--345},
  year={2013},
  publisher={SIAM}
}

@book{pardo2018statistical,
  title={Statistical inference based on divergence measures},
  author={Pardo, Leandro},
  year={2018},
  publisher={Chapman and Hall/CRC}
}

@article{rainforth2024modern,
  title={Modern Bayesian experimental design},
  author={Rainforth, Tom and Foster, Adam and Ivanova, Desi R and Bickford Smith, Freddie},
  journal={Statistical Science},
  volume={39},
  number={1},
  pages={100--114},
  year={2024},
  publisher={Institute of Mathematical Statistics}
}

@article{brynjarsdottir2014learning,
  title={Learning about physical parameters: The importance of model discrepancy},
  author={Brynjarsd{\'o}ttir, Jenn{\`y} and O'Hagan, Anthony},
  journal={Inverse problems},
  volume={30},
  number={11},
  pages={114007},
  year={2014},
  publisher={IOP Publishing}
}

@phdthesis{feng_optimal_2015,
	type = {Thesis},
	title = {Optimal {Bayesian} experimental design in the presence of model error},
	url = {https://dspace.mit.edu/handle/1721.1/97790},
	language = {eng},
	urldate = {2024-07-09},
	school = {Massachusetts Institute of Technology},
	author = {Feng, Chi},
	year = {2015},
	note = {Accepted: 2015-07-17T19:46:48Z},
}

@inproceedings{catanach2023metrics,
  title={Metrics for {Bayesian} optimal experiment design under model misspecification},
  author={Catanach, Tommie A and Das, Niladri},
  booktitle={2023 62nd IEEE Conference on Decision and Control (CDC)},
  pages={7707--7714},
  year={2023},
  organization={IEEE}
}

@article{levine2022framework,
  title={A framework for machine learning of model error in dynamical systems},
  author={Levine, Matthew and Stuart, Andrew},
  journal={Communications of the American Mathematical Society},
  volume={2},
  number={07},
  pages={283--344},
  year={2022}
}

@article{dong2025data,
  title={Data-driven stochastic closure modeling via conditional diffusion model and neural operator},
  author={Dong, Xinghao and Chen, Chuanqi and Wu, Jin-Long},
  journal={Journal of Computational Physics},
  volume={534},
  pages={114005},
  year={2025},
  publisher={Elsevier}
}

@article{ebers2024discrepancy,
  title={Discrepancy modeling framework: Learning missing physics, modeling systematic residuals, and disambiguating between deterministic and random effects},
  author={Ebers, Megan R and Steele, Katherine M and Kutz, J Nathan},
  journal={SIAM Journal on Applied Dynamical Systems},
  volume={23},
  number={1},
  pages={440--469},
  year={2024},
  publisher={SIAM}
}

@article{iglesias2013ensemble,
  title={Ensemble {Kalman} methods for inverse problems},
  author={Iglesias, Marco A and Law, Kody JH and Stuart, Andrew M},
  journal={Inverse Problems},
  volume={29},
  number={4},
  pages={045001},
  year={2013},
  publisher={IOP Publishing}
}

@article{chen2022autodifferentiable,
  title={Autodifferentiable ensemble {Kalman} filters},
  author={Chen, Yuming and Sanz-Alonso, Daniel and Willett, Rebecca},
  journal={SIAM Journal on Mathematics of Data Science},
  volume={4},
  number={2},
  pages={801--833},
  year={2022},
  publisher={SIAM}
}

@article{kennedy2001bayesian,
  title={Bayesian calibration of computer models},
  author={Kennedy, Marc C and O'Hagan, Anthony},
  journal={Journal of the Royal Statistical Society: Series B (Statistical Methodology)},
  volume={63},
  number={3},
  pages={425--464},
  year={2001},
  publisher={Wiley Online Library}
}

@book{lindley1972bayesian,
  title={Bayesian statistics: A review},
  author={Lindley, Dennis Victor},
  year={1972},
  publisher={SIAM}
}

@article{huan2024optimal,
  title={Optimal experimental design: Formulations and computations},
  author={Huan, Xun and Jagalur, Jayanth and Marzouk, Youssef},
  journal={Acta Numerica},
  volume={33},
  pages={715--840},
  year={2024},
  publisher={Cambridge University Press}
}

@article{lindley1956measure,
  title={On a measure of the information provided by an experiment},
  author={Lindley, Dennis V},
  journal={The Annals of Mathematical Statistics},
  volume={27},
  number={4},
  pages={986--1005},
  year={1956},
  publisher={Institute of Mathematical Statistics}
}

@article{huan2014gradient,
  title={Gradient-based stochastic optimization methods in {Bayesian} experimental design},
  author={Huan, Xun and Marzouk, Youssef M},
  journal={International Journal for Uncertainty Quantification},
  volume={4},
  number={6},
  year={2014},
  publisher={Begel House Inc.}
}

@article{yang2025bayesian,
title = {Bayesian experimental design for model discrepancy calibration: An auto-differentiable ensemble {Kalman} inversion approach},
journal = {Journal of Computational Physics},
volume = {545},
pages = {114469},
year = {2026},
issn = {0021-9991},
doi = {https://doi.org/10.1016/j.jcp.2025.114469},
url = {https://www.sciencedirect.com/science/article/pii/S002199912500751X},
author = {Huchen Yang and Xinghao Dong and Jin-Long Wu},
}

@article{shen2023bayesian,
  title={Bayesian sequential optimal experimental design for nonlinear models using policy gradient reinforcement learning},
  author={Shen, Wanggang and Huan, Xun},
  journal={Computer Methods in Applied Mechanics and Engineering},
  volume={416},
  pages={116304},
  year={2023},
  publisher={Elsevier}
}

@article{YANG2025118198,
title = {Active learning of model discrepancy with {Bayesian} experimental design},
journal = {Computer Methods in Applied Mechanics and Engineering},
volume = {446},
pages = {118198},
year = {2025},
issn = {0045-7825},
doi = {https://doi.org/10.1016/j.cma.2025.118198},
url = {https://www.sciencedirect.com/science/article/pii/S0045782525004700},
author = {Huchen Yang and Chuanqi Chen and Jin-Long Wu},
}

@article{chaloner1995bayesian,
  title={Bayesian experimental design: A review},
  author={Chaloner, Kathryn and Verdinelli, Isabella},
  journal={Statistical science},
  pages={273--304},
  year={1995},
  publisher={JSTOR}
}

@article{jones2016bayes,
  title={Bayes linear analysis for {Bayesian} optimal experimental design},
  author={Jones, Matthew and Goldstein, Michael and Jonathan, Philip and Randell, David},
  journal={Journal of Statistical Planning and Inference},
  volume={171},
  pages={115--129},
  year={2016},
  publisher={Elsevier}
}

@article{ryan2016review,
  title={A review of modern computational algorithms for {Bayesian} optimal design},
  author={Ryan, Elizabeth G and Drovandi, Christopher C and McGree, James M and Pettitt, Anthony N},
  journal={International Statistical Review},
  volume={84},
  number={1},
  pages={128--154},
  year={2016},
  publisher={Wiley Online Library}
}

@article{grunwald_inconsistency_2017,
  title={Inconsistency of {Bayesian} inference for misspecified linear models, and a proposal for repairing it},
  author={Gr{\"u}nwald, Peter and Van Ommen, Thijs},
  journal = {Bayesian Analysis},
  pages = {1069--1103},
  year={2017}
}

@article{gibbs2002choosing,
  title={On choosing and bounding probability metrics},
  author={Gibbs, Alison L and Su, Francis Edward},
  journal={International statistical review},
  volume={70},
  number={3},
  pages={419--435},
  year={2002},
  publisher={Wiley Online Library}
}

@inproceedings{go2022robust,
  title={Robust expected information gain for optimal {Bayesian} experimental design using ambiguity sets},
  author={Go, Jinwoo and Isaac, Tobin},
  booktitle={Uncertainty in Artificial Intelligence},
  pages={728--737},
  year={2022},
  organization={PMLR}
}

@article{wu2023comparison,
  title={A comparison of learning rate selection methods in generalized {Bayesian} inference},
  author={Wu, Pei-Shien and Martin, Ryan},
  journal={Bayesian Analysis},
  volume={18},
  number={1},
  pages={105--132},
  year={2023},
  publisher={International Society for Bayesian Analysis}
}

@article{holmes2017assigning,
  title={Assigning a value to a power likelihood in a general {Bayesian} model},
  author={Holmes, Chris C and Walker, Stephen G},
  journal={Biometrika},
  volume={104},
  number={2},
  pages={497--503},
  year={2017},
  publisher={Oxford University Press}
}

@article{csiszar1967information,
  title={Information-type measures of difference of probability distributions and indirect observation},
  author={CSISZAR, I},
  journal={Studia Scientiarum Mathematicarum Hungarica},
  volume={2},
  pages={229--318},
  year={1967}
}

@article{lin2002divergence,
  title={Divergence measures based on the {Shannon} entropy},
  author={Lin, Jianhua},
  journal={IEEE Transactions on Information theory},
  volume={37},
  number={1},
  pages={145--151},
  year={2002},
  publisher={IEEE}
}

@article{marsaglia2003evaluating,
  title={Evaluating {Kolmogorov}'s distribution},
  author={Marsaglia, George and Tsang, Wai Wan and Wang, Jingbo},
  journal={Journal of statistical software},
  volume={8},
  pages={1--4},
  year={2003}
}

@inbook{Stephens1992,
  author    = {Stephens, M. A.},
  editor    = {Kotz, Samuel and Johnson, Norman L.},
  title     = {Introduction to Kolmogorov (1933) On the Empirical Determination of a Distribution},
  booktitle = {Breakthroughs in Statistics: Methodology and Distribution},
  year      = {1992},
  publisher = {Springer},
  address   = {New York, NY},
  pages     = {93--105},
  isbn      = {978-1-4612-4380-9},
  doi       = {10.1007/978-1-4612-4380-9_9},
  url       = {https://doi.org/10.1007/978-1-4612-4380-9_9},
}

@article{muller1997integral,
  title={Integral probability metrics and their generating classes of functions},
  author={M{\"u}ller, Alfred},
  journal={Advances in applied probability},
  volume={29},
  number={2},
  pages={429--443},
  year={1997},
  publisher={Cambridge University Press}
}

@article{liese2006divergences,
  title={On divergences and informations in statistics and information theory},
  author={Liese, Friedrich and Vajda, Igor},
  journal={IEEE Transactions on Information Theory},
  volume={52},
  number={10},
  pages={4394--4412},
  year={2006},
  publisher={IEEE}
}

@book{raiffa2000applied,
  title={Applied statistical decision theory},
  author={Raiffa, Howard and Schlaifer, Robert},
  year={2000},
  publisher={John Wiley \& Sons}
}

@book{degroot2005optimal,
  title={Optimal statistical decisions},
  author={DeGroot, Morris H},
  year={2005},
  publisher={John Wiley \& Sons}
}

@article{sebastiani2000maximum,
  title={Maximum entropy sampling and optimal {Bayesian} experimental design},
  author={Sebastiani, Paola and Wynn, Henry P},
  journal={Journal of the Royal Statistical Society: Series B (Statistical Methodology)},
  volume={62},
  number={1},
  pages={145--157},
  year={2000},
  publisher={Wiley Online Library}
}

\clearpage
\appendix
\section{Derivation of \(W_2\) for Delta Distributions}\label{appendix: derivation}

Before deriving the closed‐form expressions for \(W_2\), we recall why in one dimension the Wasserstein‐2 distance can be written in terms of two CDFs (and their inverses), and how those inverses take the explicit piecewise forms below.

\begin{enumerate}
  \item \textbf{CDF and quantile function.}  For any probability distribution \(P\) on \(\mathbb{R}\):
  \[
    F_P(x)=\Pr_{X\sim P}\bigl[X\le x\bigr]
    \quad\text{is its cumulative distribution function (CDF),}
  \]
  and its \emph{quantile function} (inverse CDF) is defined by
  \[
    F_P^{-1}(u)
    =\inf\{\,x:F_P(x)\ge u\}, 
    \qquad u\in[0,1].
  \]
  Intuitively, \(F_P^{-1}(u)\) is the location in the real line at which the fraction \(u\) of the total probability mass has been accumulated.

  \item \textbf{One‐dimensional optimal transport.}  A fundamental theorem in optimal transport on the real line states that the \(L^2\)‐Wasserstein distance between two distributions \(P\) and \(Q\) admits the simple “monotone‐rearrangement” formula
  \[
    W_2^2(P,Q)
    = \int_{0}^{1} \bigl(F_P^{-1}(u)\;-\;F_Q^{-1}(u)\bigr)^2 \,du.
  \]
  Here \(u\) plays the role of a “mass coordinate” ranging from 0 to 1.

  \item \textbf{Why piecewise expressions?}  
  \begin{itemize}
    \item For a \emph{delta distribution} \(P=\delta_{x_0}\), the CDF jumps from 0 to 1 at \(x_0\), so
    \[
      F_P^{-1}(u)=x_0,\quad \forall\,u\in[0,1].
    \]
    \item For the \emph{uniform distribution} \(Q=U[-A,A]\), 
    \[
      F_Q(x)=\frac{x + A}{2A},\quad x\in[-A,A],
    \]
    and hence its inverse is the linear map
    \[
      F_Q^{-1}(u) = -A + 2A\,u,\quad u\in[0,1].
    \]
  \end{itemize}
  Substituting these explicit quantile functions into the one‐dimensional formula yields the piecewise integrals we compute below.
\end{enumerate}

\textbf{For the three distributions}:
\begin{itemize}
    \item a delta distribution at 0, $\delta(0)$;
    \item two symmetric delta distributions, $0.5\delta(-\Delta x)+0.5\delta(\Delta x)$;
    \item a shifted symmetric delta distribution, $0.5\delta(-\Delta x+\mu)+0.5\delta(\Delta x+\mu)$, with  $\Delta x < A$ and $\Delta x +\mu< A$.
\end{itemize}

We compute
\[
W_2^2(P,Q)
=\int_0^1\bigl(F_P^{-1}(u)-F_Q^{-1}(u)\bigr)^2\,du,
\]
where \(Q=U[-A,A]\) has
\[
F_Q^{-1}(u)=-A+2Au,\quad u\in[0,1].
\]

\textbf{For a delta distribution} \(P=\delta(x-x_0)\), 
\[
F_P^{-1}(u)=x_0\quad(0\le u\le1),
\]
and one obtains
\begin{align*}
W_2^2(\delta_{x_0},\,U[-A,A])
&=\int_0^1\bigl(x_0 -(-A+2Au)\bigr)^2du
=\int_0^1\bigl(A + x_0 -2Au\bigr)^2du\\
&=\int_0^1\bigl((A+x_0)^2 -4A(A+x_0)u +4A^2u^2\bigr)\,du\\
&=(A+x_0)^2 -2A(A+x_0) +\tfrac{4}{3}A^2
=x_0^2 +\tfrac{1}{3}A^2.
\end{align*}
In particular:
\[
W_2^{(1)} = \sqrt{0^2 + \tfrac{A^2}{3}} = \frac{A}{\sqrt3}.
\]

\textbf{For two symmetric deltas} 
\(
P^{(2)} = \tfrac12\delta_{-\Delta x} + \tfrac12\delta_{+\Delta x},
\)
the quantile function is
\[
F_{P^{(2)}}^{-1}(u)
=\begin{cases}
-\Delta x,&0\le u<0.5,\\
+\Delta x,&0.5\le u\le1.
\end{cases}
\]
Hence
\begin{align*}
W_2^2(P^{(2)},Q)
&=\int_0^{0.5}(A-\Delta x-2Au)^2du
+\int_{0.5}^{1}(A+\Delta x-2Au)^2du\\
&=\Delta x^2 -A\,\Delta x + \frac{A^2}{3},
\end{align*}
so
\[
W_2^{(2)}=\sqrt{\Delta x^2 -A\,\Delta x + \tfrac{A^2}{3}}.
\]

\textbf{Finally, for the shifted symmetric deltas}
\(
P^{(3)} = \tfrac12\delta_{-(\Delta x-\mu)} + \tfrac12\delta_{+(\Delta x+\mu)},
\)
one finds
\[
W_2^2(P^{(3)},Q)
=\mu^2 + \Delta x^2 -A\,\Delta x + \frac{A^2}{3},
\]
and hence
\[
W_2^{(3)}
=\sqrt{\mu^2 + \Delta x^2 -A\,\Delta x + \tfrac{A^2}{3}}.
\]

\section{A reverse case without model discrepancy}\label{appendix: reverse case}

\begin{figure}[H]
    \centering
    \includegraphics[width=\linewidth]{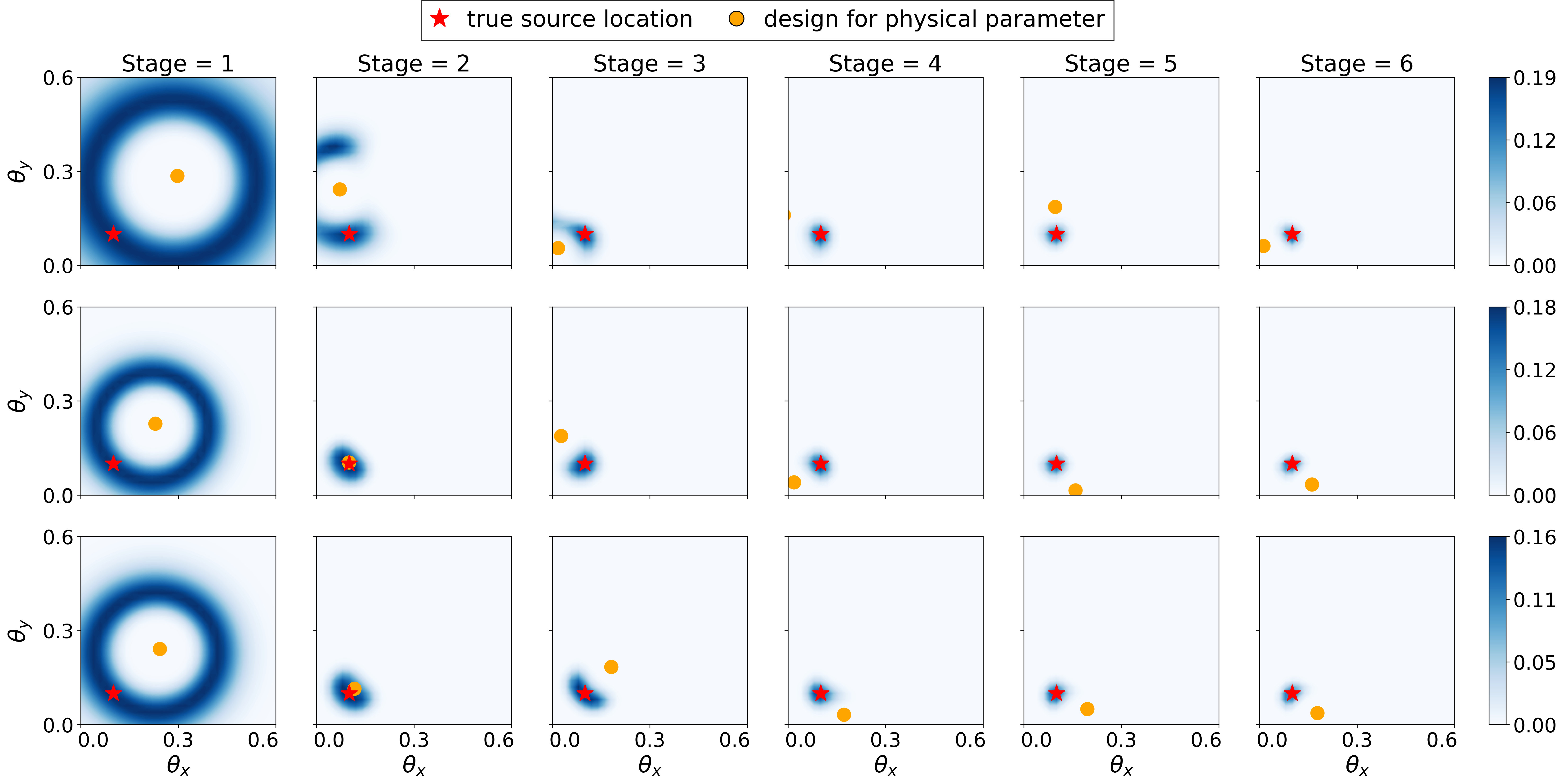}
    \caption{Without error (case 5 in the convection-diffusion problem), using expected information gain: posterior. From top to bottom: KL, $W_2$, $W_1$. }
    \label{fig: no error expected posterior reverse}
\end{figure}

\begin{figure}[H]
  \centering
  \begin{subfigure}[t]{0.48\textwidth}
    \centering
    \includegraphics[width=0.8\linewidth]{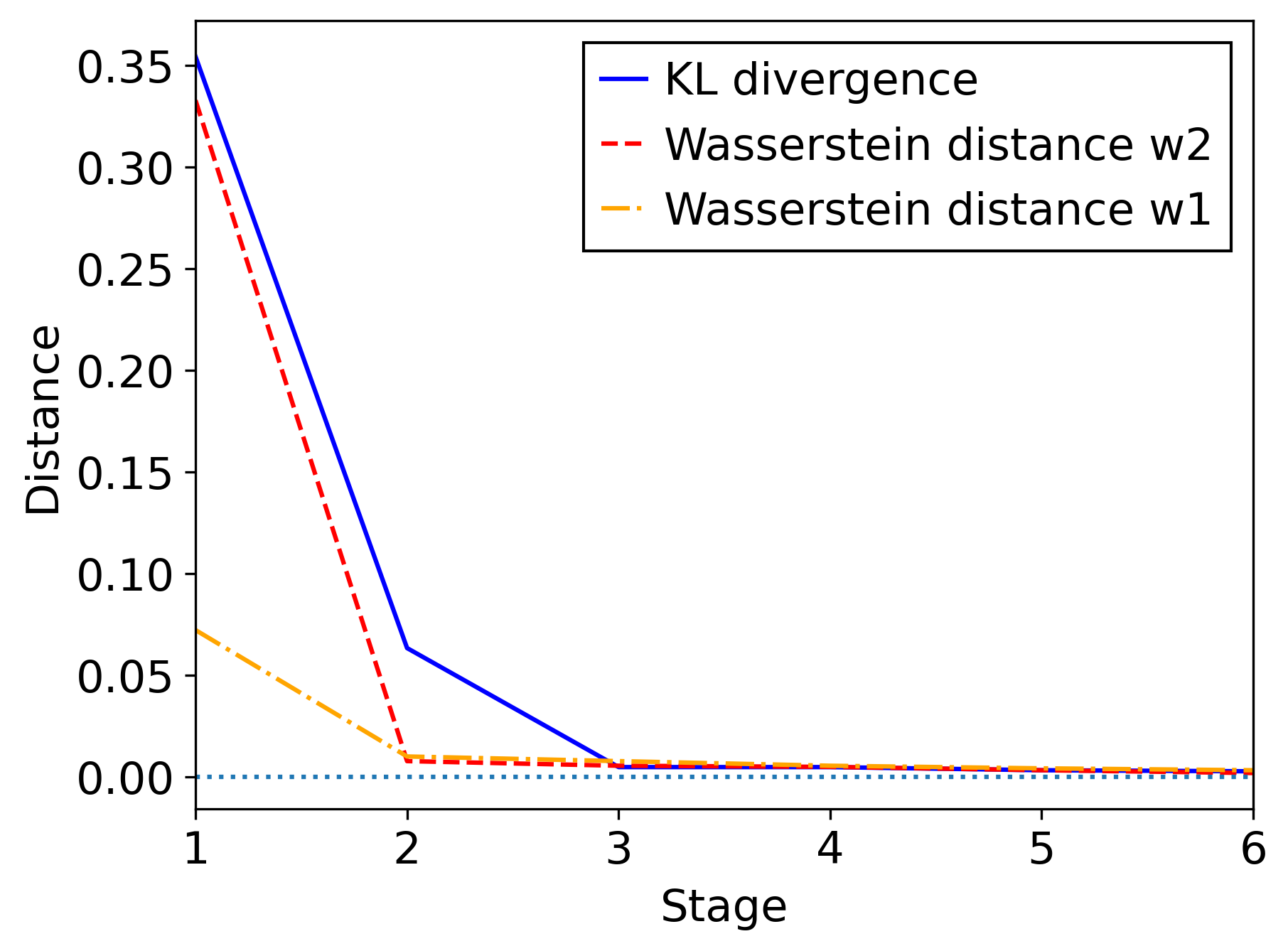}
    \caption{Distance}
    \label{fig: no error expected distance reverse}
  \end{subfigure}
  \hfill
  \begin{subfigure}[t]{0.48\textwidth}
    \centering
    \includegraphics[width=0.8\linewidth]{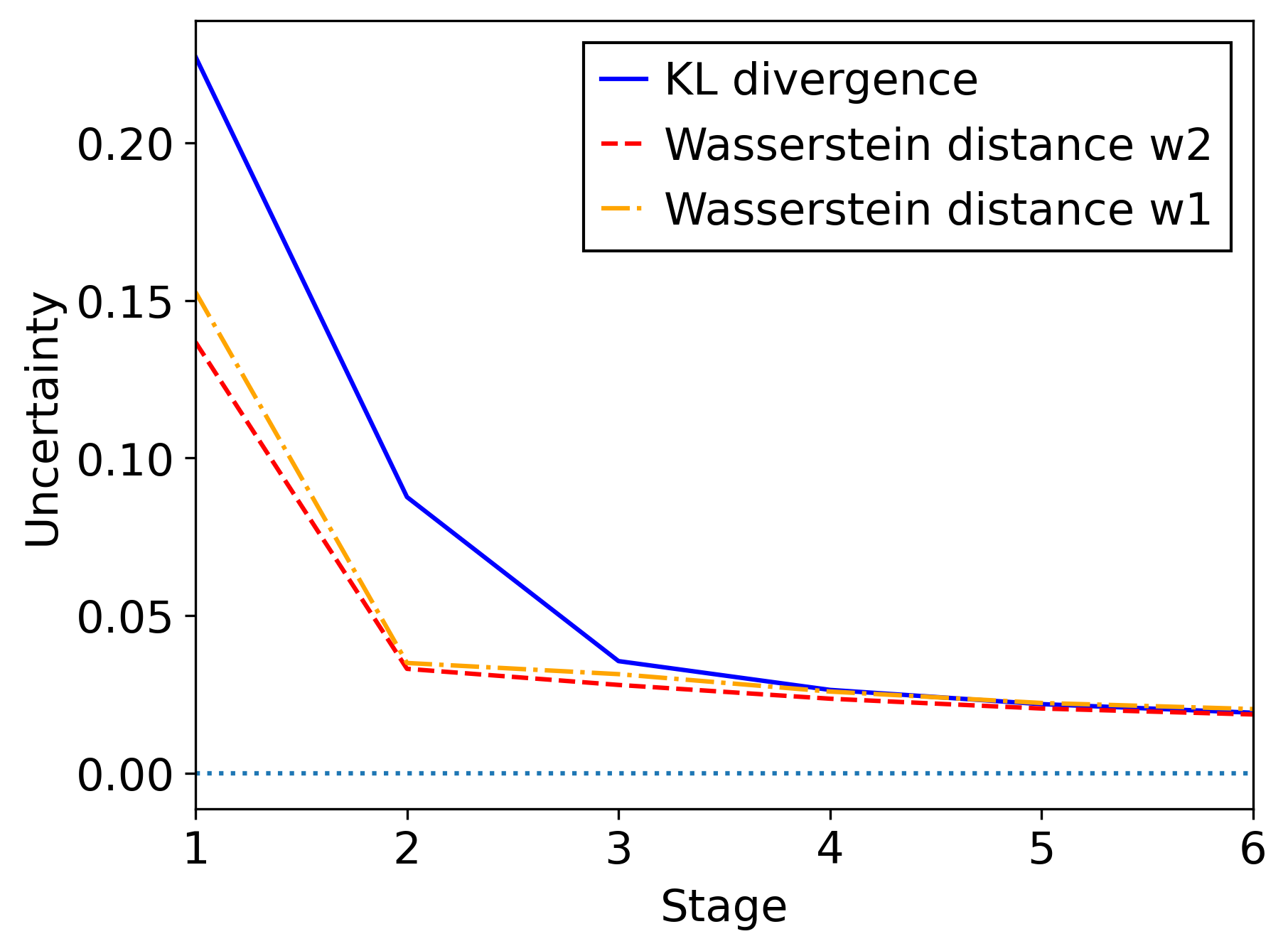}
    \caption{Uncertainty}
    \label{fig: no error expected uncertainty reverse}
  \end{subfigure}
  \caption{Distance and uncertainty evolution for Case 5 in the convection-diffusion problem, where the designs are selected using expected information gain without model discrepancy.}
  \label{fig: no error expected distance and uncertainty reverse}
\end{figure}

Here, we present a representative case in which the source term is located at $\{0.1, 0.1\}$. This location is notably closer to the design generated by the Wasserstein-based criterion (around $\{0.2, 0.2\}$) and farther from that of the KL-based design $\{0.3, 0.3\}$. In this setting, the Wasserstein-based design yields lower uncertainty in the first stage (Fig.~\ref{fig: no error expected posterior reverse}), which subsequently leads to consistently smaller uncertainty across all posterior updates (Fig.~\ref{fig: no error expected uncertainty reverse}).

The key factor underlying this behavior is the distance of the source term to the Wasserstein-based design. Geometrically, this implies that the perpendicular bisector of the line segment connecting the KL-based and Wasserstein-based designs partitions the entire square domain into two regions. The lower-left region, where the source term lies in this case, satisfies the condition for which the Wasserstein-based design achieves faster uncertainty reduction during posterior updates. When the source term lies very close to this bisector, the posterior performance of both designs becomes nearly indistinguishable. This observation further indicates that the region where the Wasserstein-based design outperforms is smaller than that of the KL-based design. Consequently, in the majority of cases without model discrepancy, the KL-based design tends to produce more effective posterior inference.

\section{Additional results of an acoustic inverse problem}
\label{apd: Numerical case: acoustic inverse}

In this appendix, we present an acoustic inverse problem as an additional example to show that the main findings of this work are not specific to the decaying Gaussian source problem. This case has been studied in previous work~\cite{wu2023large, o2022learning} and has been extended to model discrepancy setting~\cite{YANG2025118198}. In this work, we change the inferred parameter from the source strength to the source location for a clearer comparison between KL divergence and Wasserstein-2 distance.

The wave equation:
\begin{equation}
    \frac{\partial^2 u(\mathbf{r},t)}{\partial t^2}-c^2\nabla^2u(\mathbf{r},t)=0, \quad \mathbf{r} \in \Omega
\end{equation}
describes how a physical quantity propagates through space and time. For the steady-state or single-frequency case, we assume that the solution has a harmonic time dependence. By substituting an ansatz of the form
\begin{equation}
    u(\mathbf{r},t)=U(\mathbf{r})e^{i\omega t},
\end{equation}
into the wave equation, the time-dependent part factors out. This process leads to Helmholtz equation for the spatial component:
\begin{equation}
    \nabla^2 U(\mathbf{r})+k^2U(\mathbf{r})=0,
\end{equation}
where the wave number $k=\omega / c=2\pi/\lambda$, where $c$ represents the phase speed and $\lambda$ represents the wavelength. The complex acoustic pressure amplitude function $U(\mathbf{r})$ is complex-valued, which represents both the amplitude and phase information at each point $\mathbf{r}$. The Helmholtz equation is central to understanding the spatial behavior of waves under monochromatic conditions, making it invaluable in fields such as acoustics, electromagnetics, and quantum mechanics.

We consider the problem of a point source acoustic wave propagating in a two-dimensional inhomogeneous medium:
\begin{equation}
    \begin{aligned}
        -\nabla^2 U(x,y)-e^{2m(x,y)}k^2U(x,y)&=f, \quad \{x,y\}\in \Omega,\\
        f&=A\delta (x-x_0,y-y_0),\\
        \nabla U \cdot \mathbf{n}&=0 ~\text{on}~ \Gamma_\text{top},\\
        \text{PML boundary condition}~&~\text{on }~\partial \Omega/\Gamma_\text{top}.
    \end{aligned}
\end{equation}
where $e^{2m(x,y)}$ represents the spatially varying medium property, such as the local reflection coefficient or refractive index contrast. $f$ represents a time-harmonic point source of amplitude $A$, located at $\{x_0,y_0\}$ and modeled by Dirac delta function $\delta$. The computational domain $\Omega$ is a square region, subject to a reflective (Neumann) boundary condition on the top boundary $ \Gamma_\text{top}$ and surrounded by perfectly matched layers (PML) on the remaining three boundaries to simulate wave absorption. Due to the inhomogeneity, the medium’s physical properties vary with position, which leads to complex propagation behavior. Increasing $m(x,y)$ reduces the local wave speed and increases the effective refractive index, resulting in stronger reflection and denser wavefronts with more rapid spatial oscillations.

\textbf{Problem setup}: Assume the unknown physical parameter is the source $x$-location $x_0$ with true value $\frac{129}{256}$ (the grid node closest to $0.5$), and a structural model discrepancy lies in the medium properties. The true form of $m^\dagger (x,y)$ in system is given by
\begin{equation}
    m^\dagger(x,y)=\frac{2}{3} x,
\end{equation}
and the modeled $m(x,y)$ is
\begin{equation}
    m(x,y)=\frac{2}{3}x^2.
\end{equation}
A neural network (denoted by $\text{NN}$) is utilized to capture the discrepancy between the modeled and true medium properties: $\text{NN} \rightarrow \frac{2}{3}(x-x^2)$. The network takes the spatial coordinate $x$ as input and outputs a scalar medium parameter $m$ at spatial location $\{x,y\}$: $\text{NN}: x \mapsto m$. The architecture is a fully connected feed-forward network featuring a single hidden layer with five nodes, which results in a total of 16 parameters $\boldsymbol{\theta}_\text{NN} \subseteq \mathbb{R}^{16}.$

 The general goal of this case is to seek a design $\mathbf{d}$ specifying where to measure $U$ and use the measurement $\mathbf{y}=U(\mathbf{d})+\epsilon,~\epsilon\sim \mathcal{CN}(0,\sigma^2)$ to infer the source $x$-location $x_0$. A brief procedure for each sequential experiment is described by
 \begin{itemize}
     \item Find optimal $\mathbf{d}_1$ for source $x$-location $x_0$ by standard BED method.
     \item Observe $\mathbf{y}_1$ and update a posterior for $x_0$: $p(x_0|\mathbf{y}_1,\mathbf{d}_1,\boldsymbol{\theta}_\text{NN})$ with a MAP estimation $x_0^*$.
     \item Find optimal $\mathbf{d}_2$ for network parameter $\boldsymbol{\theta}_\text{NN}$ by AD-EKI method.
     \item Observe $\mathbf{y}_2$ and optimize the network parameter $\boldsymbol{\theta}_\text{NN}$ by maximizing likelihood $p(\mathbf{y}_2|\mathbf{d}_2,x_0^*,\boldsymbol{\theta}_\text{NN})$.
     \item Use the updated model $\boldsymbol{\theta}_\text{NN}^*$ to re-update a posterior $p(x_0|\mathbf{y}_1,\mathbf{d}_1,\boldsymbol{\theta}_\text{NN}^*)$.
 \end{itemize}
The posterior of each experiment will be utilized as the prior for the next experiment. And the updated model is also passed into the next experiment.

\textbf{Numerical Setting and Solver}: The computational domain $\Omega$ is defined as $[0,1.5]^2$, discretized uniformly into $128\times 128$ cells. The wave number is set to $k=5.0$ and the source is located at $\{x_0,y_0\}=\{0.5,1.25\}$ (to be exact $\{\frac{129}{256},\frac{321}{256}\}$) with known amplitude $A=1.5$. The potential measure places are constrained to the set $\{\frac{3}{64}+\frac{3}{256}i,\frac{93}{64}\}_{i=0}^{120}$. These unusual-looking fractions arise because the source and candidate design locations are snapped to the nearest grid points. The prior distribution of $x_0$ is assumed to be uniform over $[0,1.5]$, discretized into 127 equally spaced points. A finite element method solver is developed entirely in JAX for its automatic differentiation capabilities.

To compute the Wasserstein distance $W_2$ for the one-dimensional physical parameter $x_0$ on grid points, we use the analytical expression:
\begin{equation}
    W_2^2(p,q)=\int_0^1 |F_p^{-1}(t)-F_q^{-1}(t)|^2dt,
\end{equation}
where $F_p$ and $F_q$ are the cumulative distribution functions induced by the discrete probability masses $p$ and $q$.

\subsection{Results without model discrepancy}

\begin{figure}[H]
  \centering
  \begin{subfigure}[t]{0.48\textwidth}
    \centering
    \includegraphics[width=0.9\linewidth]{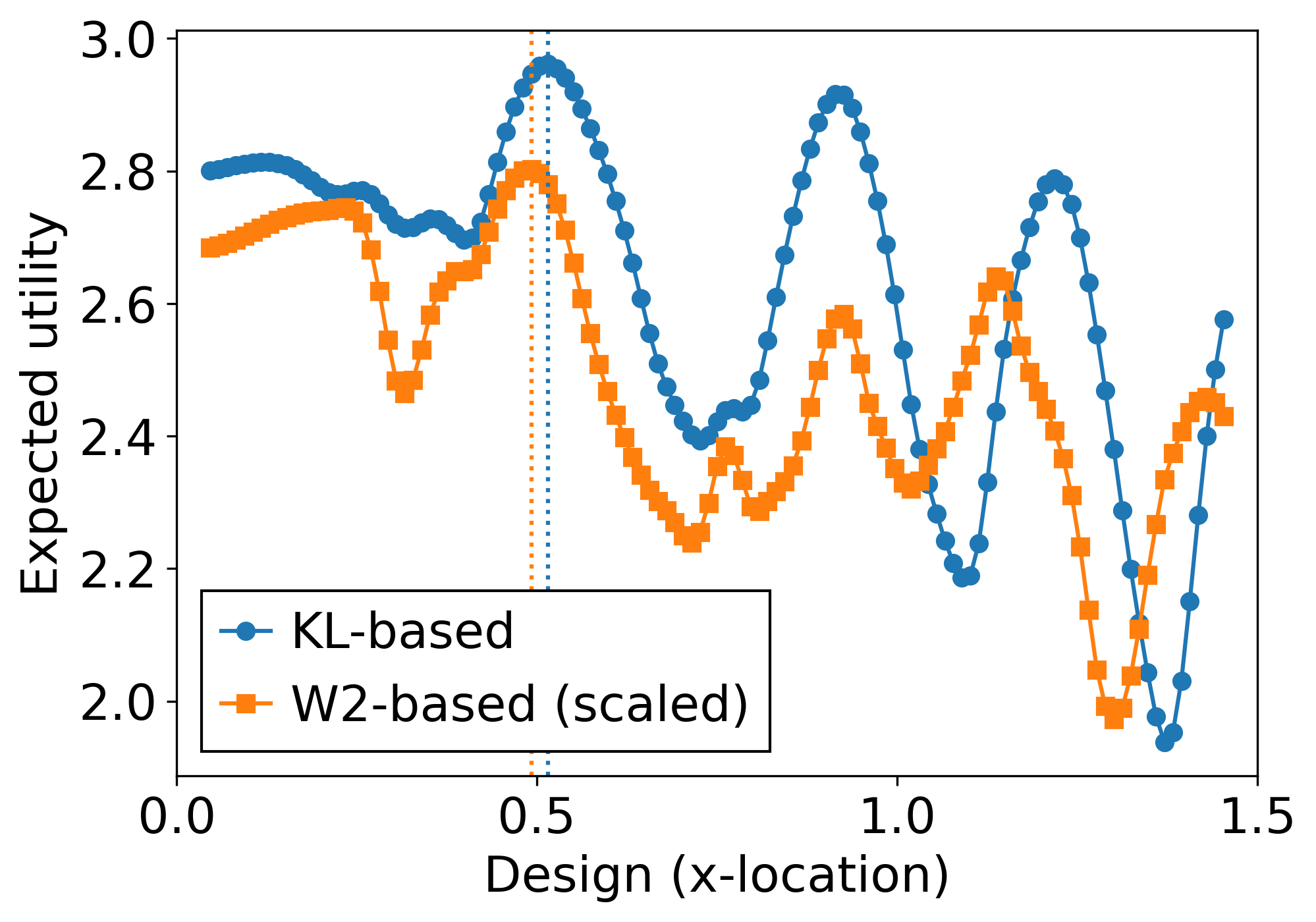}
    \caption{Without model discrepancy.}
    \label{fig: acoustic no error eig}
  \end{subfigure}
  \hfill
  \begin{subfigure}[t]{0.48\textwidth}
    \centering
    \includegraphics[width=0.9\linewidth]{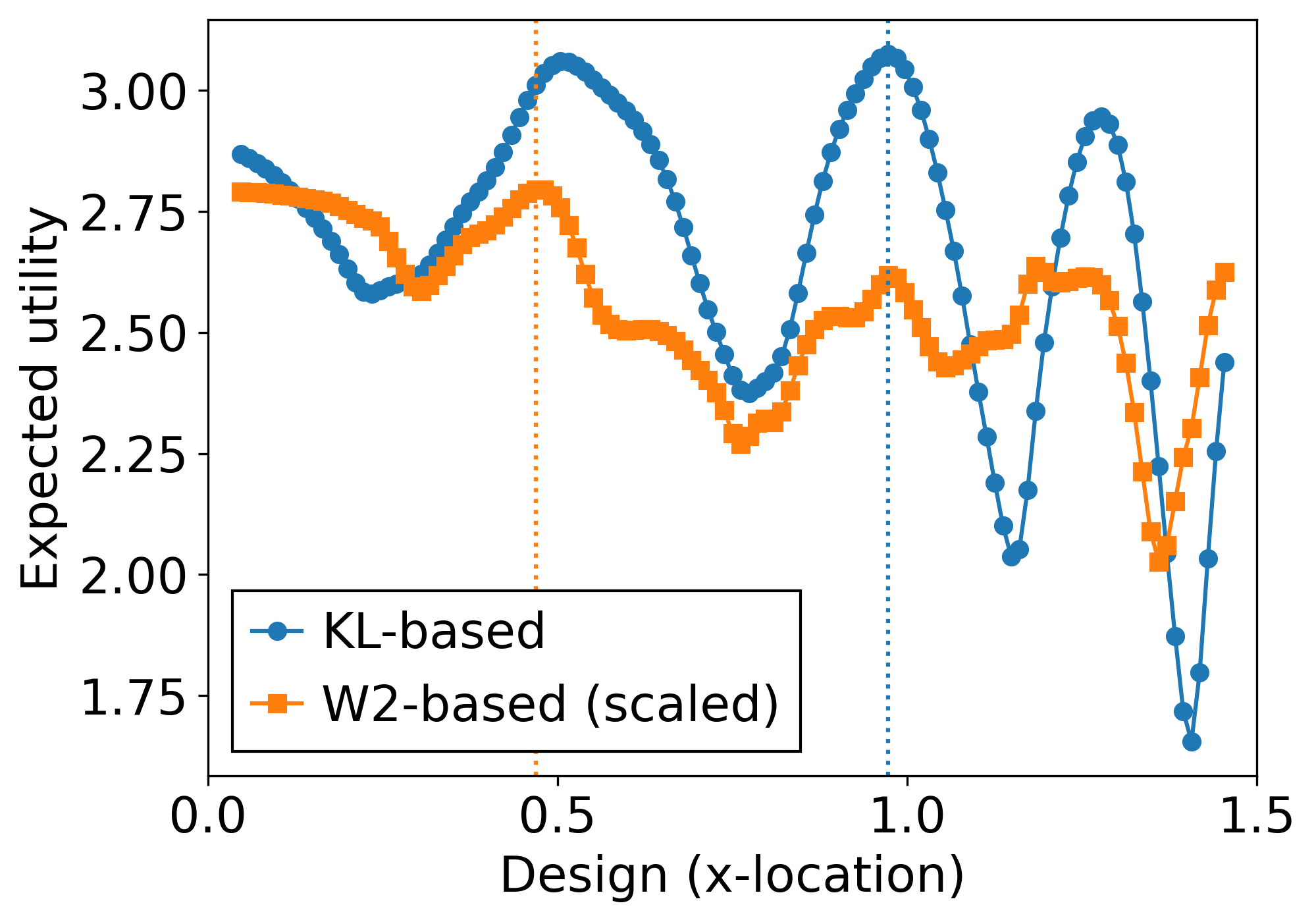}
    \caption{With model discrepancy.}
    \label{fig: acoustic error eig}
  \end{subfigure}
  \caption{Reward map for the acoustic inverse problem. The dotted lines indicate the global maxima of the expected utility.}
  \label{fig: acoustic eig}
\end{figure}

\begin{figure}[H]
    \centering
    \includegraphics[width=0.95\linewidth]{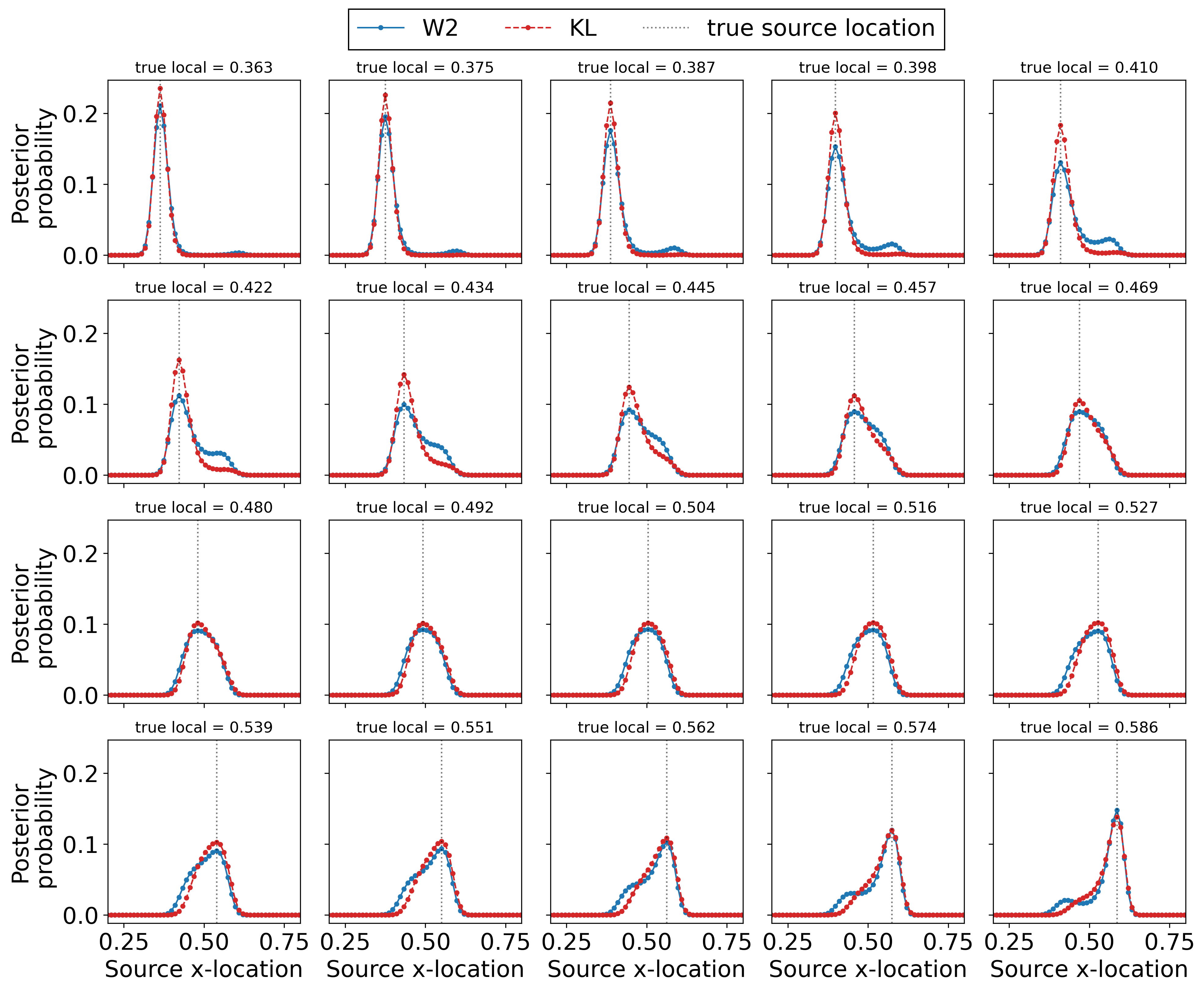}
    \caption{Posterior distribution of source $x$-location $x_0$ in the acoustic inverse problem under different true values.}
    \label{fig:Posterior under different true value}
\end{figure}

We first consider the case without model discrepancy. Figure~\ref{fig: acoustic no error eig} shows the reward map, and Fig.~\ref{fig:Posterior under different true value} shows the posterior distribution of the source $x$-coordinate $x_0$ after one measurement for different true source locations. Due to the false reward effect, the optimal design based on the expected Wasserstein distance is slightly different from that based on the expected KL divergence. As a result, regardless of the true source location, the posterior obtained from the Wasserstein-based optimal design is less concentrated than that obtained from the KL-based design. This is consistent with our conclusion that Wasserstein distance tends to produce more diffuse posteriors.

\subsection{Results with model discrepancy}

\begin{figure}[H]
    \centering
    \includegraphics[width=0.95\linewidth]{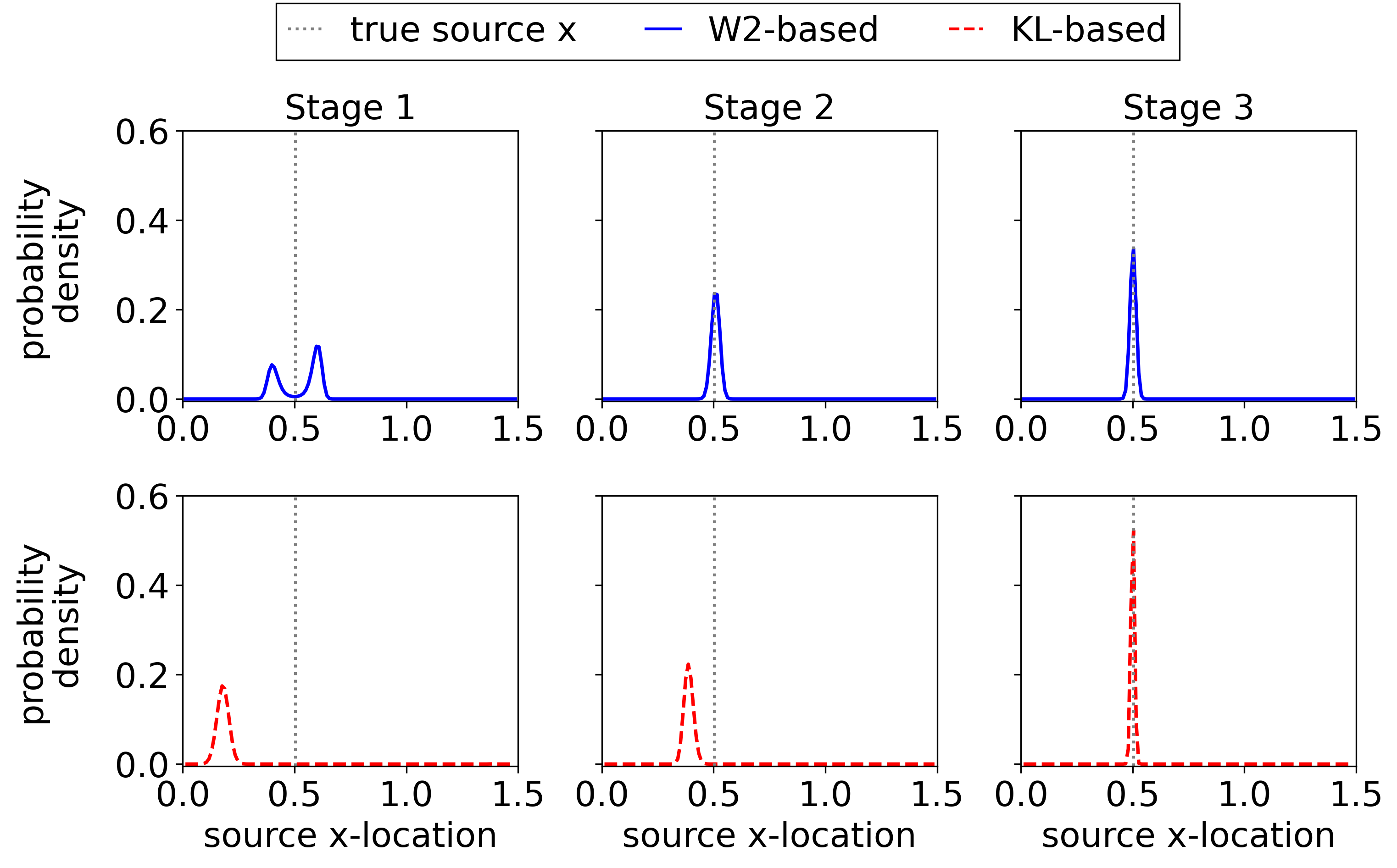}
    \caption{Posterior distribution of source $x$-location $x_0$ in the acoustic inverse problem under model discrepancy.}
    \label{fig: acoustic posterior under model discrepancy}
\end{figure}

We next consider the case with model discrepancy. Figure~\ref{fig: acoustic error eig} compares the reward maps between KL and $W_2$ with model discrepancy. In the model discrepancy case, the optimal designs selected by different criteria become more distinct. We perform three stages of the sequential AD-EKI framework with model correction and show the stage-wise posterior distributions in Fig.~\ref{fig: acoustic posterior under model discrepancy}. Due to the different design choices, the first-stage KL-based posterior is shifted farther away from the true value, and the KL-based posterior probability at the true value is much smaller than the Wasserstein-based one. Consequently, the Wasserstein-based criteria locate the true source in two stages, one stage fewer than the KL-based criterion. This is consistent with our conclusion that, under model discrepancy, Wasserstein distance can preserve more posterior uncertainty near the true value and thus requires less effort for subsequent correction.

\section{Sensitivity study for regularization factor}\label{apd:sensitivity}

\begin{figure}[H]
    \centering
    \includegraphics[width=0.8\linewidth]{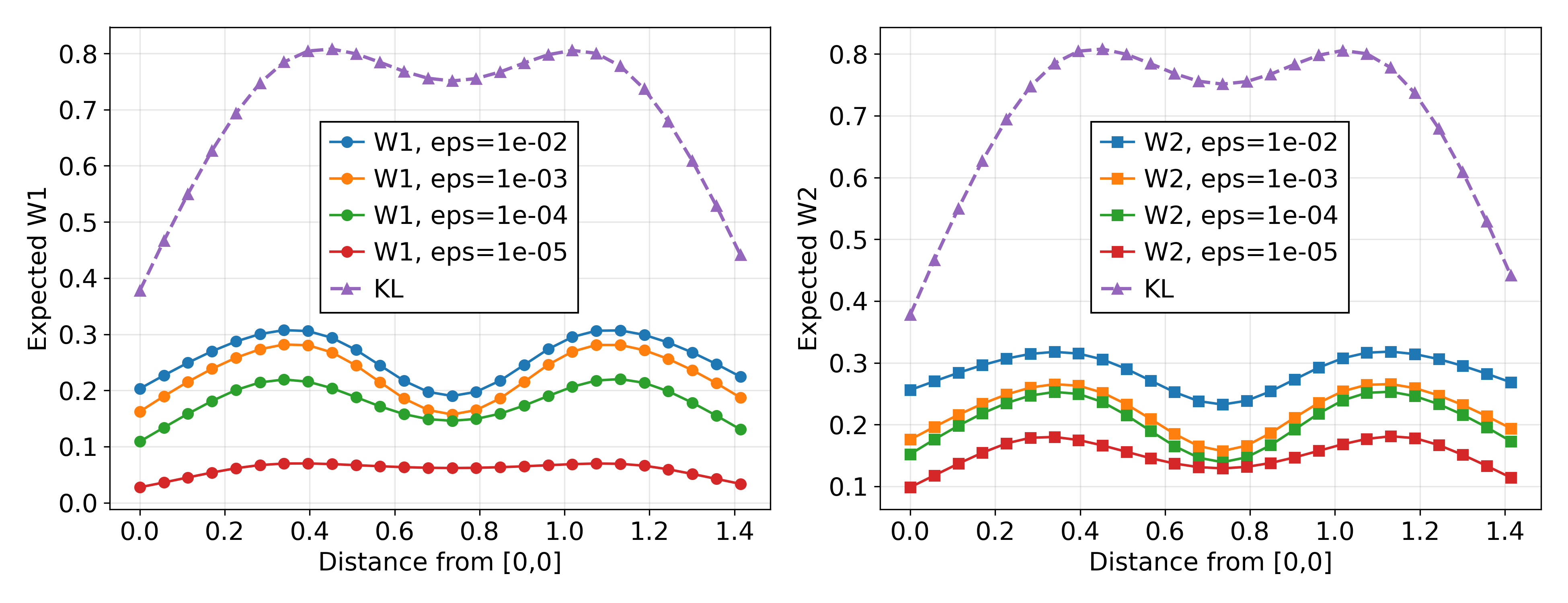}
    \caption{Expected utility changing with epsilon (for the Case 4 in the convection-diffusion problem)}
    \label{fig:sensitivity_map}
\end{figure}

\begin{figure}[H]
    \centering
    \includegraphics[width=0.5\linewidth]{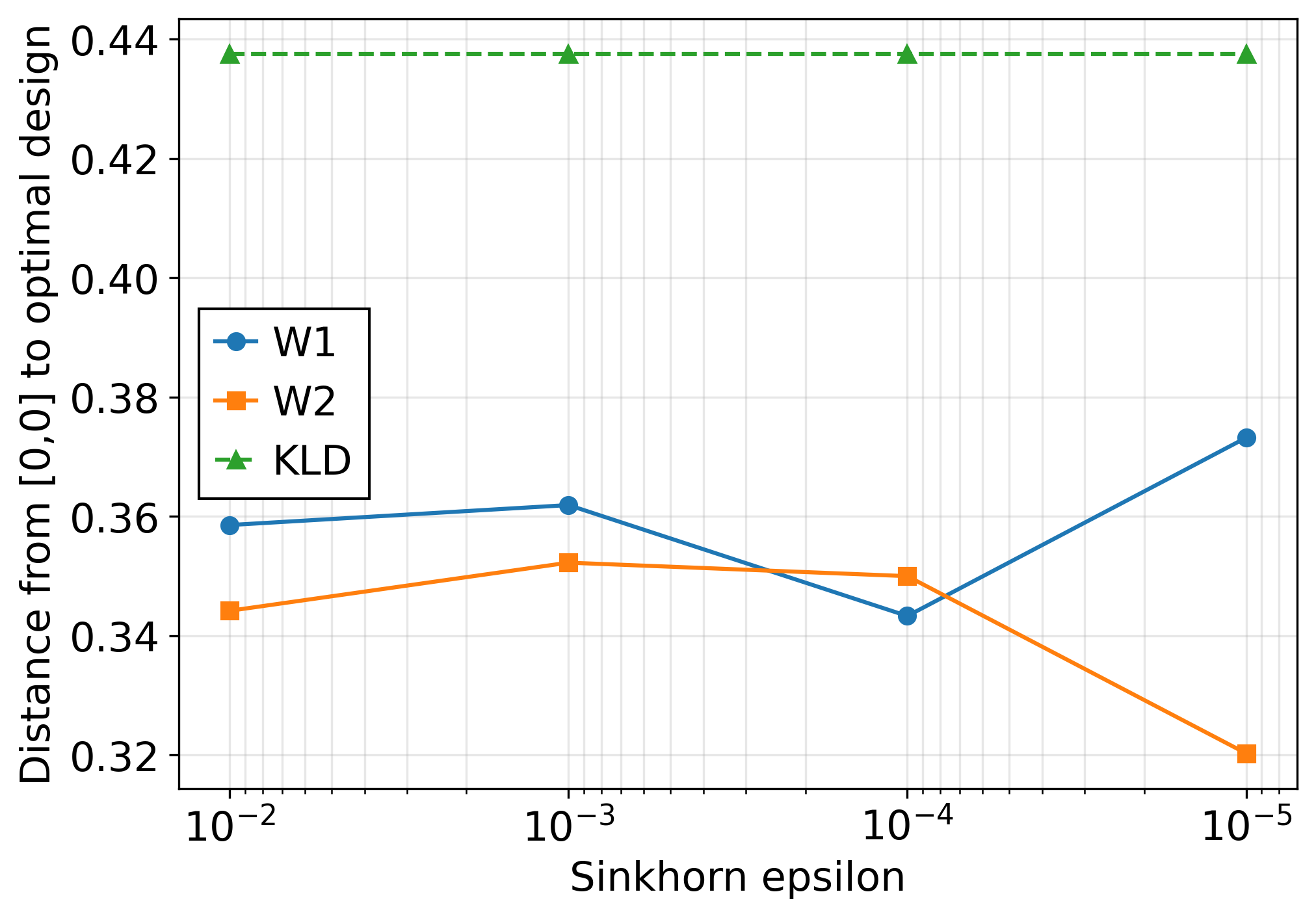}
    \caption{Optimal design position changing with epsilon (for the Case 4 in the convection-diffusion problem)}
    \label{fig:sensitivity_design}
\end{figure}

In this work, the Wasserstein utilities are computed using an entropically regularized discrete optimal transport formulation and are then used to rank candidate designs. To assess the robustness of the resulting utility landscape and design ranking, we conducted a sensitivity analysis with respect to the regularization parameter $\varepsilon$ for the convection-diffusion case under model discrepancy in Section~\ref{sec: case 4}. Specifically, we varied the regularization parameter from $1e^{-2}$ to $1e^{-5}$ and examined the resulting expected utility and optimal design. For visualization and comparison, the candidate design points were restricted to the diagonal of the field $[0,1]^2$. As $\varepsilon$ changes, the absolute values of the expected utilities vary (Fig.~\ref{fig:sensitivity_map}), but the overall shape of the utility curves and the location of the optimal design remains stable. The resulting optimal design varies only within a very small range and remains clearly separated from the KL-based optimal design (Fig.~\ref{fig:sensitivity_design}). This suggests that, in this experiment, the design optimization result is not highly sensitive to the choice of the regularization parameter, and that the observed differences between KL- and Wasserstein-based utilities are not artifacts of this regularization parameter. 

\end{document}